\documentclass{article} % For LaTeX2e
\usepackage{iclr2027_conference,times}

\usepackage{amsmath,amsfonts,bm}

\def\eqref#1{equation~\ref{#1}}
\def\1{\bm{1}}

\DeclareMathAlphabet{\mathsfit}{\encodingdefault}{\sfdefault}{m}{sl}
\SetMathAlphabet{\mathsfit}{bold}{\encodingdefault}{\sfdefault}{bx}{n}

\usepackage[utf8]{inputenc}
\usepackage[T1]{fontenc}
\usepackage{url}
\usepackage{booktabs}
\usepackage{graphicx}
\usepackage{float}
\usepackage{algorithm}
\usepackage{algpseudocode}
\usepackage{wrapfig}
\usepackage{caption}
\usepackage{afterpage}
\usepackage{microtype}
\usepackage{xcolor}
\usepackage{colortbl}
\usepackage{pifont}
\usepackage{array}
\usepackage{hyperref}
\usepackage{tikz}
\hypersetup{hidelinks}
\definecolor{ipdgreen}{HTML}{2B8A3E}
\definecolor{ipdred}{HTML}{B03A2E}
\definecolor{ipdblue}{HTML}{2F5D8A}
\definecolor{ipdindigo}{HTML}{552EDC}
\definecolor{ipdcoral}{HTML}{EB4E3D}

\title{In Their Own Words:\\Reasoning Traces Tailored for Small Models Make Them Better Reasoners}

\author{Jaehoon Kim, Kwangwook Seo \& Dongha Lee\thanks{Corresponding author.} \\
Yonsei University \\
\texttt{\{jaeh8nkim,donalee\}@yonsei.ac.kr}
}

\iclrfinalcopy % Uncomment for camera-ready version, but NOT for submission.
\begin{document}

\maketitle
\lhead{Preprint.}

\begin{abstract}
  % !TEX root = iclr2027_conference.tex
Stronger learning signals do not necessarily produce a stronger small reasoning model.
Off-policy supervised fine-tuning (SFT) supplies stronger traces that may be incompatible with the student.
On-policy post-training instead remains constrained by the quality of the student's own trajectories: reinforcement learning (RL) provides little learning signal because a small student rarely produces a correct rollout.
Consequently, effective post-training must supply stronger reasoning in a form the student can learn from.
We hypothesize that a small number of tokens with very low probability under the student can make an otherwise strong reasoning trace difficult to learn.
To test this hypothesis, we introduce Interleaved-Policy Distillation (IPD), which generates SFT data balancing teacher guidance and student compatibility.
During generation, the student replaces teacher proposals that have low probability under its own policy, allowing the teacher to continue reasoning from a prefix the student can support.
IPD improves reasoning across model families, datasets, and varying degrees of student intervention; for example, it raises the seven-benchmark average from 24.55 to 29.72 on Qwen3-0.6B/s1K.
It outperforms baselines with comparable training budgets and broadens problem-solving coverage, while ablations show that filtering traces or masking losses cannot match the gains from revising trajectories during generation.
Analyses link transfer failures to the rare tokens the student finds least likely and show that the source of the tokens also matters.
Small models reason better when they learn strong solutions in a form they can reach in their own words.

\end{abstract}

\begin{figure}[H]
  \centering
  \includegraphics[width=\linewidth]{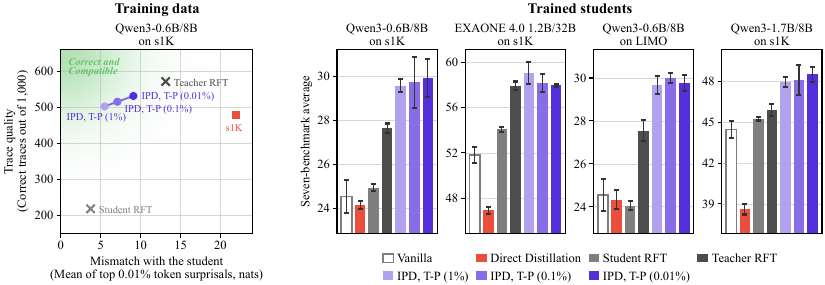}
  \caption{Training data in the primary setting and student performance across four settings. All IPD results shown use the teacher-propose (T-P) variant. \emph{Left:} Each point is one method's training set, plotted by verified correct traces and the mean of the top 0.01\% token surprisals under the student. Student RFT is compatible but teaches little; the original s1K traces used for Direct Distillation are strong but incompatible. \emph{Right:} Direct Distillation degrades the student below Vanilla, while IPD outperforms every other method in seven-benchmark average accuracy. Error bars show one standard deviation across three training seeds. \textbf{Takeaway:} correct and compatible data trains the best student.}
  \label{fig_main_results}
\end{figure}

\clearpage
% !TEX root = iclr2027_conference.tex
\section{Introduction}\label{sec_intro}

Stronger learning signals do not necessarily produce a stronger small reasoning model.
Off-policy supervised fine-tuning (SFT) provides stronger traces that may be incompatible with the student: fine-tuning Phi-4-Mini on s1K~\citep{muennighoff2025s1} reduced AIME24 accuracy from 10.0 to 3.0 and MATH500 accuracy from 71.8 to 47.0~\citep{xu2025phi4mini}.
On-policy methods instead remain constrained by the quality of the student's own rollouts.
Reinforcement learning (RL) provides little learning signal because a small student rarely produces a correct rollout.
On-policy distillation (OPD) supplies dense teacher feedback, but student rollouts can remain far from the teacher's reasoning distribution, forcing the teacher to condition on low-quality prefixes that weaken its guidance~\citep{zhu2026manyfaces}.
Effective post-training must therefore supply stronger reasoning in a form the student can learn from.

\afterpage{\begin{figure}[H]
\centering
\includegraphics[width=\linewidth]{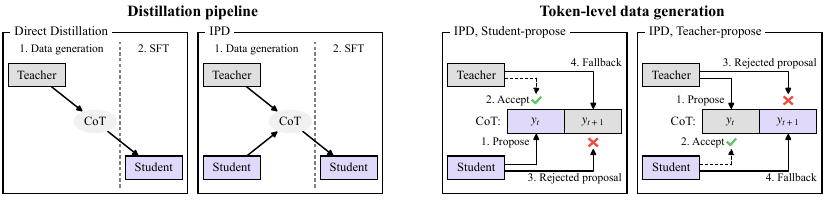}
\caption{\emph{Left:} Direct Distillation trains on fixed teacher traces; IPD constructs each trace from both policies. \emph{Right:} a proposal is accepted at $t$ and rejected at $t+1$, where the other policy supplies a fallback token. Both tokens enter the shared prefix and SFT targets.}
\label{fig_interleaved_paths}
\end{figure}
}

We hypothesize that a small number of tokens with very low student probability can make otherwise strong reasoning difficult to learn.
A trace can appear compatible on average even when it contains a few tokens the student struggles to follow.
Only a small fraction of tokens steer the reasoning that follows~\citep{wang2025beyond}, so once a problematic token enters the prefix, the rest of the solution builds on a choice the student would rarely make.
These problematic tokens cannot be treated in isolation because their effects are embedded in subsequent reasoning.
The challenge is to preserve strong reasoning while making both the trajectory and its learning targets compatible with the student.

To test this hypothesis, we introduce Interleaved-Policy Distillation (IPD), in which a teacher and student jointly construct each reasoning trace.
In teacher-propose IPD, the teacher proposes each token, the student replaces proposals whose probability under its own policy falls below a threshold $\tau$, and the teacher continues from the revised prefix.
IPD trains on the resulting interleaved traces, making the student a source of supervision as well as conditioning prefixes.
Reversing the roles gives student-propose IPD, a control that tests whether the proposal direction drives the gains (Figure~\ref{fig_interleaved_paths}).

Strong reasoning transfers more effectively when its trajectories are adapted to the student.
Across four settings spanning Qwen3 and EXAONE students~\citep{qwen2025qwen3,lg2025exaone} with 0.6B--1.7B parameters and the s1K and LIMO datasets~\citep{muennighoff2025s1,ye2025limo}, Direct Distillation leaves the student below its initial performance.
Teacher-propose IPD consistently improves performance and achieves the highest seven-benchmark average in every setting, outperforming both Student RFT and Teacher RFT~\citep{yuan2023rft} with only 1,000 s1K or 800 LIMO training questions.
For example, on Qwen3-0.6B/s1K, IPD raises the average from 24.55 to 29.72.
% 실험결과의 핵심 해석 -> 공통 패턴 -> 효과 크기를 보여주는 대표 수치 제시

This advantage persists across two orders of magnitude in $\tau$, with teacher-propose IPD exceeding Teacher RFT in all twelve setting--threshold pairs.
In the primary setting, it also outperforms RL and OPD under comparable training budgets.
Increasing Student RFT from 8 to 100 attempts more than doubles its correct traces but leaves performance unchanged.
Ablations show that filtering traces, masking token losses, mixing teacher and student distributions, and matching teacher logits on interleaved prefixes fall short of teacher-propose IPD.
These comparisons support selectively revising trajectories during generation and learning from the jointly generated tokens.

IPD reduces the rare, severe mismatches that motivate its interleaving rule.
We quantify these through the mean of the top 0.01\% token surprisals under the initial student.
In the primary setting, teacher-propose IPD reduces this mean from 21.99 nats for Direct Distillation on s1K to 5.50--9.09 nats (Figure~\ref{fig_main_results}).
Mean surprisal is a weaker guide: student-propose IPD yields lower mean surprisal than teacher-propose IPD, yet trains weaker students.
Surprisal alone, however, does not fully explain transfer: the source of the reasoning tokens also matters.
Even at a comparable top-0.01\% mean, teacher-propose IPD outperforms student-propose IPD.
This gap suggests that compatibility and token source jointly shape the value of supervision.

IPD also expands the range of problems the student can solve on its own.
At the largest matched sampling budget, it leads Vanilla and Direct Distillation on Pass@$k$ in 27 of 28 benchmark--setting pairs.
Whereas average accuracy measures how often a sampled solution is correct, Pass@$k$ measures whether any of $k$ rollouts is.
This is the quantity that limits RL, which learns from a problem only when the student produces a correct rollout.
Broader coverage therefore motivates IPD as an initialization for subsequent on-policy learning, giving RL more successes to reinforce and OPD more useful contexts for teacher feedback.
\textbf{Small models reason better when they learn strong solutions in a form that they can reach in their own words.}

\clearpage
% !TEX root = iclr2027_conference.tex
\section{Related Work}\label{sec_related}

\paragraph{Student-aligned distillation.}
Teacher traces differ in how easily a given student can learn from them.
Prior work ranks teachers using statistics of student gradients~\citep{panigrahi2026grace}, selects responses using student likelihood~\citep{zhang2025grape} or rank-to-surprisal ratios~\citep{yang2026rsr}, and contrasts teacher- and student-generated responses during training~\citep{ko2025distillm2}.
Other approaches score responses using student likelihood within local contexts~\citep{just2025local}, train on teacher prefixes the student judges sufficient and then completes~\citep{liu2026palign}, or trace teacher-originated sentences across models to select data~\citep{liu2025provenance}.

\paragraph{Reshaping teacher guidance.}
OPD provides teacher feedback on student rollouts~\citep{agarwal2024gkd}, but their prefixes can weaken that guidance~\citep{zhu2026manyfaces}.
Making teaching more useful to the student can involve adapting teacher prompts and policies~\citep{kim2024promptkd,cetin2025rlt}, optimizing the teacher with DPO on student reference-loss changes~\citep{li2025montessori}, or designing data and curricula using knowledge probes~\citep{he2026pedagogical}.
Teacher guidance can also structure the traces themselves, through expert prefixes that anchor RL rollouts~\citep{zhang2025breadbranchedrolloutsexpert} or explicit reasoning-step roles~\citep{wen2025scaffolding}.
The student can participate directly by selecting teacher continuations using perplexity~\citep{he2026genssd}, or completing and rewriting teacher-started solutions~\citep{yang2026sfd}.
Other work combines teacher ranking of student-proposed token candidates with distribution matching weighted by student--teacher entropy gaps~\citep{kim2026explainwordsimprovingreasoning}.

These methods select or construct traces over whole responses or multi-token continuations, or apply token-level feedback to sequences the student has already generated.
IPD instead uses token-level acceptance and replacement during generation and trains directly on the resulting tokens.
Speculative Knowledge Distillation (SKD) also interleaves student proposals and teacher corrections, but matches the teacher distribution at each position~\citep{xu2025speculativekd}, so student tokens serve only as conditioning prefixes.
Our primary configuration, teacher-propose IPD, also reverses SKD's proposal direction: the teacher proposes, and the student replaces rejected proposals.
Our control, student-propose IPD, shares SKD's proposal direction but retains IPD's jointly generated SFT targets.

\section{Method}\label{sec_method}
\subsection{Student Support}\label{sec_problem_setup}

Let $x$ denote a problem, $y=(y_1,\ldots,y_L)$ a reasoning trace, and $p_T$ and $p_S$ the teacher and initial student policies.
The models share a tokenizer so that each can score the other's token proposals.
Given a prefix $y_{<t}$, a candidate next token $z$ has student support at threshold $\tau$ when $p_S(z\mid x,y_{<t})\geq\tau$.
Its student surprisal is $-\log p_S(z\mid x,y_{<t})$, measured in nats.
For scale, 4.61 nats corresponds to a student probability of $1\%$, and 20 nats to about $2\times10^{-9}$.

To assess training datasets, we use \emph{upper-tail mean surprisal}: the mean of the top $1\%$, $0.1\%$, or $0.01\%$ of a dataset's token surprisals under the student.
These means reveal rare mismatches that an average over all tokens can obscure, with smaller fractions isolating more extreme mismatches.
The top-$0.01\%$ mean covers the few tokens the student finds least likely, which our analyses link to transfer failures; we call it the \emph{extreme-tail mean surprisal}.

\clearpage
\subsection{Teacher-Propose IPD}\label{sec_teacher_propose}
\begin{wrapfigure}{r}{0.34\linewidth}
  \vspace{\dimexpr-\baselineskip+4pt\relax}
  \hrule height.8pt depth0pt \kern2pt
  \captionsetup{labelsep=space,labelfont=bf,singlelinecheck=false,skip=0pt}
  \captionof{algorithm}{Interleaved-policy rollout of one trace}
  \label{alg_ipd_rollout}
  \kern2pt\hrule\kern2pt
  \footnotesize
  \begin{algorithmic}[1]
    \Require problem $x$, proposer $p_P$, gate $p_G$, threshold $\tau$, budget $L$
    \For{$t = 1,\dots,L$}
      \State $z \sim p_P(\cdot \mid x, y_{<t})$
      \If{$p_G(z \mid x, y_{<t}) < \tau$}
        \State $z \sim p_G(\cdot \mid x, y_{<t})$
      \EndIf
      \State $y_t \gets z$
      \State \textbf{if} $y_t$ terminates \textbf{then break}
    \EndFor
    \State \Return $y$
  \end{algorithmic}
  \kern2pt\hrule
  \vspace{\dimexpr-\baselineskip-10pt\relax}
\end{wrapfigure}

Our primary configuration, teacher-propose IPD (T-P), lets the teacher lead generation and the student intervene at proposals it finds unlikely.
Both policies condition on the shared prefix $y_{<t}$.
The teacher samples a token $z$, which is accepted if $p_S(z\mid x,y_{<t})\geq\tau$.
Otherwise, the student samples a replacement, or fallback token, from its own policy without a second support test.
Both policies remain frozen during generation.

Algorithm~\ref{alg_ipd_rollout} expresses this rule using a proposer $p_P$ and a gate $p_G$, which scores proposals and supplies replacements.
T-P sets $p_P=p_T$ and $p_G=p_S$; student-propose IPD (S-P) reverses these roles.
T-P thus adapts teacher reasoning to the student, whereas S-P corrects student reasoning with teacher tokens.
Because each committed token extends the shared prefix, a replacement also redirects the proposals that follow.

We then fine-tune the student on the resulting traces with ordinary SFT.
Every committed token, including each student replacement, is therefore an SFT target.

At $\tau=0$, T-P accepts every proposal and reduces to teacher-only generation; at the other extreme, rejecting every proposal reduces it to student-only generation.
At intermediate thresholds, accepted teacher proposals satisfy
\begin{equation}\label{eq_ceiling}
-\log p_S(y_t\mid x,y_{<t})\leq\log(1/\tau).
\end{equation}
The ceilings are 4.61, 6.91, and 9.21 nats for $\tau=1\%,0.1\%,0.01\%$.
Student replacements are not gated, so a dataset's upper-tail mean surprisal can exceed these ceilings.
The threshold therefore controls how readily the student replaces teacher proposals and redirects subsequent reasoning.

\section{IPD Robustly Improves Reasoning Distillation}\label{sec_experiments}
\subsection{Consistent Gains Across Models and Datasets}\label{sec_generality}

Teacher-propose IPD achieves the highest seven-benchmark average across all four model--dataset settings.
Table~\ref{tab_family_results} covers Qwen3-0.6B/8B on s1K and LIMO, EXAONE 1.2B/32B on s1K, and Qwen3-1.7B/8B on s1K, with model pairs written as student/teacher.
Its baselines span the trade-off between strong reasoning and student compatibility.
Direct Distillation uses the released traces of s1K-1.1 and LIMO-v2 (s1K and LIMO hereafter).
Student RFT and Teacher RFT use correctness-verified traces sampled for the same questions from the student and the teacher, respectively.
Vanilla is the initial student.
Percentages in IPD labels denote the acceptance threshold $\tau$, and we use $\tau=0.1\%$ as the representative configuration.

\label{sec_dataset_construction}
Within each setting, all methods use the same generation budget, SFT schedule, and evaluation, so they differ only in their training data.
Each data-generation method has a budget of eight attempts per question.
Unsolved IPD questions contribute partial traces of at most 128 tokens generated by the same IPD condition~\citep{ji2025upft}.
We select the SFT schedule through an independent held-out hyperparameter search for each setting (Appendix~\ref{app_hyperparameters}), yielding 15 epochs at $3\times10^{-5}$ for the Qwen3 settings and 30 epochs at $10^{-5}$ for EXAONE.
Evaluation uses Avg@8 on AIME24/25/26 and HMMT26, and Avg@4 on MATH500, OlympiadBench, and GPQA-D~\citep{maa2025aime,hmmt2026,lightman2024verify,he2024olympiadbench,rein2024gpqa}, averaged over three training seeds.

The results order these methods by how well their data combine strength with compatibility.
Direct Distillation, whose traces are the least compatible, falls below Vanilla in all four settings; Student RFT gains less than Teacher RFT, and Teacher RFT less than T-P, in every setting.
Neither strength nor compatibility is enough alone.
In our primary setting, Qwen3-0.6B/8B on s1K, T-P at $\tau=0.1\%$ raises the average from 24.55 to 29.72, compared with 24.15 for Direct Distillation and 27.65 for Teacher RFT.
Across settings it improves on Vanilla by 3.64--6.32 points, and the gain is broad: it holds in all 28 benchmark--setting pairs.
\textbf{Adapting teacher reasoning to the student yields consistent gains across the tested models and datasets.}

\clearpage
\begin{table}[H]
  \caption{Accuracy (\%) on seven benchmarks in four settings. Entries are mean $\pm$ sample standard deviation across three seeds; each seed sets training and evaluation, or only evaluation for Vanilla. Average is the unweighted benchmark mean, and IPD percentages are $\tau$. Blue and red shading mark gains and losses relative to Vanilla; bold marks the best value per column and setting.}
  \label{tab_family_results}
  \vspace{3pt}
  \centering
  \scriptsize
  \setlength{\tabcolsep}{2pt}
  \begin{tabular}{lcccccccc}
    \toprule
    \textbf{Method} & \textbf{AIME24} & \textbf{AIME25} & \textbf{AIME26} & \textbf{HMMT26} & \textbf{MATH500} & \textbf{Olympiad} & \textbf{GPQA-D} & \textbf{Average} \\
    \midrule
    \rowcolor{gray!15}\multicolumn{9}{l}{\textbf{Qwen3-0.6B/8B on s1K}} \\
    \hspace{0.75em}Vanilla & $6.25_{\pm 1.67}$ & $15.00_{\pm 1.91}$ & $10.42_{\pm 1.10}$ & $8.96_{\pm 0.79}$ & $71.07_{\pm 0.40}$ & $37.82_{\pm 0.26}$ & $22.35_{\pm 0.91}$ & $24.55$ \\
    \hspace{0.75em}Direct Distillation & \cellcolor{ipdcoral!44}$4.86_{\pm 0.24}$ & \cellcolor{ipdindigo!18}$18.33_{\pm 2.20}$ & \cellcolor{ipdcoral!5}$10.14_{\pm 0.64}$ & \cellcolor{ipdcoral!11}$8.46_{\pm 1.22}$ & \cellcolor{ipdindigo!2}$71.63_{\pm 0.75}$ & \cellcolor{ipdcoral!10}$35.87_{\pm 0.27}$ & \cellcolor{ipdcoral!23}$19.78_{\pm 0.07}$ & \cellcolor{ipdcoral!3}$24.15$ \\
    \hspace{0.75em}Student RFT & \cellcolor{ipdcoral!4}$6.11_{\pm 1.34}$ & \cellcolor{ipdcoral!7}$14.44_{\pm 0.64}$ & \cellcolor{ipdcoral!3}$10.28_{\pm 0.48}$ & \cellcolor{ipdcoral!28}$7.70_{\pm 0.95}$ & \cellcolor{ipdindigo!1}$71.42_{\pm 1.35}$ & \cellcolor{ipdindigo!2}$38.24_{\pm 0.27}$ & \cellcolor{ipdindigo!16}$26.43_{\pm 1.07}$ & \cellcolor{ipdindigo!2}$24.95$ \\
    \hspace{0.75em}Teacher RFT & \cellcolor{ipdindigo!16}$7.50_{\pm 1.91}$ & \cellcolor{ipdindigo!26}$19.86_{\pm 0.64}$ & \cellcolor{ipdindigo!5}$11.11_{\pm 3.07}$ & \cellcolor{ipdindigo!26}$11.87_{\pm 2.22}$ & \cellcolor{ipdindigo!13}$74.40_{\pm 0.40}$ & \cellcolor{ipdindigo!14}$41.31_{\pm 0.25}$ & \cellcolor{ipdindigo!21}$27.48_{\pm 0.65}$ & \cellcolor{ipdindigo!12}$27.65$ \\
    \hspace{0.75em}IPD, S-P (1\%) & \cellcolor{ipdindigo!64}$\mathbf{11.25}_{\pm 0.42}$ & \cellcolor{ipdindigo!24}$19.58_{\pm 0.00}$ & \cellcolor{ipdindigo!44}$\mathbf{16.11}_{\pm 2.71}$ & \cellcolor{ipdindigo!30}$12.37_{\pm 2.19}$ & \cellcolor{ipdindigo!22}$76.45_{\pm 1.55}$ & \cellcolor{ipdindigo!21}$43.13_{\pm 0.41}$ & \cellcolor{ipdindigo!10}$24.79_{\pm 1.03}$ & \cellcolor{ipdindigo!18}$29.10$ \\
    \hspace{0.75em}IPD, S-P (0.1\%) & \cellcolor{ipdindigo!41}$9.44_{\pm 3.01}$ & \cellcolor{ipdindigo!26}$19.86_{\pm 0.24}$ & \cellcolor{ipdindigo!29}$14.17_{\pm 0.72}$ & \cellcolor{ipdindigo!36}$13.01_{\pm 0.44}$ & \cellcolor{ipdindigo!20}$76.02_{\pm 0.98}$ & \cellcolor{ipdindigo!15}$41.59_{\pm 0.70}$ & \cellcolor{ipdindigo!8}$24.28_{\pm 0.65}$ & \cellcolor{ipdindigo!15}$28.34$ \\
    \hspace{0.75em}IPD, S-P (0.01\%) & \cellcolor{ipdindigo!39}$9.31_{\pm 1.05}$ & \cellcolor{ipdindigo!19}$18.61_{\pm 1.46}$ & \cellcolor{ipdindigo!39}$15.56_{\pm 1.58}$ & \cellcolor{ipdindigo!18}$10.98_{\pm 1.00}$ & \cellcolor{ipdindigo!15}$74.77_{\pm 0.50}$ & \cellcolor{ipdindigo!16}$41.86_{\pm 0.60}$ & \cellcolor{ipdindigo!31}$30.05_{\pm 1.98}$ & \cellcolor{ipdindigo!17}$28.73$ \\
    \hspace{0.75em}IPD, T-P (1\%) & \cellcolor{ipdindigo!52}$10.28_{\pm 2.44}$ & \cellcolor{ipdindigo!27}$20.14_{\pm 2.10}$ & \cellcolor{ipdindigo!13}$12.08_{\pm 1.10}$ & \cellcolor{ipdindigo!42}$13.64_{\pm 1.65}$ & \cellcolor{ipdindigo!27}$\mathbf{77.78}_{\pm 0.35}$ & \cellcolor{ipdindigo!21}$\mathbf{43.18}_{\pm 0.71}$ & \cellcolor{ipdindigo!30}$29.97_{\pm 1.50}$ & \cellcolor{ipdindigo!20}$29.58$ \\
    \hspace{0.75em}IPD, T-P (0.1\%) & \cellcolor{ipdindigo!50}$10.14_{\pm 3.15}$ & \cellcolor{ipdindigo!30}$\mathbf{20.56}_{\pm 0.48}$ & \cellcolor{ipdindigo!28}$14.03_{\pm 2.77}$ & \cellcolor{ipdindigo!30}$12.37_{\pm 1.09}$ & \cellcolor{ipdindigo!21}$76.30_{\pm 1.06}$ & \cellcolor{ipdindigo!20}$42.89_{\pm 0.85}$ & \cellcolor{ipdindigo!38}$31.78_{\pm 1.39}$ & \cellcolor{ipdindigo!21}$29.72$ \\
    \hspace{0.75em}IPD, T-P (0.01\%) & \cellcolor{ipdindigo!25}$8.19_{\pm 1.05}$ & \cellcolor{ipdindigo!26}$19.86_{\pm 2.96}$ & \cellcolor{ipdindigo!21}$13.19_{\pm 1.46}$ & \cellcolor{ipdindigo!46}$\mathbf{14.14}_{\pm 2.87}$ & \cellcolor{ipdindigo!17}$75.43_{\pm 0.50}$ & \cellcolor{ipdindigo!14}$41.41_{\pm 0.09}$ & \cellcolor{ipdindigo!59}$\mathbf{37.16}_{\pm 1.58}$ & \cellcolor{ipdindigo!21}$\mathbf{29.91}$ \\
    \midrule
    \rowcolor{gray!15}\multicolumn{9}{l}{\textbf{EXAONE 4.0 1.2B/32B on s1K}} \\
    \hspace{0.75em}Vanilla & $38.47_{\pm 1.34}$ & $41.11_{\pm 3.49}$ & $49.58_{\pm 2.92}$ & $30.18_{\pm 2.74}$ & $89.82_{\pm 0.20}$ & $66.01_{\pm 1.15}$ & $47.69_{\pm 1.19}$ & $51.84$ \\
    \hspace{0.75em}Direct Distillation & \cellcolor{ipdcoral!27}$33.19_{\pm 1.68}$ & \cellcolor{ipdcoral!21}$36.94_{\pm 1.97}$ & \cellcolor{ipdcoral!28}$44.03_{\pm 0.24}$ & \cellcolor{ipdindigo!6}$31.69_{\pm 0.87}$ & $89.77_{\pm 0.32}$ & \cellcolor{ipdcoral!36}$58.86_{\pm 0.06}$ & \cellcolor{ipdcoral!69}$33.92_{\pm 0.70}$ & \cellcolor{ipdcoral!25}$46.92$ \\
    \hspace{0.75em}Student RFT & \cellcolor{ipdindigo!8}$40.42_{\pm 1.50}$ & \cellcolor{ipdindigo!13}$44.44_{\pm 1.68}$ & \cellcolor{ipdindigo!17}$53.75_{\pm 0.42}$ & \cellcolor{ipdindigo!11}$32.95_{\pm 0.38}$ & \cellcolor{ipdindigo!3}$90.48_{\pm 0.43}$ & \cellcolor{ipdindigo!7}$67.68_{\pm 0.72}$ & \cellcolor{ipdindigo!5}$48.82_{\pm 2.27}$ & \cellcolor{ipdindigo!9}$54.08$ \\
    \hspace{0.75em}Teacher RFT & \cellcolor{ipdindigo!40}$\mathbf{48.47}_{\pm 0.48}$ & \cellcolor{ipdindigo!51}$53.89_{\pm 1.73}$ & \cellcolor{ipdindigo!31}$57.36_{\pm 0.64}$ & \cellcolor{ipdindigo!24}$\mathbf{36.11}_{\pm 1.22}$ & \cellcolor{ipdindigo!6}$91.30_{\pm 0.96}$ & \cellcolor{ipdindigo!9}$68.20_{\pm 0.62}$ & \cellcolor{ipdindigo!10}$50.21_{\pm 0.39}$ & \cellcolor{ipdindigo!24}$57.93$ \\
    \hspace{0.75em}IPD, S-P (1\%) & \cellcolor{ipdindigo!23}$44.17_{\pm 2.32}$ & \cellcolor{ipdindigo!40}$51.11_{\pm 2.14}$ & \cellcolor{ipdindigo!29}$56.81_{\pm 2.37}$ & \cellcolor{ipdindigo!13}$33.33_{\pm 0.66}$ & \cellcolor{ipdindigo!5}$91.07_{\pm 0.69}$ & \cellcolor{ipdindigo!12}$68.93_{\pm 0.63}$ & \cellcolor{ipdindigo!8}$49.58_{\pm 1.36}$ & \cellcolor{ipdindigo!18}$56.43$ \\
    \hspace{0.75em}IPD, S-P (0.1\%) & \cellcolor{ipdindigo!27}$45.14_{\pm 0.87}$ & \cellcolor{ipdindigo!26}$47.50_{\pm 0.72}$ & \cellcolor{ipdindigo!23}$55.28_{\pm 1.73}$ & \cellcolor{ipdindigo!14}$33.71_{\pm 1.37}$ & \cellcolor{ipdindigo!4}$90.75_{\pm 0.26}$ & \cellcolor{ipdindigo!10}$68.55_{\pm 0.80}$ & \cellcolor{ipdindigo!11}$50.46_{\pm 0.84}$ & \cellcolor{ipdindigo!16}$55.91$ \\
    \hspace{0.75em}IPD, S-P (0.01\%) & \cellcolor{ipdindigo!19}$43.19_{\pm 2.37}$ & \cellcolor{ipdindigo!32}$49.03_{\pm 1.05}$ & \cellcolor{ipdindigo!25}$55.83_{\pm 1.82}$ & \cellcolor{ipdindigo!19}$34.85_{\pm 0.76}$ & $89.93_{\pm 0.86}$ & \cellcolor{ipdindigo!11}$68.73_{\pm 0.74}$ & \cellcolor{ipdindigo!13}$50.93_{\pm 0.95}$ & \cellcolor{ipdindigo!17}$56.07$ \\
    \hspace{0.75em}IPD, T-P (1\%) & \cellcolor{ipdindigo!38}$47.92_{\pm 1.10}$ & \cellcolor{ipdindigo!63}$56.81_{\pm 1.68}$ & \cellcolor{ipdindigo!45}$60.83_{\pm 1.50}$ & \cellcolor{ipdindigo!20}$35.10_{\pm 1.53}$ & \cellcolor{ipdindigo!8}$91.75_{\pm 0.39}$ & \cellcolor{ipdindigo!14}$\mathbf{69.57}_{\pm 0.62}$ & \cellcolor{ipdindigo!15}$51.35_{\pm 2.21}$ & \cellcolor{ipdindigo!29}$\mathbf{59.05}$ \\
    \hspace{0.75em}IPD, T-P (0.1\%) & \cellcolor{ipdindigo!33}$46.81_{\pm 0.64}$ & \cellcolor{ipdindigo!54}$54.58_{\pm 3.31}$ & \cellcolor{ipdindigo!47}$\mathbf{61.39}_{\pm 2.55}$ & \cellcolor{ipdindigo!11}$32.83_{\pm 1.16}$ & \cellcolor{ipdindigo!8}$\mathbf{91.77}_{\pm 0.38}$ & \cellcolor{ipdindigo!7}$67.74_{\pm 0.26}$ & \cellcolor{ipdindigo!17}$\mathbf{51.98}_{\pm 1.20}$ & \cellcolor{ipdindigo!25}$58.16$ \\
    \hspace{0.75em}IPD, T-P (0.01\%) & \cellcolor{ipdindigo!33}$46.81_{\pm 2.06}$ & \cellcolor{ipdindigo!65}$\mathbf{57.36}_{\pm 0.87}$ & \cellcolor{ipdindigo!46}$60.97_{\pm 4.05}$ & \cellcolor{ipdindigo!11}$32.95_{\pm 3.41}$ & \cellcolor{ipdindigo!6}$91.38_{\pm 0.21}$ & \cellcolor{ipdindigo!14}$69.54_{\pm 0.61}$ & \cellcolor{ipdcoral!5}$46.72_{\pm 1.49}$ & \cellcolor{ipdindigo!24}$57.96$ \\
    \midrule
    \rowcolor{gray!15}\multicolumn{9}{l}{\textbf{Qwen3-0.6B/8B on LIMO}} \\
    \hspace{0.75em}Vanilla & $6.25_{\pm 1.67}$ & $15.00_{\pm 1.91}$ & $10.42_{\pm 1.10}$ & $8.96_{\pm 0.79}$ & $71.07_{\pm 0.40}$ & $37.82_{\pm 0.26}$ & $22.35_{\pm 0.91}$ & $24.55$ \\
    \hspace{0.75em}Direct Distillation & \cellcolor{ipdindigo!25}$8.19_{\pm 0.87}$ & \cellcolor{ipdindigo!8}$16.53_{\pm 1.34}$ & \cellcolor{ipdcoral!3}$10.28_{\pm 0.96}$ & \cellcolor{ipdcoral!17}$8.21_{\pm 0.44}$ & \cellcolor{ipdindigo!1}$71.25_{\pm 0.61}$ & \cellcolor{ipdindigo!3}$38.55_{\pm 0.54}$ & \cellcolor{ipdcoral!44}$17.38_{\pm 0.26}$ & \cellcolor{ipdcoral!2}$24.34$ \\
    \hspace{0.75em}Student RFT & \cellcolor{ipdcoral!62}$4.31_{\pm 2.55}$ & \cellcolor{ipdindigo!1}$15.14_{\pm 1.88}$ & \cellcolor{ipdcoral!16}$9.58_{\pm 1.67}$ & \cellcolor{ipdcoral!6}$8.71_{\pm 1.14}$ & \cellcolor{ipdcoral!10}$69.07_{\pm 2.29}$ & \cellcolor{ipdcoral!2}$37.50_{\pm 0.39}$ & \cellcolor{ipdindigo!7}$24.12_{\pm 0.66}$ & \cellcolor{ipdcoral!4}$24.06$ \\
    \hspace{0.75em}Teacher RFT & \cellcolor{ipdindigo!20}$7.78_{\pm 0.24}$ & \cellcolor{ipdindigo!7}$16.39_{\pm 1.97}$ & \cellcolor{ipdindigo!44}$16.11_{\pm 1.05}$ & \cellcolor{ipdindigo!26}$11.87_{\pm 0.44}$ & \cellcolor{ipdindigo!13}$74.43_{\pm 0.88}$ & \cellcolor{ipdindigo!14}$41.23_{\pm 0.43}$ & \cellcolor{ipdindigo!10}$24.96_{\pm 3.03}$ & \cellcolor{ipdindigo!12}$27.54$ \\
    \hspace{0.75em}IPD, S-P (1\%) & \cellcolor{ipdindigo!27}$8.33_{\pm 1.50}$ & \cellcolor{ipdindigo!20}$18.75_{\pm 0.00}$ & \cellcolor{ipdindigo!32}$14.58_{\pm 1.10}$ & \cellcolor{ipdindigo!51}$\mathbf{14.65}_{\pm 0.58}$ & \cellcolor{ipdindigo!22}$76.60_{\pm 0.74}$ & \cellcolor{ipdindigo!21}$43.15_{\pm 0.31}$ & \cellcolor{ipdindigo!19}$26.98_{\pm 1.71}$ & \cellcolor{ipdindigo!18}$29.01$ \\
    \hspace{0.75em}IPD, S-P (0.1\%) & \cellcolor{ipdindigo!21}$7.92_{\pm 0.42}$ & \cellcolor{ipdindigo!21}$18.89_{\pm 0.48}$ & \cellcolor{ipdindigo!19}$12.92_{\pm 0.72}$ & \cellcolor{ipdindigo!25}$11.74_{\pm 0.76}$ & \cellcolor{ipdindigo!18}$75.58_{\pm 0.29}$ & \cellcolor{ipdindigo!15}$41.47_{\pm 0.32}$ & \cellcolor{ipdindigo!21}$27.48_{\pm 1.90}$ & \cellcolor{ipdindigo!14}$28.00$ \\
    \hspace{0.75em}IPD, S-P (0.01\%) & \cellcolor{ipdindigo!36}$9.03_{\pm 0.64}$ & \cellcolor{ipdindigo!6}$16.11_{\pm 1.05}$ & \cellcolor{ipdindigo!32}$14.58_{\pm 1.50}$ & \cellcolor{ipdindigo!20}$11.24_{\pm 1.87}$ & \cellcolor{ipdindigo!16}$75.07_{\pm 0.42}$ & \cellcolor{ipdindigo!14}$41.20_{\pm 0.49}$ & \cellcolor{ipdindigo!10}$24.79_{\pm 0.98}$ & \cellcolor{ipdindigo!12}$27.43$ \\
    \hspace{0.75em}IPD, T-P (1\%) & \cellcolor{ipdindigo!43}$9.58_{\pm 1.91}$ & \cellcolor{ipdindigo!23}$\mathbf{19.31}_{\pm 0.87}$ & \cellcolor{ipdindigo!30}$14.31_{\pm 2.14}$ & \cellcolor{ipdindigo!34}$12.75_{\pm 0.22}$ & \cellcolor{ipdindigo!28}$\mathbf{77.98}_{\pm 0.40}$ & \cellcolor{ipdindigo!29}$\mathbf{45.07}_{\pm 0.06}$ & \cellcolor{ipdindigo!25}$28.70_{\pm 1.10}$ & \cellcolor{ipdindigo!20}$29.67$ \\
    \hspace{0.75em}IPD, T-P (0.1\%) & \cellcolor{ipdindigo!39}$9.31_{\pm 1.05}$ & \cellcolor{ipdindigo!16}$17.92_{\pm 0.42}$ & \cellcolor{ipdindigo!45}$\mathbf{16.25}_{\pm 1.10}$ & \cellcolor{ipdindigo!35}$12.88_{\pm 1.00}$ & \cellcolor{ipdindigo!23}$76.78_{\pm 0.65}$ & \cellcolor{ipdindigo!25}$44.09_{\pm 0.60}$ & \cellcolor{ipdindigo!42}$\mathbf{32.74}_{\pm 2.15}$ & \cellcolor{ipdindigo!22}$\mathbf{30.00}$ \\
    \hspace{0.75em}IPD, T-P (0.01\%) & \cellcolor{ipdindigo!68}$\mathbf{11.53}_{\pm 1.46}$ & \cellcolor{ipdindigo!23}$\mathbf{19.31}_{\pm 1.05}$ & \cellcolor{ipdindigo!35}$15.00_{\pm 1.25}$ & \cellcolor{ipdindigo!41}$13.51_{\pm 0.58}$ & \cellcolor{ipdindigo!25}$77.22_{\pm 0.46}$ & \cellcolor{ipdindigo!24}$43.79_{\pm 0.54}$ & \cellcolor{ipdindigo!23}$28.03_{\pm 0.33}$ & \cellcolor{ipdindigo!21}$29.77$ \\
    \midrule
    \rowcolor{gray!15}\multicolumn{9}{l}{\textbf{Qwen3-1.7B/8B on s1K}} \\
    \hspace{0.75em}Vanilla & $31.81_{\pm 2.29}$ & $32.64_{\pm 1.68}$ & $33.47_{\pm 2.44}$ & $25.38_{\pm 1.37}$ & $88.72_{\pm 0.28}$ & $57.15_{\pm 1.00}$ & $42.21_{\pm 1.07}$ & $44.48$ \\
    \hspace{0.75em}Direct Distillation & \cellcolor{ipdcoral!44}$24.86_{\pm 0.96}$ & \cellcolor{ipdcoral!37}$26.67_{\pm 0.72}$ & \cellcolor{ipdcoral!35}$27.64_{\pm 2.51}$ & \cellcolor{ipdcoral!13}$23.74_{\pm 1.58}$ & \cellcolor{ipdcoral!8}$87.10_{\pm 0.10}$ & \cellcolor{ipdcoral!15}$54.17_{\pm 0.13}$ & \cellcolor{ipdcoral!81}$26.01_{\pm 1.46}$ & \cellcolor{ipdcoral!29}$38.60$ \\
    \hspace{0.75em}Student RFT & \cellcolor{ipdindigo!6}$33.19_{\pm 1.88}$ & \cellcolor{ipdindigo!11}$35.42_{\pm 2.20}$ & \cellcolor{ipdindigo!9}$35.69_{\pm 1.58}$ & \cellcolor{ipdindigo!8}$27.27_{\pm 1.00}$ & \cellcolor{ipdcoral!2}$88.33_{\pm 0.45}$ & \cellcolor{ipdindigo!2}$57.62_{\pm 0.41}$ & \cellcolor{ipdcoral!15}$39.27_{\pm 0.91}$ & \cellcolor{ipdindigo!3}$45.26$ \\
    \hspace{0.75em}Teacher RFT & \cellcolor{ipdcoral!5}$30.97_{\pm 1.73}$ & \cellcolor{ipdindigo!9}$35.00_{\pm 0.42}$ & \cellcolor{ipdindigo!23}$39.31_{\pm 2.06}$ & \cellcolor{ipdindigo!6}$26.89_{\pm 1.74}$ & \cellcolor{ipdindigo!1}$88.95_{\pm 0.18}$ & \cellcolor{ipdindigo!4}$58.10_{\pm 0.39}$ & \cellcolor{ipdcoral!1}$42.05_{\pm 0.55}$ & \cellcolor{ipdindigo!6}$45.90$ \\
    \hspace{0.75em}IPD, S-P (1\%) & \cellcolor{ipdindigo!11}$34.44_{\pm 0.87}$ & \cellcolor{ipdindigo!10}$35.14_{\pm 0.48}$ & \cellcolor{ipdindigo!24}$39.44_{\pm 0.64}$ & \cellcolor{ipdindigo!9}$27.65_{\pm 2.37}$ & \cellcolor{ipdindigo!3}$89.48_{\pm 0.13}$ & \cellcolor{ipdindigo!7}$58.79_{\pm 0.79}$ & \cellcolor{ipdcoral!7}$40.78_{\pm 1.34}$ & \cellcolor{ipdindigo!8}$46.53$ \\
    \hspace{0.75em}IPD, S-P (0.1\%) & \cellcolor{ipdindigo!9}$34.03_{\pm 2.96}$ & \cellcolor{ipdindigo!3}$33.33_{\pm 1.67}$ & \cellcolor{ipdindigo!28}$40.56_{\pm 2.41}$ & \cellcolor{ipdindigo!11}$28.16_{\pm 0.44}$ & \cellcolor{ipdindigo!3}$89.38_{\pm 0.51}$ & \cellcolor{ipdindigo!7}$58.96_{\pm 0.39}$ & \cellcolor{ipdindigo!3}$42.89_{\pm 0.81}$ & \cellcolor{ipdindigo!9}$46.76$ \\
    \hspace{0.75em}IPD, S-P (0.01\%) & \cellcolor{ipdindigo!11}$34.58_{\pm 1.25}$ & \cellcolor{ipdindigo!8}$34.58_{\pm 1.25}$ & \cellcolor{ipdindigo!11}$36.11_{\pm 2.93}$ & \cellcolor{ipdindigo!18}$29.92_{\pm 0.66}$ & \cellcolor{ipdindigo!2}$89.13_{\pm 0.38}$ & \cellcolor{ipdindigo!6}$58.75_{\pm 0.32}$ & \cellcolor{ipdcoral!1}$41.96_{\pm 2.08}$ & \cellcolor{ipdindigo!8}$46.44$ \\
    \hspace{0.75em}IPD, T-P (1\%) & \cellcolor{ipdindigo!7}$33.61_{\pm 1.97}$ & \cellcolor{ipdindigo!18}$37.08_{\pm 1.10}$ & \cellcolor{ipdindigo!27}$40.28_{\pm 1.05}$ & \cellcolor{ipdindigo!11}$28.03_{\pm 1.65}$ & \cellcolor{ipdindigo!5}$\mathbf{90.00}_{\pm 0.46}$ & \cellcolor{ipdindigo!10}$59.57_{\pm 0.27}$ & \cellcolor{ipdindigo!19}$\mathbf{47.05}_{\pm 0.72}$ & \cellcolor{ipdindigo!14}$47.95$ \\
    \hspace{0.75em}IPD, T-P (0.1\%) & \cellcolor{ipdindigo!19}$\mathbf{36.67}_{\pm 1.82}$ & \cellcolor{ipdindigo!19}$\mathbf{37.50}_{\pm 2.32}$ & \cellcolor{ipdindigo!23}$39.17_{\pm 2.92}$ & \cellcolor{ipdindigo!16}$29.29_{\pm 2.58}$ & \cellcolor{ipdindigo!5}$89.95_{\pm 0.46}$ & \cellcolor{ipdindigo!11}$\mathbf{59.92}_{\pm 0.37}$ & \cellcolor{ipdindigo!8}$44.32_{\pm 0.66}$ & \cellcolor{ipdindigo!15}$48.12$ \\
    \hspace{0.75em}IPD, T-P (0.01\%) & \cellcolor{ipdindigo!11}$34.44_{\pm 3.07}$ & \cellcolor{ipdindigo!18}$37.08_{\pm 2.32}$ & \cellcolor{ipdindigo!33}$\mathbf{41.81}_{\pm 1.88}$ & \cellcolor{ipdindigo!25}$\mathbf{31.57}_{\pm 1.09}$ & \cellcolor{ipdindigo!4}$89.78_{\pm 0.24}$ & \cellcolor{ipdindigo!11}$59.80_{\pm 0.20}$ & \cellcolor{ipdindigo!12}$45.29_{\pm 0.96}$ & \cellcolor{ipdindigo!16}$\mathbf{48.54}$ \\
    \bottomrule
  \end{tabular}
\end{table}

\subsection{Threshold Robustness and Proposal Direction}\label{sec_robustness}

The teacher-propose advantage does not depend on tuning the threshold.
The threshold sets how often the student intervenes: in the primary setting, the strictest threshold makes the student intervene about five times as often as the most permissive one (Table~\ref{tab_quality_compatibility}).
Yet T-P exceeds Teacher RFT in all twelve setting--threshold comparisons of Table~\ref{tab_family_results}, and its average changes little within a setting.
Which threshold is best varies across training seeds, but the best average in every setting and seed comes from T-P (Appendix~\ref{app_seed_results}).
Even the most permissive threshold already replaces the proposals the student finds least likely, and intervening more often adds little.
This is the pattern our hypothesis predicts: the obstacle lies in a few extreme tokens, not in the bulk of the teacher's trace.

Proposal direction, in contrast, matters.
Student-propose IPD interleaves the same frozen models under the same thresholds and differs only in which model proposes.
It also improves on Vanilla throughout, so interleaving helps in either direction, but even its best threshold falls below T-P's worst in every setting.
When the student leads, the teacher contributes only corrections, and the traces remain mostly the student's own.
S-P traces are shorter and solve fewer questions, whereas T-P traces are as long as the teacher's and solve nearly as many (Table~\ref{tab_quality_compatibility}).
\textbf{At every tested threshold, the student learns more from adapted teacher reasoning than from corrected student reasoning.}

\clearpage
\subsection{Gains Under Data and Budget Controls}\label{sec_controls}

IPD retains its advantage over additional sampling, RL, and OPD under comparable budgets.
In the primary setting, increasing Student RFT from 8 to 100 attempts raises the number of correct full traces from 219 to 468, but average accuracy remains essentially unchanged (24.95 versus 24.88).
IPD reaches 29.72 with 516 correct full traces, exceeding Teacher RFT at 27.65 despite its 572 traces (Tables~\ref{tab_compute_budget} and~\ref{tab_confound_controls}).
The quantity of verified solutions alone therefore does not explain its advantage.

\begin{table}[H]
\caption{Sampling and training budgets for Qwen3-0.6B/8B on s1K. Relative budget weights the maximum rollouts by the sizes of the models that generate and score them, normalized to IPD (Appendix~\ref{app_on_policy}). Every baseline except Student RFT-8 spends close to IPD's budget.}
\label{tab_compute_budget}
\centering\scriptsize
\begin{tabular}{lcccc}
\toprule
Method & Correct full traces & Training steps & Maximum rollouts & Relative budget \\
\midrule
Student RFT-8 & 219 & \phantom{1,}937 & \phantom{10}8,000 & $0.07\times$ \\
Student RFT-100 & 468 & \phantom{1,}937 & 100,000 & $0.87\times$ \\
Teacher RFT & 572 & \phantom{1,}937 & \phantom{10}8,000 & $0.93\times$ \\
GRPO & N/A & 1,000 & 128,000 & $1.12\times$ \\
OPD & N/A & \phantom{1,}500 & \phantom{10}8,000 & $1.00\times$ \\
IPD, T-P (0.1\%) & 516 & \phantom{1,}937 & \phantom{10}8,000 & $1.00\times$ \\
\bottomrule
\end{tabular}
\end{table}

\begin{table}[H]
\caption{Additional-sampling and on-policy baselines for Qwen3-0.6B/8B on s1K, under the budgets of Table~\ref{tab_compute_budget}. Scoring follows Table~\ref{tab_family_results}; bold marks the best value per column. IPD has the best average, ahead of heavier sampling, GRPO, and OPD, and has the best value on five of the seven benchmarks.}
\label{tab_confound_controls}
\centering\scriptsize\setlength{\tabcolsep}{2pt}
\begin{tabular}{lcccccccc}
\toprule
Method & AIME24 & AIME25 & AIME26 & HMMT26 & MATH500 & Olympiad & GPQA-D & Average \\
\midrule
    Vanilla & $6.25_{\pm 1.67}$ & $15.00_{\pm 1.91}$ & $10.42_{\pm 1.10}$ & $8.96_{\pm 0.79}$ & $71.07_{\pm 0.40}$ & $37.82_{\pm 0.26}$ & $22.35_{\pm 0.91}$ & $24.55$ \\
    Direct Distillation & $4.86_{\pm 0.24}$ & $18.33_{\pm 2.20}$ & $10.14_{\pm 0.64}$ & $8.46_{\pm 1.22}$ & $71.63_{\pm 0.75}$ & $35.87_{\pm 0.27}$ & $19.78_{\pm 0.07}$ & $24.15$ \\
    Student RFT-8 & $6.11_{\pm 1.34}$ & $14.44_{\pm 0.64}$ & $10.28_{\pm 0.48}$ & $7.70_{\pm 0.95}$ & $71.42_{\pm 1.35}$ & $38.24_{\pm 0.27}$ & $26.43_{\pm 1.07}$ & $24.95$ \\
    Student RFT-100 & $3.89_{\pm 1.05}$ & $16.39_{\pm 1.73}$ & $10.83_{\pm 1.25}$ & $9.34_{\pm 2.15}$ & $71.45_{\pm 0.33}$ & $38.70_{\pm 0.47}$ & $23.53_{\pm 0.62}$ & $24.88$ \\
    Teacher RFT & $7.50_{\pm 1.91}$ & $19.86_{\pm 0.64}$ & $11.11_{\pm 3.07}$ & $11.87_{\pm 2.22}$ & $74.40_{\pm 0.40}$ & $41.31_{\pm 0.25}$ & $27.48_{\pm 0.65}$ & $27.65$ \\
    GRPO & $5.83_{\pm 2.20}$ & $17.78_{\pm 1.92}$ & $12.22_{\pm 1.68}$ & $10.61_{\pm 1.37}$ & $72.62_{\pm 0.35}$ & $39.26_{\pm 0.67}$ & $18.60_{\pm 0.84}$ & $25.27$ \\
    OPD & $8.47_{\pm 1.68}$ & $\mathbf{20.97}_{\pm 0.24}$ & $\mathbf{15.28}_{\pm 1.46}$ & $10.61_{\pm 1.52}$ & $75.32_{\pm 0.28}$ & $42.84_{\pm 0.26}$ & $26.05_{\pm 1.42}$ & $28.51$ \\
    IPD, T-P (0.1\%) & $\mathbf{10.14}_{\pm 3.15}$ & $20.56_{\pm 0.48}$ & $14.03_{\pm 2.77}$ & $\mathbf{12.37}_{\pm 1.09}$ & $\mathbf{76.30}_{\pm 1.06}$ & $\mathbf{42.89}_{\pm 0.85}$ & $\mathbf{31.78}_{\pm 1.39}$ & $\mathbf{29.72}$ \\
\bottomrule
\end{tabular}
\end{table}

Another possible confound is the partial trace: for each question it fails to solve, IPD still trains on the first 128 tokens of an attempt.
We test the effect of partial traces from both sides.
First, we train IPD on its 516 full traces alone, and its average drops only slightly.
Second, we give Direct Distillation the same treatment: we keep the original s1K traces either for the questions IPD solved or for the questions whose s1K trace our verifier accepts, and train each selection with and without 128-token partial traces for the remaining questions.
Partial traces again add a little, yet every variant stays far below IPD (Appendix~\ref{app_data_controls}, Tables~\ref{tab_data_composition} and~\ref{tab_data_control_results}).
They help, but do not explain IPD's advantage.

For on-policy comparisons, we use GRPO~\citep{shao2024deepseekmath} and OPD~\citep{agarwal2024gkd}.
GRPO learns from verified correctness rewards, while OPD minimizes reverse KL to the teacher on newly sampled student rollouts.
Under comparable budgets, they average 25.27 and 28.51, both below IPD's 29.72; GRPO gains only 0.72 points over Vanilla.
Neither correctness rewards nor dense teacher feedback on student rollouts matches the gains from IPD's interleaved targets.
Appendix~\ref{app_on_policy} provides the baseline protocols and budget accounting.
\textbf{IPD's gains are not explained by trace quantity or partial traces from failed attempts, and they exceed RL and OPD under comparable budgets.}

\subsection{Ablating Trajectory Construction and Learning Targets}\label{sec_mechanism}

During distillation, a teacher token the student would almost never produce causes two problems: the student learns to reproduce it, and later reasoning builds on it.
IPD replaces only such tokens, lets the teacher continue, and trains on the result.
Tables~\ref{tab_mechanism_design} and~\ref{tab_mechanism_results} test controls that drop these steps.

The first controls keep completed traces and change only their supervision.
Trace filtering ranks s1K traces by mean surprisal, keeps the lower half as full traces, and cuts the rest to 128-token partial traces.
Loss masking drops the loss on tokens whose student probability falls below $1\%$, $0.1\%$, or $0.01\%$.
Neither regenerates anything: filtering discards the reasoning in the traces it cuts, and masking still trains on a continuation built on the masked token.
Both fall short of IPD, so the reasoning after an unsupported token has to be revised, not only the token's supervision.

\clearpage
\begin{table}[H]
\caption{Trace construction and learning target of each control. A token from the source passes the gate or is replaced by the fallback; $q_\beta$ is the geometric mean of the two policies, CE is cross-entropy, and KL is forward KL divergence. Only IPD both gates generation and trains on the realized tokens.}
\label{tab_mechanism_design}
\centering\scriptsize\setlength{\tabcolsep}{5pt}
\begin{tabular}{lcccl}
\toprule
Method & Token source & Selection or gate & Fallback & Supervision target \\
\midrule
Trace filtering & Corpus & Trace surprisal & None & Corpus token (CE) \\
Loss masking & Corpus & $p_S(z)\geq\tau$ (loss only) & None & Unmasked token (CE) \\
Dist. Interpo. & $q_\beta$ & None & None & Sampled token (CE) \\
SKD & Student & Teacher top-$K$ & Teacher & Teacher dist. (KL) \\
IPD, S-P & Student & $p_T(z)\geq\tau$ & Teacher & Realized token (CE) \\
IPD, T-P & Teacher & $p_S(z)\geq\tau$ & Student & Realized token (CE) \\
\bottomrule
\end{tabular}
\end{table}

The remaining controls generate new traces but change how the policies are combined or what the student learns.
Distribution interpolation samples every token from $q_\beta\propto p_T^\beta p_S^{1-\beta}$ with $\beta=0.5$, a geometric mean that blends the two policies everywhere but sets no floor on student probability.
SKD also interleaves the policies, accepting student proposals among the teacher's top 25 tokens and replacing the rest with teacher samples, but it trains the student to match the teacher's distribution at every position~\citep{xu2025speculativekd}, so student tokens shape its prefixes without becoming targets.
Both perform below T-P; interpolation, which weights the two policies equally, falls between the S-P and T-P averages of Table~\ref{tab_family_results}.
Appendix~\ref{app_mechanism_controls} gives their specifications.

\begin{table}[H]
\caption{Results for the controls of Table~\ref{tab_mechanism_design}, for Qwen3-0.6B/8B on s1K. Loss-masking percentages are probability thresholds. Scoring follows Table~\ref{tab_family_results}; bold marks the best value per column. No control reaches IPD's average: filtering and masking fall far behind, and interpolation and SKD come close.}
\label{tab_mechanism_results}
\centering\scriptsize\setlength{\tabcolsep}{2pt}
\begin{tabular}{lcccccccc}
\toprule
Method & AIME24 & AIME25 & AIME26 & HMMT26 & MATH500 & Olympiad & GPQA-D & Average \\
\midrule
    Vanilla & $6.25_{\pm 1.67}$ & $15.00_{\pm 1.91}$ & $10.42_{\pm 1.10}$ & $8.96_{\pm 0.79}$ & $71.07_{\pm 0.40}$ & $37.82_{\pm 0.26}$ & $22.35_{\pm 0.91}$ & $24.55$ \\
Trace filtering & $6.25_{\pm 1.10}$ & $18.19_{\pm 0.48}$ & $10.00_{\pm 0.72}$ & $10.48_{\pm 2.69}$ & $74.03_{\pm 1.04}$ & $38.50_{\pm 1.34}$ & $28.79_{\pm 1.08}$ & $26.61$ \\
Loss masking (1\%) & $7.08_{\pm 0.42}$ & $18.47_{\pm 1.05}$ & $9.72_{\pm 1.05}$ & $9.34_{\pm 1.79}$ & $73.15_{\pm 0.18}$ & $37.64_{\pm 0.47}$ & $22.77_{\pm 2.44}$ & $25.45$ \\
Loss masking (0.1\%) & $6.11_{\pm 0.24}$ & $17.50_{\pm 1.10}$ & $10.28_{\pm 2.44}$ & $10.35_{\pm 3.09}$ & $72.55_{\pm 0.25}$ & $36.82_{\pm 0.23}$ & $22.64_{\pm 1.68}$ & $25.18$ \\
Loss masking (0.01\%) & $6.67_{\pm 1.25}$ & $18.33_{\pm 1.10}$ & $9.58_{\pm 0.00}$ & $7.83_{\pm 1.16}$ & $71.90_{\pm 0.95}$ & $36.86_{\pm 0.25}$ & $23.06_{\pm 0.96}$ & $24.89$ \\
Dist. Interpo. & $10.00_{\pm 0.72}$ & $19.03_{\pm 1.05}$ & $14.72_{\pm 1.58}$ & $\mathbf{13.26}_{\pm 1.97}$ & $\mathbf{77.82}_{\pm 0.68}$ & $\mathbf{43.71}_{\pm 0.09}$ & $28.03_{\pm 1.20}$ & $29.51$ \\
SKD & $\mathbf{13.19}_{\pm 1.20}$ & $19.58_{\pm 1.25}$ & $\mathbf{15.83}_{\pm 0.83}$ & $12.50_{\pm 1.00}$ & $69.08_{\pm 0.98}$ & $40.69_{\pm 0.65}$ & $\mathbf{32.95}_{\pm 0.51}$ & $29.12$ \\
\midrule
    IPD, T-P (0.1\%) & $10.14_{\pm 3.15}$ & $\mathbf{20.56}_{\pm 0.48}$ & $14.03_{\pm 2.77}$ & $12.37_{\pm 1.09}$ & $76.30_{\pm 1.06}$ & $42.89_{\pm 0.85}$ & $31.78_{\pm 1.39}$ & $\mathbf{29.72}$ \\
\bottomrule
\end{tabular}
\end{table}

Controls that drop any of these choices fall short: regenerating the continuation, replacing selectively, and learning from the generated tokens each contribute.
\textbf{The strongest transfer comes from revising teacher reasoning during generation and retaining both policies' contributions as SFT targets.}

\section{What IPD Changes and What It Preserves}\label{sec_analysis}
\subsection{IPD Reduces Extreme-Tail Mismatch}\label{sec_tradeoff}

\begin{figure}[H]
\centering
\includegraphics[width=\linewidth]{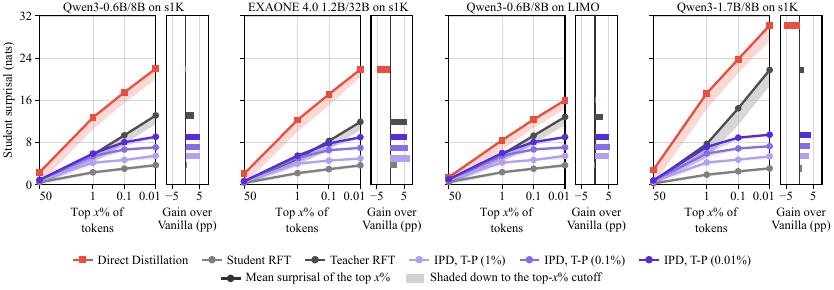}
\caption{Student-surprisal tails and gains in four settings. Curves show the mean of the top $x\%$ token surprisals, shaded to the top-$x\%$ cutoff; bars show gains over Vanilla at the mean of the top 0.01\%. Curves separate as $x$ narrows; T-P, below Direct Distillation and Teacher RFT, gains most.}
\label{fig_tail_profiles}
\end{figure}

Teacher-propose IPD keeps the teacher's reasoning while removing its rare, severe mismatches with the student.
Figure~\ref{fig_tail_profiles} follows each dataset's upper-tail mean surprisal as the tail narrows, beside the gain of the student trained on it; Table~\ref{tab_quality_compatibility} gives the diagnostics for the primary setting.
The curves fan out toward the extreme tail, where Direct Distillation is highest, Teacher RFT next, and T-P lower in every setting.
Student RFT is lower still, since its traces are the student's own.

\begin{table}[H]
\caption{Dataset diagnostics for Qwen3-0.6B/8B on s1K. \# Correct counts verified full traces; median length covers them, or all source traces for Direct Distillation. T-P keeps most of Teacher RFT's correct traces while lowering the mean of the top 0.01\% token surprisals.}
\label{tab_quality_compatibility}
\centering\scriptsize\setlength{\tabcolsep}{5pt}
\begin{tabular}{lcccccc}
\toprule
Condition & \# Correct & Med. len. (tok) & Fallback (\%) & Mean surp. (nat) & Top 0.01\% surp. (nat) & Eval. avg. \\
\midrule
Direct Distillation & 479 & 9,549.5 & N/A & 1.1229 & 21.9850 & 24.15 \\
Student RFT & 219 & 6,133.0 & N/A & 0.2366 & \phantom{0}3.7291 & 24.95 \\
Teacher RFT & 572 & 9,211.0 & N/A & 0.4400 & 13.1366 & 27.65 \\
\midrule
IPD, S-P (1\%) & 455 & 7,537.0 & 1.795 & 0.3584 & 11.5742 & 29.10 \\
IPD, S-P (0.1\%) & 367 & 7,026.0 & 0.969 & 0.3616 & 10.5838 & 28.34 \\
IPD, S-P (0.01\%) & 312 & 7,488.5 & 0.542 & 0.3732 & \phantom{0}9.9513 & 28.73 \\
\midrule
IPD, T-P (1\%) & 503 & 9,391.0 & 0.728 & 0.4182 & \phantom{0}5.4950 & 29.58 \\
IPD, T-P (0.1\%) & 516 & 9,583.5 & 0.241 & 0.4715 & \phantom{0}7.0902 & 29.72 \\
IPD, T-P (0.01\%) & 532 & 9,371.0 & 0.136 & 0.4937 & \phantom{0}9.0937 & 29.91 \\
\bottomrule
\end{tabular}
\end{table}

\clearpage

Transfer follows the same order in all four settings: Direct Distillation leaves the student below Vanilla, Teacher RFT improves it, and T-P improves it most.
Student RFT has the lowest extreme-tail mean surprisal, yet its self-generated traces train weaker students than either IPD direction in every setting.
T-P keeps the teacher's learning signal, as the teacher writes almost every token, yet its extreme-tail mean surprisal falls well below Teacher RFT's.
Neither correct-trace count, median length, nor mean surprisal explains this: the best configuration has neither the most correct traces nor the longest ones, and S-P datasets have lower mean surprisal than T-P datasets yet train weaker students (Tables~\ref{tab_quality_compatibility} and~\ref{tab_quality_compatibility_settings}).
\textbf{Low extreme-tail surprisal supports transfer when paired with an external learning signal.}

\subsection{Teacher Proposals Add Value Beyond Compatibility}\label{sec_proposal_source}
\begin{figure}[H]
\centering
\includegraphics[width=\linewidth]{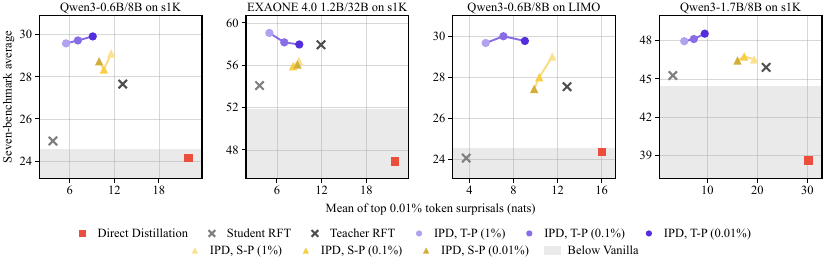}
\caption{Student performance against the mean of the top 0.01\% token surprisals of its training set. Each point is one training set; lines connect the thresholds of each IPD direction, and shading marks averages below Vanilla. T-P sits above S-P in every setting, even where the two overlap in surprisal.}
\label{fig_tail_scatter}
\end{figure}

The external learning signal also separates the two IPD directions.
T-P outperforms S-P in every setting, even where their extreme-tail mean surprisals overlap (Figure~\ref{fig_tail_scatter}).
EXAONE shows this directly: one T-P dataset has higher extreme-tail mean surprisal than every S-P dataset, yet trains a better student than any of them.

In T-P, the teacher writes almost every token; in S-P, it writes only the rare fallbacks (Table~\ref{tab_quality_compatibility}).
S-P data thus carry the external signal only as corrections to the student's own reasoning, whereas T-P data carry the teacher's reasoning throughout.
\textbf{More teacher supervision strengthens transfer at comparable extreme-tail surprisal.}

\subsection{Token Replacements and Reasoning Quality}\label{sec_selective}

Student fallbacks steer the teacher toward plainer continuations, and the resulting traces, judged slightly weaker than s1K as training data, train a student that reasons better.
Table~\ref{tab_token_categories} classifies each rejected teacher proposal and its student fallback, from surface form to the discourse pivots that shift reasoning stages (Appendix~\ref{app_token_analysis}).
Fallbacks shift mass toward surface form and discourse connectives, and away from reasoning-content words and pivots.
Where the teacher proposes a pivot such as ``Wait'', the student often continues the argument with a connective such as ``Similarly'' or ``Hence''.

\begin{table}[H]
\centering
\begin{minipage}[t]{0.49\linewidth}
\vspace{0pt}
\caption{Token-category shares, T-P ($\tau=0.1\%$). Subscripts: relative change from rejected tokens.}
\label{tab_token_categories}
\centering\scriptsize\setlength{\tabcolsep}{3pt}
\begin{tabular}{lcc}
\toprule
Category & \begin{tabular}[c]{@{}c@{}}Rejected\\T. tokens (\%)\end{tabular} & \begin{tabular}[c]{@{}c@{}}Fallback\\S. tokens (\%)\end{tabular} \\
\midrule
Surface form & 24.59 & 34.16\textsubscript{\textcolor{ipdindigo}{$+$39\%}} \\
Computational content & 14.16 & 15.81\textsubscript{\textcolor{ipdindigo}{$+$12\%}} \\
Reasoning content & 57.30 & 44.45\textsubscript{\textcolor{ipdcoral}{$-$22\%}} \\
Discourse connective & \phantom{0}2.82 & \phantom{0}5.26\textsubscript{\textcolor{ipdindigo}{$+$87\%}} \\
Discourse pivot & \phantom{0}1.13 & \phantom{0}0.32\textsubscript{\textcolor{ipdcoral}{$-$72\%}} \\
\bottomrule
\end{tabular}
\end{minipage}\hfill
\begin{minipage}[t]{0.49\linewidth}
\vspace{0pt}
\caption{IPD win rate (\%) against s1K, mean $\pm$ sample standard deviation across three judges.}
\label{tab_trace_comparisons}
\centering\scriptsize\setlength{\tabcolsep}{3pt}
\begin{tabular}{lcc}
\toprule
Criterion & \begin{tabular}[c]{@{}c@{}}\textbf{Training traces}\\IPD data vs. s1K\end{tabular} & \begin{tabular}[c]{@{}c@{}}\textbf{Sampled traces}\\IPD- vs. s1K-trained\end{tabular} \\
\midrule
Problem groundedness & $54.4_{\pm 2.3}$ & $55.8_{\pm 2.2}$ \\
Step coherence & $51.9_{\pm 2.9}$ & $56.2_{\pm 2.1}$ \\
Logical validity & $50.3_{\pm 2.7}$ & $54.4_{\pm 2.9}$ \\
No redundancy & $41.1_{\pm 2.6}$ & $50.1_{\pm 1.9}$ \\
Overall preference & $47.7_{\pm 2.9}$ & $54.1_{\pm 2.3}$ \\
\bottomrule
\end{tabular}
\end{minipage}
\end{table}

Table~\ref{tab_trace_comparisons} reports blind judgments on both-correct pairs: IPD against s1K training data, and traces sampled from the students trained on each (Appendix~\ref{app_trace_quality}).
As training data, IPD traces are better grounded but less concise than s1K, and s1K is slightly preferred overall.
The IPD-trained student is then preferred overall and on groundedness, coherence, and validity, and is as concise.
\textbf{IPD trades a little trace quality for compatibility and trains a student that reasons better.}

\subsection{IPD Broadens Problem-Solving Coverage}\label{sec_coverage}
\begin{figure}[H]
\centering
\includegraphics[width=\linewidth]{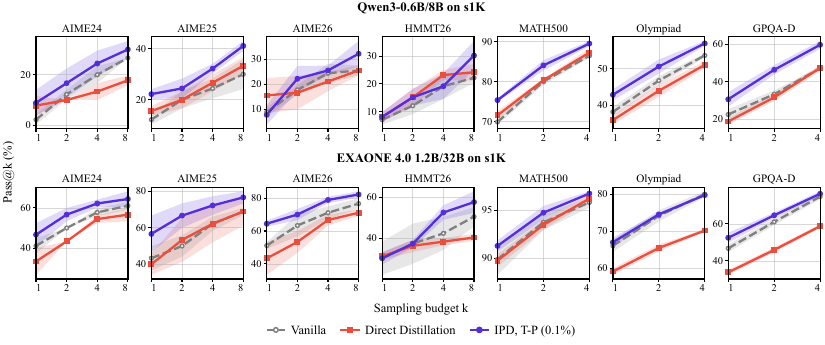}
\caption{Pass@$k$ under matched decoding on two s1K settings. Curves average three seeds; bands show one sample standard deviation. IPD leads both baselines at the largest budget in 13 of 14 panels.}
\label{fig_passk}
\end{figure}

IPD expands the range of problems the student can solve on its own.
Pass@$k$ measures whether any of $k$ sampled solutions is correct, which is what limits RL: it learns from a problem only when the student produces a correct rollout.
At the largest matched sampling budget, IPD leads Vanilla and Direct Distillation in 27 of 28 benchmark--setting pairs (Figure~\ref{fig_passk}; Tables~\ref{tab_passk_qwen_s1k}--\ref{tab_passk_qwen17_s1k}).
IPD therefore gives RL more successes to reinforce and OPD more useful prefixes for teacher feedback.
\textbf{Broader problem-solving coverage motivates IPD as an initialization for subsequent on-policy learning.}

% !TEX root = iclr2027_conference.tex
\section{Conclusion}\label{sec_conclusion}

IPD builds SFT targets in which the student's own tokens replace teacher proposals it would rarely produce, and the teacher continues from the revised prefix.
Across model families, datasets, and thresholds, IPD outperforms direct distillation and RFT.
Transfer failures trace to the few tokens the student finds least likely, while teacher-supplied tokens add value beyond compatibility.
Broader problem-solving coverage suggests IPD could initialize subsequent RL and OPD.
\textbf{Small models learn reasoning paths better when those paths arrive in a form they can make their own.}

\clearpage

\bibliography{iclr2027_conference}

@inproceedings{muennighoff2025s1,
    title = "s1: Simple test-time scaling",
    author = "Muennighoff, Niklas  and
      Yang, Zitong  and
      Shi, Weijia  and
      Li, Xiang Lisa  and
      Fei-Fei, Li  and
      Hajishirzi, Hannaneh  and
      Zettlemoyer, Luke  and
      Liang, Percy  and
      Cand{\`e}s, Emmanuel  and
      Hashimoto, Tatsunori",
    editor = "Christodoulopoulos, Christos  and
      Chakraborty, Tanmoy  and
      Rose, Carolyn  and
      Peng, Violet",
    booktitle = "Proceedings of the 2025 Conference on Empirical Methods in Natural Language Processing",
    month = nov,
    year = "2025",
    address = "Suzhou, China",
    publisher = "Association for Computational Linguistics",
    url = "https://aclanthology.org/2025.emnlp-main.1025/",
    doi = "10.18653/v1/2025.emnlp-main.1025",
    pages = "20275--20321",
    ISBN = "979-8-89176-332-6"
}

@inproceedings{
ye2025limo,
title={{LIMO}: Less is More for Reasoning},
author={Yixin Ye and Zhen Huang and Yang Xiao and Ethan Chern and Shijie Xia and Pengfei Liu},
booktitle={Second Conference on Language Modeling},
year={2025},
url={https://openreview.net/forum?id=T2TZ0RY4Zk}
}

@misc{xu2025phi4mini,
      title={{Phi-4-Mini-Reasoning}: Exploring the Limits of Small Reasoning Language Models in Math}, 
      author={Haoran Xu and Baolin Peng and Hany Awadalla and Dongdong Chen and Yen-Chun Chen and Mei Gao and Young Jin Kim and Yunsheng Li and Liliang Ren and Yelong Shen and Shuohang Wang and Weijian Xu and Jianfeng Gao and Weizhu Chen},
      year={2025},
      eprint={2504.21233},
      archivePrefix={arXiv},
      primaryClass={cs.CL},
      url={https://arxiv.org/abs/2504.21233}, 
}

@misc{yuan2023rft,
      title={Scaling Relationship on Learning Mathematical Reasoning with Large Language Models}, 
      author={Zheng Yuan and Hongyi Yuan and Chengpeng Li and Guanting Dong and Keming Lu and Chuanqi Tan and Chang Zhou and Jingren Zhou},
      year={2023},
      eprint={2308.01825},
      archivePrefix={arXiv},
      primaryClass={cs.CL},
      url={https://arxiv.org/abs/2308.01825}, 
}

@inproceedings{agarwal2024gkd,
 author = {Agarwal, Rishabh and Vieillard, Nino and Zhou, Yongchao and Stanczyk, Piotr and Ramos Garea, Sabela and Geist, Matthieu and Bachem, Olivier},
 booktitle = {International Conference on Learning Representations},
 editor = {B. Kim and Y. Yue and S. Chaudhuri and K. Fragkiadaki and M. Khan and Y. Sun},
 pages = {21246--21263},
 title = {On-Policy Distillation of Language Models: Learning from Self-Generated Mistakes},
 url = {https://proceedings.iclr.cc/paper_files/paper/2024/file/5be69a584901a26c521c2b51e40a4c20-Paper-Conference.pdf},
 volume = {2024},
 year = {2024}
}

@InProceedings{ko2025distillm2,
  title = 	 {{D}isti{LLM}-2: A Contrastive Approach Boosts the Distillation of {LLM}s},
  author =       {Ko, Jongwoo and Chen, Tianyi and Kim, Sungnyun and Ding, Tianyu and Liang, Luming and Zharkov, Ilya and Yun, Se-Young},
  booktitle = 	 {Proceedings of the 42nd International Conference on Machine Learning},
  pages = 	 {31044--31062},
  year = 	 {2025},
  editor = 	 {Singh, Aarti and Fazel, Maryam and Hsu, Daniel and Lacoste-Julien, Simon and Berkenkamp, Felix and Maharaj, Tegan and Wagstaff, Kiri and Zhu, Jerry},
  volume = 	 {267},
  series = 	 {Proceedings of Machine Learning Research},
  month = 	 {13--19 Jul},
  publisher =    {PMLR},
  url = 	 {https://proceedings.mlr.press/v267/ko25a.html}
}

@inproceedings{xu2025speculativekd,
 author = {Xu, Wenda and Han, Rujun and Wang, Zifeng and Le, Long T. and Madeka, Dhruv and Li, Lei and Wang, William Yang and Agarwal, Rishabh and Lee, Chen-Yu and Pfister, Tomas},
 booktitle = {International Conference on Learning Representations},
 editor = {Y. Yue and A. Garg and N. Peng and F. Sha and R. Yu},
 pages = {64616--64646},
 title = {Speculative Knowledge Distillation: Bridging the Teacher-Student Gap Through Interleaved Sampling},
 url = {https://proceedings.iclr.cc/paper_files/paper/2025/file/a2747a3844ca1e4667fbff3f558eb39b-Paper-Conference.pdf},
 volume = {2025},
 year = {2025}
}

@misc{zhu2026manyfaces,
      title={The Many Faces of On-Policy Distillation: Pitfalls, Mechanisms, and Fixes}, 
      author={Siqi Zhu and Xuyan Ye and Hongyu Lu and Weiye Shi and Ge Liu},
      year={2026},
      eprint={2605.11182},
      archivePrefix={arXiv},
      primaryClass={cs.AI},
      url={https://arxiv.org/abs/2605.11182}, 
}

@inproceedings{zhang2025grape,
 author = {Zhang, Dylan and Dai, Qirun and Peng, Hao},
 booktitle = {Advances in Neural Information Processing Systems},
 doi = {10.52202/085713-4717},
 editor = {D. Belgrave and C. Zhang and H. Lin and R. Pascanu and P. Koniusz and M. Ghassemi and N. Chen},
 pages = {141172--141208},
 publisher = {Curran Associates, Inc.},
 title = {The Best Instruction-Tuning Data are Those That Fit},
 url = {https://proceedings.neurips.cc/paper_files/paper/2025/file/cea04322465ad2f261f08e5b47ba9e7a-Paper-Conference.pdf},
 volume = {38, Main Conference},
 year = {2025}
}

@inproceedings{kim2024promptkd,
    title = "{P}rompt{KD}: Distilling Student-Friendly Knowledge for Generative Language Models via Prompt Tuning",
    author = "Kim, Gyeongman  and
      Jang, Doohyuk  and
      Yang, Eunho",
    editor = "Al-Onaizan, Yaser  and
      Bansal, Mohit  and
      Chen, Yun-Nung",
    booktitle = "Findings of the Association for Computational Linguistics: EMNLP 2024",
    month = nov,
    year = "2024",
    address = "Miami, Florida, USA",
    publisher = "Association for Computational Linguistics",
    url = "https://aclanthology.org/2024.findings-emnlp.364/",
    doi = "10.18653/v1/2024.findings-emnlp.364",
    pages = "6266--6282"
}

@inproceedings{just2025local,
  title={The Signal is in the Steps: Local Scoring for Reasoning Data Selection},
  author={Hoang Anh Just and Myeongseob Ko and Ruoxi Jia},
  booktitle={Forty-third International Conference on Machine Learning},
  year={2026},
  url={https://openreview.net/forum?id=GcB3a6IonG}
}

@inproceedings{yang2026rsr,
    title = "Which Reasoning Trajectories Teach Students to Reason Better? {A} Simple Metric of Informative Alignment",
    author = "Yang, Yuming  and
      Lai, Mingyoung  and
      Zhao, Wanxu  and
      Fan, Xiaoran  and
      Xi, Zhiheng  and
      Wu, Mingqi  and
      Huang, Chiyue  and
      Zhao, Jun  and
      Lv, Haijun  and
      Tong, Jian  and
      Zhou, Yunhua  and
      Zou, Yicheng  and
      Guo, Qipeng  and
      Gui, Tao  and
      Zhang, Qi  and
      Huang, Xuanjing",
    editor = "Liakata, Maria  and
      Moreira, Viviane P.  and
      Zhang, Jiajun  and
      Jurgens, David",
    booktitle = "Proceedings of the 64th Annual Meeting of the {A}ssociation for {C}omputational {L}inguistics (Volume 1: Long Papers)",
    month = jul,
    year = "2026",
    address = "San Diego, California, United States",
    publisher = "Association for Computational Linguistics",
    url = "https://aclanthology.org/2026.acl-long.1950/",
    doi = "10.18653/v1/2026.acl-long.1950",
    pages = "42123--42150",
    ISBN = "979-8-89176-390-6"
}

@inproceedings{liu2026palign,
    title = "Long-Chain Reasoning Distillation via Adaptive Prefix Alignment",
    author = "Liu, Zhenghao  and
      Wu, Zhuoyang  and
      Li, Xinze  and
      Yan, Yukun  and
      Wang, Shuo  and
      Chen, Zulong  and
      Gu, Yu  and
      Yu, Ge  and
      Sun, Maosong",
    editor = "Liakata, Maria  and
      Moreira, Viviane P.  and
      Zhang, Jiajun  and
      Jurgens, David",
    booktitle = "Proceedings of the 64th Annual Meeting of the {A}ssociation for {C}omputational {L}inguistics (Volume 1: Long Papers)",
    month = jul,
    year = "2026",
    address = "San Diego, California, United States",
    publisher = "Association for Computational Linguistics",
    url = "https://aclanthology.org/2026.acl-long.822/",
    doi = "10.18653/v1/2026.acl-long.822",
    pages = "18041--18059",
    ISBN = "979-8-89176-390-6"
}

@inproceedings{
yang2026sfd,
title={Making Complex Reasoning Student-Friendly: A Hybrid {LLM}-to-{SLM} Distillation Framework},
author={Yongjin Yang and Yinghui He and Jiarui Liu and Zhijing Jin},
booktitle={The 1st Workshop on Scaling Post-training for LLMs},
year={2026},
url={https://openreview.net/forum?id=6jP5PDOmqN}
}

@inproceedings{he2026pedagogical,
 author = {He, Bowei and Chen, Yankai and Zhang, Xiaokun and Kong, Linghe and Yu, Philip S. and Liu, Xue and Ma, Chen},
 booktitle = {International Conference on Learning Representations},
 editor = {C. Vondrick and B. Hariharan and C. Raffel and L. Pinto and D. Yang and A. Faust},
 pages = {112387--112424},
 title = {Pedagogically-Inspired Data Synthesis for Language Model Knowledge Distillation},
 url = {https://proceedings.iclr.cc/paper_files/paper/2026/file/b693a240cf1009bff9fa4422141c9392-Paper-Conference.pdf},
 volume = {2026},
 year = {2026}
}

@misc{he2026genssd,
      title={Student-in-the-Loop Chain-of-Thought Distillation via Generation-Time Selection}, 
      author={Chaoqun He and Yingfa Chen and Chaojun Xiao and Xu Han and Lijie Wen},
      year={2026},
      eprint={2604.02819},
      archivePrefix={arXiv},
      primaryClass={cs.CL},
      url={https://arxiv.org/abs/2604.02819}, 
}

@inproceedings{cetin2025rlt,
 author = {Cetin, Edoardo and Zhao, Tianyu and Tang, Yujin},
 booktitle = {Advances in Neural Information Processing Systems},
 doi = {10.52202/085713-3586},
 editor = {D. Belgrave and C. Zhang and H. Lin and R. Pascanu and P. Koniusz and M. Ghassemi and N. Chen},
 pages = {107533--107567},
 publisher = {Curran Associates, Inc.},
 title = {Reinforcement Learning Teachers of Test Time Scaling},
 url = {https://proceedings.neurips.cc/paper_files/paper/2025/file/9a6b278218966499194491f55ccf8b75-Paper-Conference.pdf},
 volume = {38, Main Conference},
 year = {2025}
}

@inproceedings{li2025montessori,
 author = {Li, Xiaochuan and Yu, Zichun and Xiong, Chenyan},
 booktitle = {International Conference on Learning Representations},
 editor = {Y. Yue and A. Garg and N. Peng and F. Sha and R. Yu},
 pages = {74809--74834},
 title = {{Montessori-Instruct}: Generate Influential Training Data Tailored for Student Learning},
 url = {https://proceedings.iclr.cc/paper_files/paper/2025/file/ba1d33849b963efc6b5d3082ad68f480-Paper-Conference.pdf},
 volume = {2025},
 year = {2025}
}

@misc{qwen2025qwen3,
      title={Qwen3 Technical Report}, 
      author={An Yang and Anfeng Li and Baosong Yang and Beichen Zhang and Binyuan Hui and Bo Zheng and Bowen Yu and Chang Gao and Chengen Huang and Chenxu Lv and Chujie Zheng and Dayiheng Liu and Fan Zhou and Fei Huang and Feng Hu and Hao Ge and Haoran Wei and Huan Lin and Jialong Tang and Jian Yang and Jianhong Tu and Jianwei Zhang and Jianxin Yang and Jiaxi Yang and Jing Zhou and Jingren Zhou and Junyang Lin and Kai Dang and Keqin Bao and Kexin Yang and Le Yu and Lianghao Deng and Mei Li and Mingfeng Xue and Mingze Li and Pei Zhang and Peng Wang and Qin Zhu and Rui Men and Ruize Gao and Shixuan Liu and Shuang Luo and Tianhao Li and Tianyi Tang and Wenbiao Yin and Xingzhang Ren and Xinyu Wang and Xinyu Zhang and Xuancheng Ren and Yang Fan and Yang Su and Yichang Zhang and Yinger Zhang and Yu Wan and Yuqiong Liu and Zekun Wang and Zeyu Cui and Zhenru Zhang and Zhipeng Zhou and Zihan Qiu},
      year={2025},
      eprint={2505.09388},
      archivePrefix={arXiv},
      primaryClass={cs.CL},
      url={https://arxiv.org/abs/2505.09388}, 
}

@misc{lg2025exaone,
      title={{EXAONE} 4.0: Unified Large Language Models Integrating Non-reasoning and Reasoning Modes}, 
      author={Kyunghoon Bae and Eunbi Choi and Kibong Choi and Stanley Jungkyu Choi and Yemuk Choi and Kyubeen Han and Seokhee Hong and Junwon Hwang and Taewan Hwang and Joonwon Jang and Hyojin Jeon and Kijeong Jeon and Gerrard Jeongwon Jo and Hyunjik Jo and Jiyeon Jung and Euisoon Kim and Hyosang Kim and Jihoon Kim and Joonkee Kim and Seonghwan Kim and Soyeon Kim and Sunkyoung Kim and Yireun Kim and Yongil Kim and Youchul Kim and Edward Hwayoung Lee and Gwangho Lee and Haeju Lee and Honglak Lee and Jinsik Lee and Kyungmin Lee and Sangha Park and Young Min Paik and Yongmin Park and Youngyong Park and Sanghyun Seo and Sihoon Yang and Heuiyeen Yeen and Sihyuk Yi and Hyeongu Yun},
      year={2025},
      eprint={2507.11407},
      archivePrefix={arXiv},
      primaryClass={cs.CL},
      url={https://arxiv.org/abs/2507.11407}, 
}

@inproceedings{
rein2024gpqa,
title={{GPQA}: A Graduate-Level {Google}-Proof {Q\&A} Benchmark},
author={David Rein and Betty Li Hou and Asa Cooper Stickland and Jackson Petty and Richard Yuanzhe Pang and Julien Dirani and Julian Michael and Samuel R. Bowman},
booktitle={First Conference on Language Modeling},
year={2024},
url={https://openreview.net/forum?id=Ti67584b98}
}

@inproceedings{he2024olympiadbench,
    title = "{O}lympiad{B}ench: A Challenging Benchmark for Promoting {AGI} with Olympiad-Level Bilingual Multimodal Scientific Problems",
    author = "He, Chaoqun  and
      Luo, Renjie  and
      Bai, Yuzhuo  and
      Hu, Shengding  and
      Thai, Zhen Leng  and
      Shen, Junhao  and
      Hu, Jinyi  and
      Han, Xu  and
      Huang, Yujie  and
      Zhang, Yuxiang  and
      Liu, Jie  and
      Qi, Lei  and
      Liu, Zhiyuan  and
      Sun, Maosong",
    editor = "Ku, Lun-Wei  and
      Martins, Andre  and
      Srikumar, Vivek",
    booktitle = "Proceedings of the 62nd Annual Meeting of the Association for Computational Linguistics (Volume 1: Long Papers)",
    month = aug,
    year = "2024",
    address = "Bangkok, Thailand",
    publisher = "Association for Computational Linguistics",
    url = "https://aclanthology.org/2024.acl-long.211/",
    doi = "10.18653/v1/2024.acl-long.211",
    pages = "3828--3850"
}

@inproceedings{wang2025beyond,
 author = {Wang, Shenzhi and Yu, Le and Gao, Chang and Zheng, Chujie and Liu, Shixuan and Lu, Rui and Dang, Kai and Chen, Xiong-Hui and Yang, Jianxin and Zhang, Zhenru and Liu, Yuqiong and Yang, An and Zhao, Andrew and Yue, Yang and Song, Shiji and Yu, Bowen and Huang, Gao and Lin, Junyang},
 booktitle = {Advances in Neural Information Processing Systems},
 doi = {10.52202/085713-3850},
 editor = {D. Belgrave and C. Zhang and H. Lin and R. Pascanu and P. Koniusz and M. Ghassemi and N. Chen},
 pages = {115452--115486},
 publisher = {Curran Associates, Inc.},
 title = {Beyond the 80/20 Rule: High-Entropy Minority Tokens Drive Effective Reinforcement Learning for {LLM} Reasoning},
 url = {https://proceedings.neurips.cc/paper_files/paper/2025/file/a797c2d2e0c1fdabf4d1ab8cd0b465c6-Paper-Conference.pdf},
 volume = {38, Main Conference},
 year = {2025}
}

@inproceedings{lippmann2025stylesubstancedistilledlanguage,
  title={Style over Substance: Distilled Language Models Reason Via Stylistic Replication},
  author={Philip Lippmann and Jie Yang},
  booktitle={Second Conference on Language Modeling},
  year={2025},
  url={https://openreview.net/forum?id=5wAfbEs34A}
}

@inproceedings{zhang2025breadbranchedrolloutsexpert,
 author = {Zhang, Xuechen and Huang, Zijian and Li, Yingcong and Ni, Chenshun and Chen, Jiasi and Oymak, Samet},
 booktitle = {Advances in Neural Information Processing Systems},
 doi = {10.52202/085713-3235},
 editor = {D. Belgrave and C. Zhang and H. Lin and R. Pascanu and P. Koniusz and M. Ghassemi and N. Chen},
 pages = {96726--96752},
 publisher = {Curran Associates, Inc.},
 title = {{BREAD}: Branched Rollouts from Expert Anchors Bridge {SFT} \& {RL} for Reasoning},
 url = {https://proceedings.neurips.cc/paper_files/paper/2025/file/8b69f0fac9b83f5ba43c7a851a841a61-Paper-Conference.pdf},
 volume = {38, Main Conference},
 year = {2025}
}

@inproceedings{kim2026explainwordsimprovingreasoning,
 author = {Kim, Minsang and Baek, Seung Jun},
 booktitle = {International Conference on Learning Representations},
 editor = {C. Vondrick and B. Hariharan and C. Raffel and L. Pinto and D. Yang and A. Faust},
 pages = {103174--103196},
 title = {Explain in Your Own Words: Improving Reasoning via Token-Selective Dual Knowledge Distillation},
 url = {https://proceedings.iclr.cc/paper_files/paper/2026/file/a89331e2ecbbd9d9618bc94717a91155-Paper-Conference.pdf},
 volume = {2026},
 year = {2026}
}

@inproceedings{ji2025upft,
 author = {Ji, Ke and Xu, Jiahao and Liang, Tian and Liu, Qiuzhi and He, Zhiwei and Chen, Xingyu and Liu, Xiaoyuan and Wang, Zhijie and Chen, Junying and Wang, Benyou and Tu, Zhaopeng and Mi, Haitao and Yu, Dong},
 booktitle = {Advances in Neural Information Processing Systems},
 doi = {10.52202/085713-0314},
 editor = {D. Belgrave and C. Zhang and H. Lin and R. Pascanu and P. Koniusz and M. Ghassemi and N. Chen},
 pages = {9305--9325},
 publisher = {Curran Associates, Inc.},
 title = {The First Few Tokens Are All You Need: An Efficient and Effective Unsupervised Prefix Fine-Tuning Method for Reasoning Models},
 url = {https://proceedings.neurips.cc/paper_files/paper/2025/file/0d6da26625f780df1572cf7c8dc3faa6-Paper-Conference.pdf},
 volume = {38, Main Conference},
 year = {2025}
}

@inproceedings{panigrahi2026grace,
 author = {Panigrahi, Abhishek and Liu, Bingbin and Malladi, Sadhika and Kakade, Sham and Goel, Surbhi},
 booktitle = {International Conference on Learning Representations},
 editor = {C. Vondrick and B. Hariharan and C. Raffel and L. Pinto and D. Yang and A. Faust},
 pages = {54779--54822},
 title = {In Good {GRACES}: Principled Teacher Selection for Knowledge Distillation},
 url = {https://proceedings.iclr.cc/paper_files/paper/2026/file/59fa5d049b777d880fdcddba2b24738f-Paper-Conference.pdf},
 volume = {2026},
 year = {2026}
}

@inproceedings{liu2025provenance,
 author = {Liu, Kaiyuan and Yan, Shaotian and Miao, Rui and Wang, Bing and Shen, Chen and Zhang, Jun and Ye, Jieping},
 booktitle = {International Conference on Learning Representations},
 editor = {C. Vondrick and B. Hariharan and C. Raffel and L. Pinto and D. Yang and A. Faust},
 pages = {82724--82743},
 title = {Where Did This Sentence Come From? {T}racing Provenance in {LLM} Reasoning Distillation},
 url = {https://proceedings.iclr.cc/paper_files/paper/2026/file/85a67c093f26d9d6f2a197aa24b50414-Paper-Conference.pdf},
 volume = {2026},
 year = {2026}
}

@inproceedings{wen2025scaffolding,
 author = {Wen, Xiangyu and Huang, Junhua and Li, Zeju and Li, Min and Zhong, Jianyuan and Xu, Zhijian and Yuan, Mingxuan and Huang, Yongxiang and Xu, Qiang},
 booktitle = {International Conference on Learning Representations},
 editor = {C. Vondrick and B. Hariharan and C. Raffel and L. Pinto and D. Yang and A. Faust},
 pages = {80433--80459},
 title = {Reasoning Scaffolding: Distilling the Flow of Thought from {LLMs}},
 url = {https://proceedings.iclr.cc/paper_files/paper/2026/file/8248b1ded388fcdbbd121bcdfea3068c-Paper-Conference.pdf},
 volume = {2026},
 year = {2026}
}

@misc{shao2024deepseekmath,
      title={{DeepSeekMath}: Pushing the Limits of Mathematical Reasoning in Open Language Models}, 
      author={Zhihong Shao and Peiyi Wang and Qihao Zhu and Runxin Xu and Junxiao Song and Xiao Bi and Haowei Zhang and Mingchuan Zhang and Y. K. Li and Y. Wu and Daya Guo},
      year={2024},
      eprint={2402.03300},
      archivePrefix={arXiv},
      primaryClass={cs.CL},
      url={https://arxiv.org/abs/2402.03300}, 
}

@misc{maa2025aime,
  title={{MAA Invitational Competitions}},
  author={{Mathematical Association of America}},
  year={n.d.},
  note={American Invitational Mathematics Examination (AIME). Accessed 23 September 2026},
  url={https://maa.org/maa-invitational-competitions/}
}

@misc{hmmt2026,
  title={{Harvard--MIT Mathematics Tournament}},
  author={{Harvard--MIT Mathematics Tournament}},
  year={2026},
  note={Competition problem set; we use the February 2026 edition},
  url={https://www.hmmt.org/www/archive/292}
}

@inproceedings{prasad2008pdtb,
    title = "The {P}enn {D}iscourse {T}ree{B}ank 2.0.",
    author = "Prasad, Rashmi  and
      Dinesh, Nikhil  and
      Lee, Alan  and
      Miltsakaki, Eleni  and
      Robaldo, Livio  and
      Joshi, Aravind  and
      Webber, Bonnie",
    editor = "Calzolari, Nicoletta  and
      Choukri, Khalid  and
      Maegaard, Bente  and
      Mariani, Joseph  and
      Odijk, Jan  and
      Piperidis, Stelios  and
      Tapias, Daniel",
    booktitle = "Proceedings of the Sixth International Conference on Language Resources and Evaluation ({LREC}'08)",
    month = may,
    year = "2008",
    address = "Marrakech, Morocco",
    publisher = "European Language Resources Association (ELRA)",
    url = "https://aclanthology.org/L08-1093/"
}

@inproceedings{zheng2023judging,
 author = {Zheng, Lianmin and Chiang, Wei-Lin and Sheng, Ying and Zhuang, Siyuan and Wu, Zhanghao and Zhuang, Yonghao and Lin, Zi and Li, Zhuohan and Li, Dacheng and Xing, Eric P. and Zhang, Hao and Gonzalez, Joseph E. and Stoica, Ion},
 booktitle = {Advances in Neural Information Processing Systems},
 doi = {10.52202/075280-2020},
 editor = {A. Oh and T. Naumann and A. Globerson and K. Saenko and M. Hardt and S. Levine},
 pages = {46595--46623},
 publisher = {Curran Associates, Inc.},
 title = {Judging {LLM}-as-a-Judge with {MT-Bench} and {Chatbot Arena}},
 url = {https://proceedings.neurips.cc/paper_files/paper/2023/file/91f18a1287b398d378ef22505bf41832-Paper-Datasets_and_Benchmarks.pdf},
 volume = {36},
 year = {2023}
}

@inproceedings{lightman2024verify,
 author = {Lightman, Hunter and Kosaraju, Vineet and Burda, Yura and Edwards, Harri and Baker, Bowen and Lee, Teddy and Leike, Jan and Schulman, John  and Sutskever, Ilya and Cobbe, Karl},
 booktitle = {International Conference on Learning Representations},
 editor = {B. Kim and Y. Yue and S. Chaudhuri and K. Fragkiadaki and M. Khan and Y. Sun},
 pages = {39578--39601},
 title = {Let\textquotesingle s Verify Step by Step},
 url = {https://proceedings.iclr.cc/paper_files/paper/2024/file/aca97732e30bcf1303bc22ac3924fd16-Paper-Conference.pdf},
 volume = {2024},
 year = {2024}
}

@article{xia2025reasoneval,
  title={Evaluating mathematical reasoning beyond accuracy},
  author={Xia, Shijie and Li, Xuefeng and Liu, Yixin and Wu, Tongshuang and Liu, Pengfei},
  journal={Proceedings of the AAAI Conference on Artificial Intelligence},
  volume={39},
  number={26},
  pages={27723--27730},
  year={2025},
  doi={10.1609/aaai.v39i26.34987},
  url={https://ojs.aaai.org/index.php/AAAI/article/view/34987}
}

@inproceedings{
golovneva2023roscoe,
title={{ROSCOE}: A Suite of Metrics for Scoring Step-by-Step Reasoning},
author={Olga Golovneva and Moya Chen and Spencer Poff and Martin Corredor and Luke Zettlemoyer and Maryam Fazel-Zarandi and Asli Celikyilmaz},
booktitle={The Eleventh International Conference on Learning Representations },
year={2023},
url={https://openreview.net/forum?id=xYlJRpzZtsY}
}

@inproceedings{jacovi2024reveal,
    title = "A Chain-of-Thought Is as Strong as Its Weakest Link: A Benchmark for Verifiers of Reasoning Chains",
    author = "Jacovi, Alon  and
      Bitton, Yonatan  and
      Bohnet, Bernd  and
      Herzig, Jonathan  and
      Honovich, Or  and
      Tseng, Michael  and
      Collins, Michael  and
      Aharoni, Roee  and
      Geva, Mor",
    editor = "Ku, Lun-Wei  and
      Martins, Andre  and
      Srikumar, Vivek",
    booktitle = "Proceedings of the 62nd Annual Meeting of the Association for Computational Linguistics (Volume 1: Long Papers)",
    month = aug,
    year = "2024",
    address = "Bangkok, Thailand",
    publisher = "Association for Computational Linguistics",
    url = "https://aclanthology.org/2024.acl-long.254/",
    doi = "10.18653/v1/2024.acl-long.254",
    pages = "4615--4634"
}

@inproceedings{lee2025evaluating,
    title = "Evaluating Step-by-step Reasoning Traces: A Survey",
    author = "Lee, Jinu  and
      Hockenmaier, Julia",
    editor = "Christodoulopoulos, Christos  and
      Chakraborty, Tanmoy  and
      Rose, Carolyn  and
      Peng, Violet",
    booktitle = "Findings of the Association for Computational Linguistics: EMNLP 2025",
    month = nov,
    year = "2025",
    address = "Suzhou, China",
    publisher = "Association for Computational Linguistics",
    url = "https://aclanthology.org/2025.findings-emnlp.94/",
    doi = "10.18653/v1/2025.findings-emnlp.94",
    pages = "1789--1814",
    ISBN = "979-8-89176-335-7"
}

@misc{ye2026coach,
      title={{LLM}-as-a-Coach: Experiential Learning for Non-Verifiable Tasks}, 
      author={Tianzhu Ye and Li Dong and Guanheng Chen and He Zhu and Xun Wu and Shaohan Huang and Furu Wei},
      year={2026},
      eprint={2607.18110},
      archivePrefix={arXiv},
      primaryClass={cs.LG},
      url={https://arxiv.org/abs/2607.18110}, 
}
\bibliographystyle{iclr2027_conference}

\clearpage
\appendix
% !TEX root = iclr2027_conference.tex
\section{Experimental Setup}\label{app_experimental_details}

\subsection{Training-Data Construction}\label{app_rollout}

Table~\ref{tab_generation_controls} summarizes the generation controls for the
1,000-row s1K and 800-row LIMO sources. Student RFT, Teacher RFT, and IPD use an
eight-attempt budget per question. The correctness verifier is an LLM judge,
\texttt{gpt-5.4-nano-2026-03-17}. Each unsolved IPD question contributes a partial trace
of at most 128 tokens from its latest failed attempt.

Direct Distillation preserves all released source traces. Our verifier accepts
479 of the 1,000 s1K-1.1 traces as correct; the s1 paper reports 630
(63.0\%)~\citep{muennighoff2025s1}. LIMO-v2 provides reasoning-only traces,
with a separate answer used for judging.

\begin{table}[H]
  \caption{Dataset-generation controls for verified full traces and partial traces. Each IPD dataset covers all source questions with targets generated by the same IPD condition.}
  \label{tab_generation_controls}
  \vspace{3pt}
  \centering
  \footnotesize
  \begin{tabular}{ll}
    \toprule
    \textbf{Setting} & \textbf{Value} \\
    \midrule
    Prompt-token budget & 4{,}096 \\
    Generated-token budget & 16{,}384 \\
    Temperature / top-$p$ & 0.6 / 0.95 \\
    Full-trace attempts & At most 8 \\
    Partial-trace length & At most 128 tokens \\
    Unsolved row & Condition-native partial trace \\
    Generation close & No forced reasoning close \\
    Source coverage & 1{,}000 rows for s1K; 800 rows for LIMO \\
    \bottomrule
  \end{tabular}
\end{table}

\subsection{Supervised Fine-Tuning}

Table~\ref{tab_training_controls} lists the shared SFT controls. All headline
comparisons use full-parameter BF16 SFT with 20,000-token packing blocks,
prompt-loss masking, a microbatch of one, gradient accumulation of 16, and
AdamW with a cosine schedule. Verified generated full traces receive the student's native
end marker, whereas partial traces end abruptly. LIMO Direct Distillation
is the reasoning-only exception described above. Within a headline
setting, every method uses the same selected schedule.

\begin{table}[H]
  \caption{SFT controls for Qwen3 and EXAONE. All SFT methods within a setting share the same optimization schedule and prompt loss masking.}
  \label{tab_training_controls}
  \vspace{3pt}
  \centering
  \footnotesize
  \setlength{\tabcolsep}{3pt}
  \begin{tabular}{ll}
    \toprule
    \textbf{Setting} & \textbf{Value} \\
    \midrule
    Training type / precision & Full-parameter SFT / BF16 \\
    Packing block & 20{,}000 tokens \\
    Microbatch / accumulation / effective batch & 1 / 16 / 16 \\
    Optimizer / schedule & AdamW / cosine \\
    Loss & Prompt-masked next-token likelihood \\
    Qwen3 headline schedule & 15 epochs, $3\times10^{-5}$ \\
    EXAONE headline schedule & 30 epochs, $1\times10^{-5}$ \\
    Complete trace end & Native EOS or turn-end marker \\
    LIMO Direct Distillation end & Reasoning close, then no answer or EOS \\
    Partial-trace end & Abrupt end with no marker \\
    \bottomrule
  \end{tabular}
\end{table}

\subsection{Held-Out Hyperparameter Search}\label{app_hyperparameters}

We conduct a separate SFT hyperparameter search for each of the four headline settings.
In each search, we reserve a deterministic 10\% of the dataset and train on the
remaining 90\%. The search uses Teacher RFT as the representative off-policy
condition and T-P ($\tau=0.1\%$) as the representative interleaved-policy
condition. These are the highest-performing conditions in their respective
families across multiple headline settings. We therefore use them to select one
shared schedule per model--dataset setting rather than tuning every method
separately.

The grid crosses 5, 15, and 30 epochs with learning rates
$1\times10^{-5}$, $2\times10^{-5}$, and $3\times10^{-5}$. We select the
schedule that ranks near the top for both representative conditions on both
held-out Avg@8 and held-out Pass@8, rather than the argmax of either condition
alone. In the two grids shown, the representative conditions favor the same region of the
grid, the higher learning rates for Qwen3 and the lower ones for EXAONE, so the
shared schedule sits near the optimum of each rather than compromising between
conflicting optima. For Qwen3-0.6B/8B the selected cell is second on Avg@8 for
both representative conditions, and on Pass@8 it is second for IPD and joint
third for Teacher RFT; for EXAONE 4.0 1.2B/32B it is first on Avg@8 and second
on Pass@8 for both.

Tables~\ref{tab_holdout_qwen} and~\ref{tab_holdout_exaone} present representative
search grids for Qwen3-0.6B/8B on s1K and EXAONE on s1K.
The Qwen3-0.6B/LIMO and Qwen3-1.7B/s1K searches independently select the same
15-epoch, $3\times10^{-5}$ schedule as Qwen3-0.6B/s1K.
Table~\ref{tab_hyperparameter_grid} summarizes the selected schedules for all four settings.

\begin{table}[H]
  \caption{Held-out schedule search for Qwen3-0.6B/8B on s1K using a deterministic 90/10 split. Shading marks the schedule selected for all SFT methods within this setting.}
  \label{tab_holdout_qwen}
  \vspace{3pt}
  \centering
  \footnotesize
  \setlength{\tabcolsep}{4pt}
  \begin{tabular}{llrrrrrr}
    \toprule
    \textbf{Method} & \textbf{LR} & \textbf{Epochs} & \textbf{Avg@8} & \textbf{Pass@1} & \textbf{Pass@2} & \textbf{Pass@4} & \textbf{Pass@8} \\
    \midrule
    Teacher RFT & $1\times10^{-5}$ & 5  & 11.00 & 10.00 & 16.00 & 25.00 & 32.00 \\
    Teacher RFT & $1\times10^{-5}$ & 15 & 12.88 & 10.00 & 17.00 & 26.00 & 37.00 \\
    Teacher RFT & $1\times10^{-5}$ & 30 & 11.62 &  8.00 & 19.00 & 27.00 & 37.00 \\
    Teacher RFT & $2\times10^{-5}$ & 5  & 11.75 & 14.00 & 23.00 & 25.00 & 39.00 \\
    Teacher RFT & $2\times10^{-5}$ & 15 & 13.12 & 12.00 & 18.00 & 24.00 & 34.00 \\
    Teacher RFT & $2\times10^{-5}$ & 30 & 15.00 & 14.00 & 23.00 & 38.00 & 40.00 \\
    Teacher RFT & $3\times10^{-5}$ & 5  & 13.88 & 15.00 & 22.00 & 30.00 & 41.00 \\
    \rowcolor{gray!15} Teacher RFT & $3\times10^{-5}$ & 15 & 15.25 & 11.00 & 23.00 & 32.00 & 39.00 \\
    Teacher RFT & $3\times10^{-5}$ & 30 & 15.88 & 15.00 & 23.00 & 34.00 & 38.00 \\
    \midrule
    IPD, T-P (0.1\%) & $1\times10^{-5}$ & 5  & 14.50 & 13.00 & 19.00 & 30.00 & 39.00 \\
    IPD, T-P (0.1\%) & $1\times10^{-5}$ & 15 & 13.63 & 12.00 & 21.00 & 32.00 & 42.00 \\
    IPD, T-P (0.1\%) & $1\times10^{-5}$ & 30 & 15.00 & 16.00 & 22.00 & 35.00 & 41.00 \\
    IPD, T-P (0.1\%) & $2\times10^{-5}$ & 5  & 15.25 & 13.00 & 24.00 & 32.00 & 40.00 \\
    IPD, T-P (0.1\%) & $2\times10^{-5}$ & 15 & 14.50 & 15.00 & 25.00 & 33.00 & 41.00 \\
    IPD, T-P (0.1\%) & $2\times10^{-5}$ & 30 & 17.88 & 13.00 & 25.00 & 35.00 & 47.00 \\
    IPD, T-P (0.1\%) & $3\times10^{-5}$ & 5  & 15.62 & 14.00 & 21.00 & 30.00 & 41.00 \\
    \rowcolor{gray!15} IPD, T-P (0.1\%) & $3\times10^{-5}$ & 15 & 17.50 & 21.00 & 32.00 & 39.00 & 46.00 \\
    IPD, T-P (0.1\%) & $3\times10^{-5}$ & 30 & 14.75 & 17.00 & 24.00 & 32.00 & 42.00 \\
    \bottomrule
  \end{tabular}
\end{table}

\begin{table}[H]
  \caption{Held-out schedule search for EXAONE 4.0 1.2B/32B on s1K using a deterministic 90/10 split. Shading marks the schedule selected for all SFT methods within this setting.}
  \label{tab_holdout_exaone}
  \vspace{3pt}
  \centering
  \footnotesize
  \setlength{\tabcolsep}{4pt}
  \begin{tabular}{llrrrrrr}
    \toprule
    \textbf{Method} & \textbf{LR} & \textbf{Epochs} & \textbf{Avg@8} & \textbf{Pass@1} & \textbf{Pass@2} & \textbf{Pass@4} & \textbf{Pass@8} \\
    \midrule
    Teacher RFT & $1\times10^{-5}$ & 5  & 43.12 & 40.00 & 55.00 & 60.00 & 66.00 \\
    Teacher RFT & $1\times10^{-5}$ & 15 & 43.00 & 43.00 & 53.00 & 61.00 & 61.00 \\
    \rowcolor{gray!15} Teacher RFT & $1\times10^{-5}$ & 30 & 43.62 & 47.00 & 54.00 & 59.00 & 66.00 \\
    Teacher RFT & $2\times10^{-5}$ & 5  & 41.62 & 40.00 & 52.00 & 63.00 & 67.00 \\
    Teacher RFT & $2\times10^{-5}$ & 15 & 41.00 & 41.00 & 54.00 & 59.00 & 65.00 \\
    Teacher RFT & $2\times10^{-5}$ & 30 & 40.38 & 45.00 & 50.00 & 56.00 & 61.00 \\
    Teacher RFT & $3\times10^{-5}$ & 5  & 42.50 & 48.00 & 54.00 & 61.00 & 65.00 \\
    Teacher RFT & $3\times10^{-5}$ & 15 & 41.00 & 38.00 & 46.00 & 59.00 & 61.00 \\
    Teacher RFT & $3\times10^{-5}$ & 30 & 37.88 & 40.00 & 52.00 & 57.00 & 61.00 \\
    \midrule
    IPD, T-P (0.1\%) & $1\times10^{-5}$ & 5  & 41.12 & 43.00 & 52.00 & 60.00 & 63.00 \\
    IPD, T-P (0.1\%) & $1\times10^{-5}$ & 15 & 41.88 & 46.00 & 53.00 & 57.00 & 61.00 \\
    \rowcolor{gray!15} IPD, T-P (0.1\%) & $1\times10^{-5}$ & 30 & 44.75 & 46.00 & 55.00 & 60.00 & 65.00 \\
    IPD, T-P (0.1\%) & $2\times10^{-5}$ & 5  & 44.25 & 49.00 & 54.00 & 62.00 & 67.00 \\
    IPD, T-P (0.1\%) & $2\times10^{-5}$ & 15 & 43.50 & 45.00 & 53.00 & 59.00 & 63.00 \\
    IPD, T-P (0.1\%) & $2\times10^{-5}$ & 30 & 42.88 & 37.00 & 50.00 & 60.00 & 63.00 \\
    IPD, T-P (0.1\%) & $3\times10^{-5}$ & 5  & 42.00 & 41.00 & 52.00 & 58.00 & 64.00 \\
    IPD, T-P (0.1\%) & $3\times10^{-5}$ & 15 & 39.88 & 41.00 & 50.00 & 55.00 & 60.00 \\
    IPD, T-P (0.1\%) & $3\times10^{-5}$ & 30 & 39.25 & 36.00 & 46.00 & 55.00 & 59.00 \\
    \bottomrule
  \end{tabular}
\end{table}

\begin{table}[H]
  \caption{SFT schedules selected by independent held-out searches in all four settings. Within each setting, all SFT methods use the selected schedule.}
  \label{tab_hyperparameter_grid}
  \vspace{3pt}
  \centering
  \footnotesize
  \setlength{\tabcolsep}{4pt}
  \begin{tabular}{lcc}
    \toprule
    \textbf{Headline setting} & \textbf{Epochs} & \textbf{Learning rate} \\
    \midrule
    Qwen3-0.6B/8B on s1K & 15 & $3\times10^{-5}$ \\
    Qwen3-0.6B/8B on LIMO & 15 & $3\times10^{-5}$ \\
    Qwen3-1.7B/8B on s1K & 15 & $3\times10^{-5}$ \\
    EXAONE 4.0 1.2B/32B on s1K & 30 & $1\times10^{-5}$ \\
    \bottomrule
  \end{tabular}
\end{table}

\subsection{Evaluation and Statistical Reporting}

Evaluation permits 32,768 generated tokens, reserving the last 2,048 tokens for
the answer after at most 30,720 reasoning tokens. Sampling uses temperature
0.6 and top-$p$ 0.95. Attempt $k$ uses the evaluation base seed plus $k$.

For each checkpoint, Avg@$k$ averages accuracy over $k$ sampled attempts.
For each fine-tuned condition, seeds 42, 1, and 2026 index three independent
SFT training runs. Table~\ref{tab_family_results} reports the mean and sample
standard deviation of their benchmark scores. These training seeds are distinct
from evaluation sampling seeds. Pass@$k$ reports cumulative question-level
coverage at the available sampling budgets without a per-seed standard deviation.

\clearpage

\section{Baselines and Ablation Controls}\label{app_baselines}

\subsection{On-Policy Baselines and Budget Accounting}\label{app_on_policy}

Both baselines initialize from the released Qwen3-0.6B checkpoint and train on
the s1K questions with a learning rate of $10^{-6}$. Generation uses a 4,096-token
prompt allowance, a 16,384-token output limit, temperature 0.6, and top-$p$ 0.95.
Each update uses 16 questions, with fresh rollouts sampled from the current student.

GRPO~\citep{shao2024deepseekmath} samples eight rollouts per question and uses
binary answer-correctness rewards to compute group-relative advantages. It
performs one optimization pass per rollout batch and uses a KL penalty of 0.01
against the frozen initial student.

OPD~\citep{agarwal2024gkd} samples one rollout per question and minimizes
token-level reverse KL against the frozen Qwen3-8B teacher. The objective uses
the teacher's top-20 tokens with a stop-gradient normalizer. All sampled rollouts
contribute to the loss regardless of correctness, including incorrect full rollouts.

In the primary setting, RFT and IPD use 15 SFT epochs on 1,000 questions with
an effective batch size of 16. This gives $1{,}000\times15/16=937.5$ batches;
dropping the final half-batch yields the 937 optimizer updates reported in
Table~\ref{tab_compute_budget}.

IPD permits eight attempts per question, giving at most
$8\times1{,}000=8{,}000$ rollouts. GRPO generates
eight rollouts per question for 16 questions at each of 1,000 steps, totaling
$8\times16\times1{,}000=128{,}000$. OPD generates one rollout per question for
16 questions at each of 500 steps, totaling $16\times500=8{,}000$.
We weight maximum rollout counts by the sizes of the models used for generation
and teacher scoring. The parameter-weighted rollout budgets are therefore
$(0.6+8)\times8{,}000=68{,}800$ for IPD and OPD and
$0.6\times128{,}000=76{,}800$ for GRPO, in billions of parameters times rollouts.
Normalizing by IPD gives $1.00\times$ for OPD and $1.12\times$ for GRPO.
The same weighting gives $0.6\times8{,}000=4{,}800$ for Student RFT-8,
$0.6\times100{,}000=60{,}000$ for Student RFT-100, and $8\times8{,}000=64{,}000$
for Teacher RFT, or $0.07\times$, $0.87\times$, and $0.93\times$.

\subsection{Question-Matched and Correctness-Based Controls}\label{app_data_controls}

Question-matched controls retain original s1K traces for the 516 questions solved
by T-P at $\tau=0.1\%$. The full-only variant uses these 516 traces alone;
the full-plus-partial variant adds 128-token partial traces for the remaining
484 questions. Correctness-based controls instead retain the 479 source traces
accepted by our verifier, either alone or with 128-token partial traces from the
521 rejected traces. All Direct Distillation partial traces come from the corresponding
original s1K traces.

We additionally evaluate IPD using only its 516 verified full traces. Comparing
full-only and full-plus-partial variants measures the contribution of partial
traces, while the question-matched comparison holds the selected questions and
full-trace counts fixed. In both controls, adding partial traces does not close the gap
to IPD, and IPD trained on its full traces alone reaches 29.63 against 29.72 with
partial traces, so partial traces from failed attempts are not the main driver of IPD's gains.
Tables~\ref{tab_data_composition} and~\ref{tab_data_control_results}
report dataset composition and downstream performance.

\begin{table}[H]
\caption{Direct Distillation controls on s1K. Q-match selects questions solved by IPD, T-P (0.1\%); C-based selects verifier-accepted source traces. Partial traces come from the remaining source traces.}
\label{tab_data_composition}
\centering\footnotesize
\begin{tabular}{@{}lccc@{}}
\toprule
Control & Full traces & 128-token partial traces & Total rows \\
\midrule
Q-match Direct, full-only & 516 & \phantom{00}0 & \phantom{1,}516 \\
Q-match Direct & 516 & 484 & 1,000 \\
C-based Direct, full-only & 479 & \phantom{00}0 & \phantom{1,}479 \\
C-based Direct & 479 & 521 & 1,000 \\
\bottomrule
\end{tabular}
\end{table}

\begin{table}[H]
\caption{Benchmark accuracy of the Direct Distillation controls in Table~\ref{tab_data_composition} and IPD with and without partial traces, for Qwen3-0.6B/8B on s1K. Scoring and bold marking follow Table~\ref{tab_family_results}.}
\label{tab_data_control_results}
\centering\scriptsize\setlength{\tabcolsep}{1.6pt}
\begin{tabular}{@{}lcccccccc@{}}
\toprule
Method & AIME24 & AIME25 & AIME26 & HMMT26 & MATH500 & Olympiad & GPQA-D & Average \\
\midrule
Vanilla & $6.25_{\pm 1.67}$ & $15.00_{\pm 1.91}$ & $10.42_{\pm 1.10}$ & $8.96_{\pm 0.79}$ & $71.07_{\pm 0.40}$ & $37.82_{\pm 0.26}$ & $22.35_{\pm 0.91}$ & $24.55$ \\
Direct Distillation & $4.86_{\pm 0.24}$ & $18.33_{\pm 2.20}$ & $10.14_{\pm 0.64}$ & $8.46_{\pm 1.22}$ & $71.63_{\pm 0.75}$ & $35.87_{\pm 0.27}$ & $19.78_{\pm 0.07}$ & $24.15$ \\
Q-match Direct, full-only & $5.28_{\pm 0.64}$ & $16.94_{\pm 1.34}$ & $8.89_{\pm 0.24}$ & $9.85_{\pm 0.66}$ & $72.72_{\pm 1.17}$ & $38.06_{\pm 0.38}$ & $21.34_{\pm 0.77}$ & $24.72$ \\
Q-match Direct & $5.69_{\pm 2.96}$ & $17.36_{\pm 1.27}$ & $8.89_{\pm 1.97}$ & $10.73_{\pm 1.87}$ & $73.22_{\pm 0.16}$ & $37.33_{\pm 0.44}$ & $23.02_{\pm 1.17}$ & $25.18$ \\
C-based Direct, full-only & $5.56_{\pm 0.48}$ & $16.81_{\pm 0.87}$ & $10.97_{\pm 1.68}$ & $10.86_{\pm 1.22}$ & $73.20_{\pm 1.30}$ & $38.37_{\pm 0.37}$ & $22.43_{\pm 0.64}$ & $25.46$ \\
C-based Direct & $7.50_{\pm 1.25}$ & $17.08_{\pm 1.82}$ & $9.58_{\pm 0.83}$ & $10.48_{\pm 0.44}$ & $72.15_{\pm 0.48}$ & $38.24_{\pm 0.33}$ & $26.14_{\pm 1.22}$ & $25.88$ \\
Student RFT-8 & $6.11_{\pm 1.34}$ & $14.44_{\pm 0.64}$ & $10.28_{\pm 0.48}$ & $7.70_{\pm 0.95}$ & $71.42_{\pm 1.35}$ & $38.24_{\pm 0.27}$ & $26.43_{\pm 1.07}$ & $24.95$ \\
Student RFT-100 & $3.89_{\pm 1.05}$ & $16.39_{\pm 1.73}$ & $10.83_{\pm 1.25}$ & $9.34_{\pm 2.15}$ & $71.45_{\pm 0.33}$ & $38.70_{\pm 0.47}$ & $23.53_{\pm 0.62}$ & $24.88$ \\
Teacher RFT & $7.50_{\pm 1.91}$ & $19.86_{\pm 0.64}$ & $11.11_{\pm 3.07}$ & $11.87_{\pm 2.22}$ & $74.40_{\pm 0.40}$ & $41.31_{\pm 0.25}$ & $27.48_{\pm 0.65}$ & $27.65$ \\
IPD, T-P (0.1\%), full-only & $\mathbf{11.67}_{\pm 1.44}$ & $18.61_{\pm 1.58}$ & $13.75_{\pm 1.91}$ & $11.99_{\pm 2.28}$ & $76.25_{\pm 0.79}$ & $42.67_{\pm 0.15}$ & $\mathbf{32.49}_{\pm 1.29}$ & $29.63$ \\
IPD, T-P (0.1\%) & $10.14_{\pm 3.15}$ & $\mathbf{20.56}_{\pm 0.48}$ & $\mathbf{14.03}_{\pm 2.77}$ & $\mathbf{12.37}_{\pm 1.09}$ & $\mathbf{76.30}_{\pm 1.06}$ & $\mathbf{42.89}_{\pm 0.85}$ & $31.78_{\pm 1.39}$ & $\mathbf{29.72}$ \\
\bottomrule
\end{tabular}
\end{table}

\subsection{Trajectory and Learning-Target Ablations}\label{app_mechanism_controls}

Table~\ref{tab_mechanism_design} organizes the controls of Section~\ref{sec_mechanism}.
Trace filtering and loss masking keep completed traces and change only their
supervision; distribution interpolation and SKD generate new traces under a
different sampling rule or learning target.

Trace filtering ranks all 1,000 s1K traces by mean target-token
surprisal under the frozen Qwen3-0.6B student, breaking ties by source index.
The lower-surprisal 500 retain their full traces and termination markers; the
remaining 500 retain at most 128 generated tokens and end abruptly without EOS.
This condition selects by compatibility, not by answer correctness.
Loss masking instead retains all full traces and masks a target whenever its
frozen-student probability falls below the specified threshold. Structural and
termination tokens are subject to the same rule, while masked tokens remain
in the context used to predict later targets.

Distribution interpolation uses the IPD full-generation manifest and condition-native partial traces for unsolved questions,
with the same attempt budget, verifier, and SFT schedule. Its candidate
set is the union of the teacher's and student's top-20 tokens. Each token receives
weight $p_T^\beta p_S^{1-\beta}$ with $\beta=0.5$, normalized over that set, before
applying the shared sampling controls. This symmetric setting weights the teacher
and student equally. Candidate truncation can exclude tokens, but it does
not impose IPD's explicit student-probability floor. The frozen references and
common pipeline make the sampling rule the primary intervention.

SKD is a broader alternative rather than a single-component IPD ablation: it
changes proposal direction, uses a teacher top-$K$ acceptance rule with $K=25$,
and applies a forward-KL distillation target at every position, including those
with student-authored tokens. IPD instead trains on each realized token with
cross-entropy, so student tokens become supervision targets in both directions.
The SKD comparison therefore concerns the overall design.

\clearpage

\section{Student Compatibility and Reasoning Quality}\label{app_selective_analysis}

\subsection{Dataset Diagnostics and Student Surprisal}\label{app_quality_compatibility}

Table~\ref{tab_quality_compatibility_settings} extends Table~\ref{tab_quality_compatibility} to the other three settings.
Fallback rate and mean surprisal average row-level values over all rows, whereas upper-tail means
pool the target tokens, excluding prompts and padding, so a
dataset's row-averaged mean surprisal can differ from its token-pooled mean.

\begin{table}[H]
\caption{Dataset diagnostics for the other three settings. Columns follow Table~\ref{tab_quality_compatibility}, and surprisal is scored with each setting's frozen initial student.}
\label{tab_quality_compatibility_settings}
\centering\scriptsize\setlength{\tabcolsep}{5pt}
\begin{tabular}{lcccccc}
\toprule
Condition & \# Correct & Med. len. (tok) & Fallback (\%) & Mean surp. (nat) & Top 0.01\% surp. (nat) & Eval. avg. \\
\midrule
\rowcolor{gray!15}\multicolumn{7}{l}{\textbf{EXAONE 4.0 1.2B/32B on s1K}} \\
\hspace{0.75em}Direct Distillation & 479 & 10,148.0 & N/A & 1.0045 & 21.8501 & 46.92 \\
\hspace{0.75em}Student RFT & 531 & \phantom{0}9,461.0 & N/A & 0.1813 & \phantom{0}3.6722 & 54.08 \\
\hspace{0.75em}Teacher RFT & 554 & 11,494.0 & N/A & 0.3300 & 11.9172 & 57.93 \\
\hspace{0.75em}IPD, S-P (1\%) & 541 & \phantom{0}9,787.0 & 0.535 & 0.2487 & \phantom{0}8.9435 & 56.43 \\
\hspace{0.75em}IPD, S-P (0.1\%) & 520 & \phantom{0}9,863.0 & 0.182 & 0.2715 & \phantom{0}8.1635 & 55.91 \\
\hspace{0.75em}IPD, S-P (0.01\%) & 498 & \phantom{0}9,830.0 & 0.087 & 0.2844 & \phantom{0}8.7646 & 56.07 \\
\hspace{0.75em}IPD, T-P (1\%) & 529 & 11,430.0 & 0.272 & 0.3248 & \phantom{0}4.9858 & 59.05 \\
\hspace{0.75em}IPD, T-P (0.1\%) & 523 & 11,464.0 & 0.077 & 0.3669 & \phantom{0}6.9855 & 58.16 \\
\hspace{0.75em}IPD, T-P (0.01\%) & 512 & 11,637.5 & 0.043 & 0.3802 & \phantom{0}9.0012 & 57.96 \\
\midrule
\rowcolor{gray!15}\multicolumn{7}{l}{\textbf{Qwen3-0.6B/8B on LIMO}} \\
\hspace{0.75em}Direct Distillation & 737 & 12,093.5 & N/A & 0.7569 & 16.0060 & 24.34 \\
\hspace{0.75em}Student RFT & 203 & \phantom{0}4,448.0 & N/A & 0.2351 & \phantom{0}3.7068 & 24.06 \\
\hspace{0.75em}Teacher RFT & 404 & 11,025.5 & N/A & 0.4260 & 12.8592 & 27.54 \\
\hspace{0.75em}IPD, S-P (1\%) & 270 & \phantom{0}7,457.5 & 1.498 & 0.3403 & 11.5197 & 29.01 \\
\hspace{0.75em}IPD, S-P (0.1\%) & 206 & \phantom{0}5,043.0 & 0.806 & 0.3383 & 10.3399 & 28.00 \\
\hspace{0.75em}IPD, S-P (0.01\%) & 193 & \phantom{0}5,058.0 & 0.440 & 0.3527 & \phantom{0}9.8774 & 27.43 \\
\hspace{0.75em}IPD, T-P (1\%) & 309 & 10,777.0 & 0.486 & 0.4013 & \phantom{0}5.4746 & 29.67 \\
\hspace{0.75em}IPD, T-P (0.1\%) & 305 & 10,903.0 & 0.153 & 0.4477 & \phantom{0}7.0838 & 30.00 \\
\hspace{0.75em}IPD, T-P (0.01\%) & 312 & 10,371.5 & 0.083 & 0.4688 & \phantom{0}9.0348 & 29.77 \\
\midrule
\rowcolor{gray!15}\multicolumn{7}{l}{\textbf{Qwen3-1.7B/8B on s1K}} \\
\hspace{0.75em}Direct Distillation & 479 & 9,549.5 & N/A & 1.3475 & 30.1942 & 38.60 \\
\hspace{0.75em}Student RFT & 417 & 7,944.0 & N/A & 0.1114 & \phantom{0}3.1027 & 45.26 \\
\hspace{0.75em}Teacher RFT & 572 & 9,211.0 & N/A & 0.4313 & 21.7522 & 45.90 \\
\hspace{0.75em}IPD, S-P (1\%) & 473 & 7,761.0 & 1.291 & 0.2381 & 19.3579 & 46.53 \\
\hspace{0.75em}IPD, S-P (0.1\%) & 440 & 7,642.0 & 0.681 & 0.2134 & 17.3729 & 46.76 \\
\hspace{0.75em}IPD, S-P (0.01\%) & 435 & 7,576.0 & 0.351 & 0.1998 & 16.0313 & 46.44 \\
\hspace{0.75em}IPD, T-P (1\%) & 504 & 9,391.5 & 1.265 & 0.2944 & \phantom{0}5.3860 & 47.95 \\
\hspace{0.75em}IPD, T-P (0.1\%) & 519 & 9,451.0 & 0.645 & 0.3851 & \phantom{0}7.2953 & 48.12 \\
\hspace{0.75em}IPD, T-P (0.01\%) & 526 & 9,431.0 & 0.347 & 0.4487 & \phantom{0}9.4782 & 48.54 \\
\bottomrule
\end{tabular}
\end{table}

\subsection{Token-Replacement Analysis}\label{app_token_analysis}

Table~\ref{tab_token_categories} classifies every fallback position in the 516
complete T-P ($\tau=0.1\%$) training traces, pairing each rejected teacher proposal with
the committed student fallback; partial traces are excluded because a
128-token fragment does not establish a reasoning stage.
Classification applies to whole words: subword pieces inherit their word's class,
and a rejected teacher token that begins a word is completed by sampling the
teacher's continuation. Five classes are
assigned by a fixed precedence, so they are exclusive and exhaustive:
\emph{surface form} (markup, delimiters, punctuation, whitespace, enumeration
labels, and closed-class function words), \emph{computational content} (numerals,
identifiers, operators, and mathematical commands inside a derivation),
\emph{reasoning content} (the remaining open-class content words),
\emph{discourse connective} (a word linking steps within a reasoning stage), and
\emph{discourse pivot} (a word shifting the stage, of type realization,
verification, exploration, or integration). Pivot types and the stage-shift
criterion follow \citet{lippmann2025stylesubstancedistilledlanguage}; connectives
follow the Penn Discourse Treebank~\citep{prasad2008pdtb}.

Fixed rules label surface form and computational content directly, and flag words
from a fixed discourse lexicon as candidate connectives or pivots. Only these candidates reach an LLM judge,
\texttt{gpt-5.6-luna}, which sees the prefix, the candidate word, and its
continuation, blind to its source, and decides whether the word links clauses in
context, whether it is a pivot or a connective, and the pivot type. Of the 39 rejected teacher proposals
labeled as pivots, 18 are replaced by student connectives.

\subsection{Trace-Quality Evaluation}\label{app_trace_quality}

Table~\ref{tab_trace_comparisons} measures reasoning quality with correctness held fixed. Training traces are paired on the 387 s1K questions for which both IPD (T-P, $\tau=0.1\%$) and s1K supply a verified-correct full solution. Sampled traces pair the student trained on that IPD data with the Direct Distillation student trained on s1K, both at the headline schedule with training seed 42, on benchmark questions that each student answers correctly in at least one natively terminated attempt at evaluation seed 42; one such attempt per student is drawn uniformly. Every eligible AIME, HMMT, and GPQA-D question enters, and 80 questions each are drawn from MATH500 and OlympiadBench so that neither dominates. Judges see the complete reasoning and final answer.

Three LLM judges, \texttt{gpt-6-sol}, \texttt{claude-opus-5.5}, and \texttt{gemini-3.8-flash}, return a binary verdict, with no tie option, on four criteria and an overall preference, following the taxonomy of \citet{lee2025evaluating}: \emph{problem groundedness}, whether the trace uses only the quantities and conditions given in the problem and answers the question asked, following the hallucination score of ROSCOE~\citep{golovneva2023roscoe} and the relevance label of REVEAL~\citep{jacovi2024reveal}; \emph{step coherence}, whether every step's premises were established earlier in the trace or given in the problem, following the missing-step error of ROSCOE; \emph{logical validity}, whether every step the final argument relies on is correct in calculation and inference, following step-level correctness labels~\citep{lightman2024verify,xia2025reasoneval}; and \emph{no redundancy}, whether every step advances the solution, following the neutral-step label of ReasonEval and the redundancy and repetition scores of ROSCOE. The rubric keeps the criteria exclusive: a false claim counts against validity, an unjustified true claim against coherence, and a retracted attempt against redundancy rather than validity.

Each pair is presented once with anonymized method labels, and the A/B order is randomized, balanced within the training set and within each benchmark~\citep{zheng2023judging}. The win rate is the share of pairs whose verdict favors IPD. Table~\ref{tab_trace_comparisons} reports the mean and sample standard deviation of the three judges' win rates over the pairs all three scored, and Table~\ref{tab_judge_results} reports each judge.

\begin{table}[H]
\caption{IPD win rate (\%) against s1K by judge, for training traces and for traces sampled from the trained students. Table~\ref{tab_trace_comparisons} reports the mean across judges.}
\label{tab_judge_results}
\centering\footnotesize
\begin{tabular}{lcc}
\toprule
Criterion & \begin{tabular}[c]{@{}c@{}}\textbf{Training traces}\\IPD data vs. s1K\end{tabular} & \begin{tabular}[c]{@{}c@{}}\textbf{Sampled traces}\\IPD- vs. s1K-trained\end{tabular} \\
\midrule
\rowcolor{gray!15}\multicolumn{3}{l}{\textbf{\texttt{gpt-6-sol}}} \\
\hspace{0.75em}Problem groundedness & 55.4 & 56.2 \\
\hspace{0.75em}Step coherence & 50.8 & 53.8 \\
\hspace{0.75em}Logical validity & 47.7 & 51.0 \\
\hspace{0.75em}No redundancy & 38.1 & 48.6 \\
\hspace{0.75em}Overall preference & 45.9 & 51.4 \\
\midrule
\rowcolor{gray!15}\multicolumn{3}{l}{\textbf{\texttt{claude-opus-5.5}}} \\
\hspace{0.75em}Problem groundedness & 56.0 & 57.8 \\
\hspace{0.75em}Step coherence & 55.2 & 57.0 \\
\hspace{0.75em}Logical validity & 53.1 & 56.2 \\
\hspace{0.75em}No redundancy & 42.2 & 49.4 \\
\hspace{0.75em}Overall preference & 51.0 & 55.8 \\
\midrule
\rowcolor{gray!15}\multicolumn{3}{l}{\textbf{\texttt{gemini-3.8-flash}}} \\
\hspace{0.75em}Problem groundedness & 51.8 & 53.4 \\
\hspace{0.75em}Step coherence & 49.7 & 57.8 \\
\hspace{0.75em}Logical validity & 50.0 & 55.8 \\
\hspace{0.75em}No redundancy & 43.0 & 52.2 \\
\hspace{0.75em}Overall preference & 46.1 & 55.0 \\
\bottomrule
\end{tabular}
\end{table}

\clearpage

\section{Experiential Self-Teaching}\label{app_self_teacher}

\paragraph{Setup.}
Self-teaching replaces the larger teacher with the student itself, given privileged guidance as additional context: the frozen Qwen3-0.6B student proposes with the guidance placed before the question in its prompt, while the gate and fallback policy use the same checkpoint with the question alone.
Following LLM-as-a-Coach~\citep{ye2026coach}, we build the guidance in three steps.
An external LLM, \texttt{gpt-5.6-luna}, first writes a question-specific rubric from the question alone.
The student then generates rollouts, and our verifier checks their correctness.
Finally, the same LLM acts as a coach: from the question, rubric, rollouts, and correctness bits, it writes guidance supplying an approach, intermediate relations, corrections, and checks, averaging 313.8 tokens.
The coach never receives the reference answer or solution; notes are audited for answer leakage and used only during generation, never as SFT inputs or targets.

Self-Teacher RFT samples from the guided proposer without gating; Self-Teacher IPD uses teacher-propose gating at $\tau=0.1\%$.
All 1,000 s1K questions receive up to eight attempts, retaining an accepted correct trace or a partial trace of at most 128 tokens, and SFT uses 15 epochs at a learning rate of $3\times10^{-5}$.
The four-benchmark comparison uses one base seed, with $k=8$ for AIME and $k=4$ for GPQA-D and MATH500; Average is the unweighted mean of the four benchmark means.

\begin{table}[H]
  \caption{Four-benchmark results for guided self-teaching with Qwen3-0.6B on s1K. Error terms give the standard deviation across evaluation attempts from one evaluation seed.}
  \label{tab_self_teacher_results}
  \vspace{3pt}
  \centering
  \scriptsize
  \setlength{\tabcolsep}{4pt}
  \begin{tabular}{lccccc}
    \toprule
    \textbf{Method} & \textbf{AIME24} & \textbf{AIME25} & \textbf{GPQA-D} & \textbf{MATH500} & \textbf{Average} \\
    \midrule
    Vanilla & $7.92\pm4.34$ & $16.67\pm3.56$ & $22.10\pm2.35$ & $71.30\pm1.58$ & 29.50 \\
    Student RFT & $4.58\pm3.05$ & $13.75\pm3.30$ & $26.01\pm1.87$ & $72.50\pm0.35$ & 29.21 \\
    \midrule
    Self-Teacher RFT & $5.83\pm3.88$ & $12.50\pm3.88$ & $20.20\pm0.71$ & $68.70\pm1.80$ & 26.81 \\
    Self-Teacher IPD (0.1\%) & $4.17\pm4.27$ & $14.17\pm4.96$ & $25.63\pm2.98$ & $70.30\pm0.26$ & 28.57 \\
    \midrule
    Teacher RFT (8B) & $9.58\pm2.14$ & $20.42\pm4.86$ & $28.03\pm3.51$ & $74.25\pm0.57$ & 33.07 \\
    IPD, T-P (8B, 0.1\%) & $13.75\pm1.18$ & $20.83\pm3.88$ & $32.70\pm2.45$ & $77.10\pm1.32$ & 36.10 \\
    \bottomrule
  \end{tabular}
\end{table}

\paragraph{Results.}
Table~\ref{tab_self_teacher_results} reports the four original benchmarks for this experiment, separately from the headline seven-benchmark aggregates.
Self-Teacher IPD produces 248 correct full traces, compared with 219 for Student RFT and 206 for Self-Teacher RFT.
Guidance also makes the traces more surprising to the student: mean training-target surprisal under the frozen student rises from 0.2366 for Student RFT to 0.4353 for Self-Teacher RFT and 0.5312 for Self-Teacher IPD.
Yet the trained student is no better: Self-Teacher IPD averages 28.57, below both Student RFT (29.21) and Vanilla (29.50), and Self-Teacher RFT is lower still at 26.81.
The traces still contain only reasoning the student itself produces, consistent with Sections~\ref{sec_tradeoff} and~\ref{sec_proposal_source}: compatibility without a stronger source of reasoning adds little.

\clearpage

\section{Full Benchmark Results}\label{app_full_tables}

\subsection{Results Across Training Seeds}\label{app_seed_results}

For fine-tuned models, seeds 42, 1, and 2026 index three independent SFT training runs of the same
condition on the same data; they are not evaluation sampling seeds. The
subscripts in Tables~\ref{tab_seed_qwen_s1k}--\ref{tab_seed_qwen17_s1k} instead
give the dispersion across the $k$ matched attempt runs within one evaluation,
and the across-seed standard deviation for fine-tuned models in
Table~\ref{tab_family_results} measures how much the training seed moves the result.

Tables~\ref{tab_seed_qwen_s1k}--\ref{tab_seed_qwen17_s1k} show that the strongest
seven-benchmark average belongs to a T-P configuration in all
twelve setting and training-seed pairs.
T-P at $\tau=1\%$ exceeds Teacher RFT in all twelve pairs,
and at the representative $\tau=0.1\%$ in eleven; the exception is EXAONE at
seed 1, where it trails by $0.43$ points.

\begin{table}[H]
  \caption{Accuracy by seed for Qwen3-0.6B/8B on s1K. For fine-tuned models, seeds identify independent training runs; subscripts give the within-run standard deviation over evaluation attempts.}
  \label{tab_seed_qwen_s1k}
  \vspace{3pt}
  \centering
  \scriptsize
  \setlength{\tabcolsep}{1.6pt}
  \begin{tabular}{lrcccccccc}
    \toprule
    \textbf{Method} & \textbf{Seed} & \textbf{AIME24} & \textbf{AIME25} & \textbf{AIME26} & \textbf{HMMT26} & \textbf{MATH500} & \textbf{Olympiad} & \textbf{GPQA-D} & \textbf{Average} \\
    \midrule
    Vanilla & 42 & $7.92_{\pm 4.34}$ & $16.67_{\pm 3.56}$ & $9.58_{\pm 4.15}$ & $9.85_{\pm 5.06}$ & $71.30_{\pm 1.58}$ & $38.06_{\pm 1.45}$ & $22.10_{\pm 2.35}$ & $25.07$ \\
    Vanilla & 1 & $6.25_{\pm 4.15}$ & $15.42_{\pm 3.54}$ & $11.67_{\pm 3.09}$ & $8.71_{\pm 2.53}$ & $71.30_{\pm 1.91}$ & $37.54_{\pm 1.38}$ & $23.36_{\pm 1.12}$ & $24.89$ \\
    Vanilla & 2026 & $4.58_{\pm 5.02}$ & $12.92_{\pm 2.14}$ & $10.00_{\pm 5.63}$ & $8.33_{\pm 3.88}$ & $70.60_{\pm 0.95}$ & $37.87_{\pm 1.73}$ & $21.59_{\pm 3.43}$ & $23.70$ \\
    \midrule
    Direct Distillation & 42 & $5.00_{\pm 3.56}$ & $19.17_{\pm 2.95}$ & $10.83_{\pm 2.95}$ & $7.58_{\pm 2.81}$ & $71.20_{\pm 0.97}$ & $35.65_{\pm 1.56}$ & $19.82_{\pm 1.51}$ & $24.18$ \\
    Direct Distillation & 1 & $5.00_{\pm 3.09}$ & $15.83_{\pm 2.95}$ & $10.00_{\pm 5.35}$ & $9.85_{\pm 4.21}$ & $71.20_{\pm 1.10}$ & $36.16_{\pm 1.10}$ & $19.70_{\pm 1.80}$ & $23.96$ \\
    Direct Distillation & 2026 & $4.58_{\pm 3.05}$ & $20.00_{\pm 4.36}$ & $9.58_{\pm 6.53}$ & $7.95_{\pm 3.21}$ & $72.50_{\pm 1.89}$ & $35.79_{\pm 0.78}$ & $19.82_{\pm 2.75}$ & $24.32$ \\
    \midrule
    Student RFT & 42 & $4.58_{\pm 3.05}$ & $13.75_{\pm 3.30}$ & $10.00_{\pm 3.98}$ & $8.71_{\pm 3.78}$ & $72.50_{\pm 0.35}$ & $38.02_{\pm 1.07}$ & $26.01_{\pm 1.87}$ & $24.80$ \\
    Student RFT & 1 & $7.08_{\pm 2.78}$ & $15.00_{\pm 3.56}$ & $10.83_{\pm 4.96}$ & $7.58_{\pm 2.29}$ & $69.90_{\pm 1.41}$ & $38.54_{\pm 1.34}$ & $25.63_{\pm 1.91}$ & $24.94$ \\
    Student RFT & 2026 & $6.67_{\pm 6.17}$ & $14.58_{\pm 4.69}$ & $10.00_{\pm 5.04}$ & $6.82_{\pm 4.51}$ & $71.85_{\pm 1.00}$ & $38.17_{\pm 0.96}$ & $27.65_{\pm 6.20}$ & $25.11$ \\
    \midrule
    Teacher RFT & 42 & $9.58_{\pm 2.14}$ & $20.42_{\pm 4.86}$ & $10.00_{\pm 5.04}$ & $10.98_{\pm 2.25}$ & $74.25_{\pm 0.57}$ & $41.10_{\pm 0.90}$ & $28.03_{\pm 3.51}$ & $27.77$ \\
    Teacher RFT & 1 & $5.83_{\pm 3.88}$ & $20.00_{\pm 3.98}$ & $8.75_{\pm 3.05}$ & $14.39_{\pm 3.53}$ & $74.85_{\pm 2.30}$ & $41.25_{\pm 0.44}$ & $26.77_{\pm 1.80}$ & $27.41$ \\
    Teacher RFT & 2026 & $7.08_{\pm 3.30}$ & $19.17_{\pm 1.54}$ & $14.58_{\pm 4.69}$ & $10.23_{\pm 3.95}$ & $74.10_{\pm 0.50}$ & $41.58_{\pm 0.99}$ & $27.65_{\pm 2.81}$ & $27.77$ \\
    \midrule
    IPD, S-P (1\%) & 42 & $10.83_{\pm 4.63}$ & $19.58_{\pm 4.52}$ & $18.75_{\pm 5.33}$ & $13.64_{\pm 3.62}$ & $74.95_{\pm 0.68}$ & $42.66_{\pm 1.07}$ & $25.25_{\pm 3.22}$ & $29.38$ \\
    IPD, S-P (1\%) & 1 & $11.25_{\pm 2.48}$ & $19.58_{\pm 3.30}$ & $13.33_{\pm 4.36}$ & $9.85_{\pm 2.14}$ & $78.05_{\pm 1.34}$ & $43.32_{\pm 0.79}$ & $23.61_{\pm 1.77}$ & $28.43$ \\
    IPD, S-P (1\%) & 2026 & $11.67_{\pm 1.78}$ & $19.58_{\pm 4.15}$ & $16.25_{\pm 7.22}$ & $13.64_{\pm 3.97}$ & $76.35_{\pm 0.90}$ & $43.40_{\pm 1.13}$ & $25.51_{\pm 1.73}$ & $29.48$ \\
    \midrule
    IPD, S-P (0.1\%) & 42 & $7.92_{\pm 3.05}$ & $20.00_{\pm 5.63}$ & $14.58_{\pm 6.16}$ & $12.50_{\pm 4.42}$ & $76.10_{\pm 1.32}$ & $42.14_{\pm 0.63}$ & $24.12_{\pm 1.12}$ & $28.19$ \\
    IPD, S-P (0.1\%) & 1 & $7.50_{\pm 3.45}$ & $19.58_{\pm 4.15}$ & $13.33_{\pm 3.98}$ & $13.26_{\pm 3.60}$ & $75.00_{\pm 0.97}$ & $41.84_{\pm 1.16}$ & $25.00_{\pm 2.42}$ & $27.93$ \\
    IPD, S-P (0.1\%) & 2026 & $12.92_{\pm 7.65}$ & $20.00_{\pm 4.36}$ & $14.58_{\pm 5.62}$ & $13.26_{\pm 6.26}$ & $76.95_{\pm 2.41}$ & $40.80_{\pm 1.28}$ & $23.74_{\pm 2.51}$ & $28.89$ \\
    \midrule
    IPD, S-P (0.01\%) & 42 & $9.17_{\pm 2.95}$ & $17.08_{\pm 4.86}$ & $16.25_{\pm 3.30}$ & $11.36_{\pm 3.88}$ & $75.25_{\pm 1.51}$ & $42.54_{\pm 0.78}$ & $29.17_{\pm 1.26}$ & $28.69$ \\
    IPD, S-P (0.01\%) & 1 & $8.33_{\pm 5.35}$ & $18.75_{\pm 3.96}$ & $13.75_{\pm 4.86}$ & $9.85_{\pm 2.14}$ & $74.25_{\pm 0.82}$ & $41.43_{\pm 1.47}$ & $32.32_{\pm 2.51}$ & $28.38$ \\
    IPD, S-P (0.01\%) & 2026 & $10.42_{\pm 4.15}$ & $20.00_{\pm 3.56}$ & $16.67_{\pm 3.98}$ & $11.74_{\pm 3.41}$ & $74.80_{\pm 0.71}$ & $41.62_{\pm 2.18}$ & $28.66_{\pm 2.31}$ & $29.13$ \\
    \midrule
    IPD, T-P (1\%) & 42 & $11.25_{\pm 6.16}$ & $20.42_{\pm 2.78}$ & $12.50_{\pm 2.36}$ & $12.50_{\pm 4.11}$ & $77.75_{\pm 0.25}$ & $43.10_{\pm 0.61}$ & $31.69_{\pm 1.82}$ & $29.89$ \\
    IPD, T-P (1\%) & 1 & $7.50_{\pm 3.88}$ & $22.08_{\pm 5.62}$ & $12.92_{\pm 3.30}$ & $12.88_{\pm 4.79}$ & $78.15_{\pm 2.36}$ & $43.92_{\pm 1.04}$ & $29.17_{\pm 2.52}$ & $29.52$ \\
    IPD, T-P (1\%) & 2026 & $12.08_{\pm 5.89}$ & $17.92_{\pm 3.96}$ & $10.83_{\pm 4.27}$ & $15.53_{\pm 3.78}$ & $77.45_{\pm 0.66}$ & $42.51_{\pm 0.55}$ & $29.04_{\pm 1.34}$ & $29.34$ \\
    \midrule
    IPD, T-P (0.1\%) & 42 & $13.75_{\pm 1.18}$ & $20.83_{\pm 3.88}$ & $17.08_{\pm 4.52}$ & $11.74_{\pm 4.42}$ & $77.10_{\pm 1.32}$ & $43.84_{\pm 1.11}$ & $32.70_{\pm 2.45}$ & $31.01$ \\
    IPD, T-P (0.1\%) & 1 & $7.92_{\pm 5.02}$ & $20.00_{\pm 4.71}$ & $11.67_{\pm 5.63}$ & $11.74_{\pm 6.96}$ & $75.10_{\pm 1.14}$ & $42.21_{\pm 0.71}$ & $32.45_{\pm 0.64}$ & $28.73$ \\
    IPD, T-P (0.1\%) & 2026 & $8.75_{\pm 2.48}$ & $20.83_{\pm 4.27}$ & $13.33_{\pm 4.71}$ & $13.64_{\pm 3.62}$ & $76.70_{\pm 1.31}$ & $42.62_{\pm 1.52}$ & $30.18_{\pm 4.56}$ & $29.44$ \\
    \midrule
    IPD, T-P (0.01\%) & 42 & $8.33_{\pm 3.56}$ & $16.67_{\pm 4.71}$ & $13.33_{\pm 3.09}$ & $12.88_{\pm 3.14}$ & $75.95_{\pm 0.91}$ & $41.36_{\pm 1.17}$ & $35.35_{\pm 1.49}$ & $29.12$ \\
    IPD, T-P (0.01\%) & 1 & $9.17_{\pm 4.96}$ & $20.42_{\pm 3.30}$ & $14.58_{\pm 4.34}$ & $17.42_{\pm 3.14}$ & $74.95_{\pm 0.53}$ & $41.36_{\pm 1.08}$ & $37.88_{\pm 3.57}$ & $30.83$ \\
    IPD, T-P (0.01\%) & 2026 & $7.08_{\pm 4.52}$ & $22.50_{\pm 3.88}$ & $11.67_{\pm 4.36}$ & $12.12_{\pm 4.29}$ & $75.40_{\pm 1.19}$ & $41.51_{\pm 1.02}$ & $38.26_{\pm 2.75}$ & $29.79$ \\
    \bottomrule
  \end{tabular}
\end{table}

\begin{table}[H]
  \caption{Accuracy by seed for EXAONE 4.0 1.2B/32B on s1K. For fine-tuned models, seeds identify independent training runs; subscripts give the within-run standard deviation over evaluation attempts.}
  \label{tab_seed_exaone_s1k}
  \vspace{3pt}
  \centering
  \scriptsize
  \setlength{\tabcolsep}{1.6pt}
  \begin{tabular}{lrcccccccc}
    \toprule
    \textbf{Method} & \textbf{Seed} & \textbf{AIME24} & \textbf{AIME25} & \textbf{AIME26} & \textbf{HMMT26} & \textbf{MATH500} & \textbf{Olympiad} & \textbf{GPQA-D} & \textbf{Average} \\
    \midrule
    Vanilla & 42 & $40.00_{\pm 5.04}$ & $37.08_{\pm 2.14}$ & $49.58_{\pm 7.86}$ & $28.79_{\pm 5.37}$ & $89.60_{\pm 1.31}$ & $65.91_{\pm 1.16}$ & $46.34_{\pm 1.12}$ & $51.04$ \\
    Vanilla & 1 & $37.92_{\pm 4.34}$ & $42.92_{\pm 3.75}$ & $46.67_{\pm 3.56}$ & $33.33_{\pm 4.58}$ & $89.85_{\pm 0.66}$ & $67.21_{\pm 0.71}$ & $48.61_{\pm 1.45}$ & $52.36$ \\
    Vanilla & 2026 & $37.50_{\pm 2.95}$ & $43.33_{\pm 5.91}$ & $52.50_{\pm 4.63}$ & $28.41_{\pm 5.36}$ & $90.00_{\pm 1.42}$ & $64.91_{\pm 1.32}$ & $48.11_{\pm 1.95}$ & $52.11$ \\
    \midrule
    Direct Distillation & 42 & $31.67_{\pm 5.91}$ & $39.17_{\pm 3.88}$ & $44.17_{\pm 5.27}$ & $32.20_{\pm 2.78}$ & $89.40_{\pm 0.78}$ & $58.90_{\pm 1.39}$ & $33.96_{\pm 1.62}$ & $47.07$ \\
    Direct Distillation & 1 & $32.92_{\pm 4.86}$ & $36.25_{\pm 5.47}$ & $44.17_{\pm 4.96}$ & $30.68_{\pm 4.42}$ & $89.90_{\pm 0.48}$ & $58.79_{\pm 0.53}$ & $33.21_{\pm 3.31}$ & $46.56$ \\
    Direct Distillation & 2026 & $35.00_{\pm 6.67}$ & $35.42_{\pm 6.65}$ & $43.75_{\pm 4.86}$ & $32.20_{\pm 3.60}$ & $90.00_{\pm 0.95}$ & $58.90_{\pm 0.62}$ & $34.60_{\pm 1.68}$ & $47.12$ \\
    \midrule
    Student RFT & 42 & $39.17_{\pm 5.27}$ & $46.25_{\pm 5.18}$ & $53.33_{\pm 7.56}$ & $33.33_{\pm 2.81}$ & $90.40_{\pm 0.71}$ & $67.69_{\pm 0.61}$ & $47.10_{\pm 2.42}$ & $53.90$ \\
    Student RFT & 1 & $42.08_{\pm 4.69}$ & $44.17_{\pm 6.36}$ & $54.17_{\pm 6.11}$ & $32.58_{\pm 3.88}$ & $90.10_{\pm 1.59}$ & $66.95_{\pm 1.20}$ & $47.98_{\pm 4.76}$ & $54.00$ \\
    Student RFT & 2026 & $40.00_{\pm 5.91}$ & $42.92_{\pm 4.52}$ & $53.75_{\pm 6.53}$ & $32.95_{\pm 4.11}$ & $90.95_{\pm 0.66}$ & $68.40_{\pm 1.28}$ & $51.39_{\pm 2.42}$ & $54.34$ \\
    \midrule
    Teacher RFT & 42 & $48.75_{\pm 5.62}$ & $53.33_{\pm 6.42}$ & $57.92_{\pm 7.96}$ & $35.61_{\pm 4.79}$ & $92.40_{\pm 1.08}$ & $67.84_{\pm 1.85}$ & $50.13_{\pm 1.95}$ & $58.00$ \\
    Teacher RFT & 1 & $48.75_{\pm 4.69}$ & $55.83_{\pm 7.07}$ & $57.50_{\pm 7.07}$ & $37.50_{\pm 5.36}$ & $90.65_{\pm 1.18}$ & $67.84_{\pm 2.27}$ & $49.87_{\pm 3.63}$ & $58.28$ \\
    Teacher RFT & 2026 & $47.92_{\pm 5.33}$ & $52.50_{\pm 6.36}$ & $56.67_{\pm 6.90}$ & $35.23_{\pm 4.27}$ & $90.85_{\pm 0.57}$ & $68.92_{\pm 0.61}$ & $50.63_{\pm 1.99}$ & $57.53$ \\
    \midrule
    IPD, S-P (1\%) & 42 & $43.75_{\pm 5.18}$ & $51.67_{\pm 7.13}$ & $58.75_{\pm 7.33}$ & $32.58_{\pm 4.79}$ & $91.65_{\pm 0.72}$ & $69.62_{\pm 0.14}$ & $48.61_{\pm 3.15}$ & $56.66$ \\
    IPD, S-P (1\%) & 1 & $42.08_{\pm 7.11}$ & $48.75_{\pm 7.33}$ & $57.50_{\pm 11.23}$ & $33.71_{\pm 4.42}$ & $91.25_{\pm 0.53}$ & $68.40_{\pm 0.76}$ & $51.14_{\pm 2.20}$ & $56.12$ \\
    IPD, S-P (1\%) & 2026 & $46.67_{\pm 9.59}$ & $52.92_{\pm 5.47}$ & $54.17_{\pm 6.36}$ & $33.71_{\pm 3.00}$ & $90.30_{\pm 0.93}$ & $68.77_{\pm 1.28}$ & $48.99_{\pm 1.70}$ & $56.50$ \\
    \midrule
    IPD, S-P (0.1\%) & 42 & $44.17_{\pm 3.45}$ & $47.92_{\pm 5.33}$ & $53.33_{\pm 6.90}$ & $34.85_{\pm 5.84}$ & $90.95_{\pm 0.10}$ & $67.88_{\pm 1.65}$ & $50.25_{\pm 2.90}$ & $55.62$ \\
    IPD, S-P (0.1\%) & 1 & $45.42_{\pm 3.05}$ & $47.92_{\pm 7.75}$ & $55.83_{\pm 4.96}$ & $32.20_{\pm 4.27}$ & $90.85_{\pm 0.66}$ & $69.44_{\pm 1.52}$ & $49.75_{\pm 1.34}$ & $55.91$ \\
    IPD, S-P (0.1\%) & 2026 & $45.83_{\pm 7.51}$ & $46.67_{\pm 7.13}$ & $56.67_{\pm 4.36}$ & $34.09_{\pm 6.22}$ & $90.45_{\pm 0.66}$ & $68.32_{\pm 0.86}$ & $51.39_{\pm 1.26}$ & $56.20$ \\
    \midrule
    IPD, S-P (0.01\%) & 42 & $45.83_{\pm 4.96}$ & $50.00_{\pm 7.56}$ & $53.75_{\pm 5.47}$ & $35.61_{\pm 8.83}$ & $90.70_{\pm 0.62}$ & $68.03_{\pm 1.05}$ & $50.00_{\pm 1.43}$ & $56.27$ \\
    IPD, S-P (0.01\%) & 1 & $41.25_{\pm 5.33}$ & $49.17_{\pm 6.61}$ & $56.67_{\pm 5.91}$ & $34.85_{\pm 4.29}$ & $89.00_{\pm 0.16}$ & $69.51_{\pm 0.91}$ & $50.88_{\pm 2.20}$ & $55.90$ \\
    IPD, S-P (0.01\%) & 2026 & $42.50_{\pm 6.11}$ & $47.92_{\pm 10.07}$ & $57.08_{\pm 3.30}$ & $34.09_{\pm 6.01}$ & $90.10_{\pm 0.62}$ & $68.66_{\pm 1.75}$ & $51.89_{\pm 3.51}$ & $56.03$ \\
    \midrule
    IPD, T-P (1\%) & 42 & $47.08_{\pm 5.18}$ & $55.00_{\pm 4.36}$ & $59.58_{\pm 4.86}$ & $33.71_{\pm 3.41}$ & $92.20_{\pm 0.99}$ & $69.03_{\pm 0.93}$ & $49.49_{\pm 2.06}$ & $58.01$ \\
    IPD, T-P (1\%) & 1 & $49.17_{\pm 6.11}$ & $58.33_{\pm 8.36}$ & $62.50_{\pm 4.27}$ & $34.85_{\pm 4.86}$ & $91.55_{\pm 0.25}$ & $69.44_{\pm 0.93}$ & $53.79_{\pm 3.04}$ & $59.95$ \\
    IPD, T-P (1\%) & 2026 & $47.50_{\pm 2.95}$ & $57.08_{\pm 3.75}$ & $60.42_{\pm 3.30}$ & $36.74_{\pm 5.94}$ & $91.50_{\pm 0.48}$ & $70.25_{\pm 0.15}$ & $50.76_{\pm 3.41}$ & $59.18$ \\
    \midrule
    IPD, T-P (0.1\%) & 42 & $46.25_{\pm 3.30}$ & $55.83_{\pm 2.95}$ & $64.17_{\pm 6.11}$ & $34.09_{\pm 5.78}$ & $92.15_{\pm 1.22}$ & $67.69_{\pm 1.71}$ & $53.16_{\pm 3.94}$ & $59.05$ \\
    IPD, T-P (0.1\%) & 1 & $46.67_{\pm 2.52}$ & $57.08_{\pm 5.47}$ & $59.17_{\pm 6.61}$ & $31.82_{\pm 4.29}$ & $91.40_{\pm 0.57}$ & $68.03_{\pm 0.89}$ & $50.76_{\pm 1.96}$ & $57.85$ \\
    IPD, T-P (0.1\%) & 2026 & $47.50_{\pm 4.96}$ & $50.83_{\pm 5.27}$ & $60.83_{\pm 5.56}$ & $32.58_{\pm 6.22}$ & $91.75_{\pm 0.75}$ & $67.51_{\pm 0.27}$ & $52.02_{\pm 4.31}$ & $57.57$ \\
    \midrule
    IPD, T-P (0.01\%) & 42 & $49.17_{\pm 3.88}$ & $56.67_{\pm 5.35}$ & $57.50_{\pm 7.29}$ & $36.36_{\pm 3.97}$ & $91.45_{\pm 0.81}$ & $68.95_{\pm 1.07}$ & $45.08_{\pm 2.81}$ & $57.88$ \\
    IPD, T-P (0.01\%) & 1 & $45.42_{\pm 3.96}$ & $57.08_{\pm 6.77}$ & $65.42_{\pm 5.33}$ & $29.55_{\pm 4.79}$ & $91.55_{\pm 0.44}$ & $69.47_{\pm 1.08}$ & $47.10_{\pm 3.07}$ & $57.94$ \\
    IPD, T-P (0.01\%) & 2026 & $45.83_{\pm 5.84}$ & $58.33_{\pm 6.17}$ & $60.00_{\pm 8.54}$ & $32.95_{\pm 5.23}$ & $91.15_{\pm 0.75}$ & $70.18_{\pm 0.74}$ & $47.98_{\pm 3.14}$ & $58.06$ \\
    \bottomrule
  \end{tabular}
\end{table}

\begin{table}[H]
  \caption{Accuracy by seed for Qwen3-0.6B/8B on LIMO. For fine-tuned models, seeds identify independent training runs; subscripts give the within-run standard deviation over evaluation attempts.}
  \label{tab_seed_qwen_limo}
  \vspace{3pt}
  \centering
  \scriptsize
  \setlength{\tabcolsep}{1.6pt}
  \begin{tabular}{lrcccccccc}
    \toprule
    \textbf{Method} & \textbf{Seed} & \textbf{AIME24} & \textbf{AIME25} & \textbf{AIME26} & \textbf{HMMT26} & \textbf{MATH500} & \textbf{Olympiad} & \textbf{GPQA-D} & \textbf{Average} \\
    \midrule
    Vanilla & 42 & $7.92_{\pm 4.34}$ & $16.67_{\pm 3.56}$ & $9.58_{\pm 4.15}$ & $9.85_{\pm 5.06}$ & $71.30_{\pm 1.58}$ & $38.06_{\pm 1.45}$ & $22.10_{\pm 2.35}$ & $25.07$ \\
    Vanilla & 1 & $6.25_{\pm 4.15}$ & $15.42_{\pm 3.54}$ & $11.67_{\pm 3.09}$ & $8.71_{\pm 2.53}$ & $71.30_{\pm 1.91}$ & $37.54_{\pm 1.38}$ & $23.36_{\pm 1.12}$ & $24.89$ \\
    Vanilla & 2026 & $4.58_{\pm 5.02}$ & $12.92_{\pm 2.14}$ & $10.00_{\pm 5.63}$ & $8.33_{\pm 3.88}$ & $70.60_{\pm 0.95}$ & $37.87_{\pm 1.73}$ & $21.59_{\pm 3.43}$ & $23.70$ \\
    \midrule
    Direct Distillation & 42 & $9.17_{\pm 7.72}$ & $17.08_{\pm 5.18}$ & $10.83_{\pm 4.27}$ & $7.95_{\pm 3.21}$ & $70.85_{\pm 1.95}$ & $38.24_{\pm 0.44}$ & $17.68_{\pm 0.71}$ & $24.54$ \\
    Direct Distillation & 1 & $7.50_{\pm 3.45}$ & $15.00_{\pm 5.35}$ & $9.17_{\pm 4.27}$ & $8.71_{\pm 1.94}$ & $70.95_{\pm 0.60}$ & $38.24_{\pm 1.21}$ & $17.30_{\pm 2.65}$ & $23.84$ \\
    Direct Distillation & 2026 & $7.92_{\pm 3.05}$ & $17.50_{\pm 4.96}$ & $10.83_{\pm 3.88}$ & $7.95_{\pm 3.95}$ & $71.95_{\pm 0.57}$ & $39.17_{\pm 1.23}$ & $17.17_{\pm 2.10}$ & $24.64$ \\
    \midrule
    Student RFT & 42 & $2.08_{\pm 2.48}$ & $17.08_{\pm 4.15}$ & $7.92_{\pm 2.48}$ & $8.71_{\pm 2.53}$ & $71.70_{\pm 1.87}$ & $37.05_{\pm 1.45}$ & $24.49_{\pm 1.91}$ & $24.15$ \\
    Student RFT & 1 & $3.75_{\pm 5.76}$ & $15.00_{\pm 6.42}$ & $9.58_{\pm 3.30}$ & $9.85_{\pm 4.21}$ & $67.50_{\pm 1.18}$ & $37.69_{\pm 1.55}$ & $23.36_{\pm 1.51}$ & $23.82$ \\
    Student RFT & 2026 & $7.08_{\pm 3.30}$ & $13.33_{\pm 2.52}$ & $11.25_{\pm 3.54}$ & $7.58_{\pm 2.81}$ & $68.00_{\pm 1.45}$ & $37.76_{\pm 1.46}$ & $24.49_{\pm 0.97}$ & $24.21$ \\
    \midrule
    Teacher RFT & 42 & $7.92_{\pm 3.54}$ & $17.92_{\pm 4.34}$ & $15.00_{\pm 3.09}$ & $11.36_{\pm 3.88}$ & $74.70_{\pm 0.58}$ & $41.06_{\pm 1.21}$ & $28.41_{\pm 2.20}$ & $28.05$ \\
    Teacher RFT & 1 & $7.50_{\pm 5.27}$ & $17.08_{\pm 2.14}$ & $16.25_{\pm 6.03}$ & $12.12_{\pm 4.29}$ & $75.15_{\pm 1.05}$ & $41.73_{\pm 1.68}$ & $22.73_{\pm 2.14}$ & $27.51$ \\
    Teacher RFT & 2026 & $7.92_{\pm 2.48}$ & $14.17_{\pm 4.96}$ & $17.08_{\pm 3.30}$ & $12.12_{\pm 1.62}$ & $73.45_{\pm 1.18}$ & $40.91_{\pm 1.77}$ & $23.74_{\pm 2.47}$ & $27.06$ \\
    \midrule
    IPD, S-P (1\%) & 42 & $7.92_{\pm 4.69}$ & $18.75_{\pm 2.48}$ & $13.75_{\pm 7.00}$ & $14.02_{\pm 3.21}$ & $77.40_{\pm 2.73}$ & $42.99_{\pm 0.95}$ & $25.38_{\pm 3.23}$ & $28.60$ \\
    IPD, S-P (1\%) & 1 & $10.00_{\pm 3.56}$ & $18.75_{\pm 5.89}$ & $15.83_{\pm 3.88}$ & $14.77_{\pm 1.94}$ & $75.95_{\pm 1.94}$ & $43.51_{\pm 0.85}$ & $28.79_{\pm 2.70}$ & $29.66$ \\
    IPD, S-P (1\%) & 2026 & $7.08_{\pm 3.30}$ & $18.75_{\pm 4.34}$ & $14.17_{\pm 3.45}$ & $15.15_{\pm 2.29}$ & $76.45_{\pm 0.66}$ & $42.95_{\pm 0.83}$ & $26.77_{\pm 2.86}$ & $28.76$ \\
    \midrule
    IPD, S-P (0.1\%) & 42 & $8.33_{\pm 3.56}$ & $18.33_{\pm 3.98}$ & $13.33_{\pm 3.98}$ & $11.74_{\pm 4.71}$ & $75.25_{\pm 1.89}$ & $41.77_{\pm 1.54}$ & $26.39_{\pm 1.91}$ & $27.88$ \\
    IPD, S-P (0.1\%) & 1 & $7.50_{\pm 3.45}$ & $19.17_{\pm 6.84}$ & $12.08_{\pm 3.96}$ & $10.98_{\pm 4.84}$ & $75.70_{\pm 0.48}$ & $41.51_{\pm 0.90}$ & $29.67_{\pm 1.99}$ & $28.09$ \\
    IPD, S-P (0.1\%) & 2026 & $7.92_{\pm 3.96}$ & $19.17_{\pm 3.45}$ & $13.33_{\pm 2.52}$ & $12.50_{\pm 6.57}$ & $75.80_{\pm 0.82}$ & $41.14_{\pm 0.87}$ & $26.39_{\pm 4.07}$ & $28.03$ \\
    \midrule
    IPD, S-P (0.01\%) & 42 & $9.58_{\pm 4.15}$ & $17.08_{\pm 5.76}$ & $14.17_{\pm 6.11}$ & $12.50_{\pm 3.00}$ & $75.20_{\pm 0.85}$ & $41.77_{\pm 0.83}$ & $24.49_{\pm 2.66}$ & $27.83$ \\
    IPD, S-P (0.01\%) & 1 & $8.33_{\pm 3.98}$ & $15.00_{\pm 5.35}$ & $13.33_{\pm 5.04}$ & $12.12_{\pm 2.81}$ & $75.40_{\pm 1.19}$ & $40.91_{\pm 1.74}$ & $25.88_{\pm 2.78}$ & $27.28$ \\
    IPD, S-P (0.01\%) & 2026 & $9.17_{\pm 2.95}$ & $16.25_{\pm 2.14}$ & $16.25_{\pm 3.30}$ & $9.09_{\pm 3.24}$ & $74.60_{\pm 1.98}$ & $40.91_{\pm 0.37}$ & $23.99_{\pm 3.34}$ & $27.18$ \\
    \midrule
    IPD, T-P (1\%) & 42 & $11.67_{\pm 4.71}$ & $18.33_{\pm 6.42}$ & $16.67_{\pm 3.09}$ & $12.88_{\pm 3.53}$ & $78.05_{\pm 1.43}$ & $45.10_{\pm 0.54}$ & $28.41_{\pm 2.27}$ & $30.16$ \\
    IPD, T-P (1\%) & 1 & $9.17_{\pm 4.63}$ & $20.00_{\pm 5.04}$ & $13.75_{\pm 6.53}$ & $12.50_{\pm 4.11}$ & $78.35_{\pm 0.50}$ & $45.10_{\pm 1.02}$ & $27.78_{\pm 1.70}$ & $29.52$ \\
    IPD, T-P (1\%) & 2026 & $7.92_{\pm 3.96}$ & $19.58_{\pm 3.30}$ & $12.50_{\pm 2.36}$ & $12.88_{\pm 2.69}$ & $77.55_{\pm 1.48}$ & $44.99_{\pm 0.95}$ & $29.92_{\pm 1.91}$ & $29.34$ \\
    \midrule
    IPD, T-P (0.1\%) & 42 & $9.17_{\pm 2.95}$ & $17.92_{\pm 4.69}$ & $17.50_{\pm 5.56}$ & $13.64_{\pm 3.24}$ & $76.15_{\pm 1.18}$ & $44.73_{\pm 1.60}$ & $31.44_{\pm 2.20}$ & $30.08$ \\
    IPD, T-P (0.1\%) & 1 & $10.42_{\pm 4.52}$ & $18.33_{\pm 4.36}$ & $15.42_{\pm 3.05}$ & $11.74_{\pm 5.23}$ & $76.75_{\pm 0.93}$ & $43.99_{\pm 1.50}$ & $31.57_{\pm 2.81}$ & $29.75$ \\
    IPD, T-P (0.1\%) & 2026 & $8.33_{\pm 6.67}$ & $17.50_{\pm 4.27}$ & $15.83_{\pm 5.84}$ & $13.26_{\pm 4.56}$ & $77.45_{\pm 0.38}$ & $43.55_{\pm 1.50}$ & $35.23_{\pm 1.56}$ & $30.16$ \\
    \midrule
    IPD, T-P (0.01\%) & 42 & $10.00_{\pm 2.52}$ & $19.17_{\pm 7.07}$ & $15.00_{\pm 5.35}$ & $14.02_{\pm 3.95}$ & $77.00_{\pm 1.55}$ & $43.18_{\pm 1.10}$ & $28.41_{\pm 2.04}$ & $29.54$ \\
    IPD, T-P (0.01\%) & 1 & $11.67_{\pm 5.63}$ & $20.42_{\pm 4.15}$ & $16.25_{\pm 3.75}$ & $13.64_{\pm 2.29}$ & $77.75_{\pm 1.29}$ & $44.07_{\pm 1.95}$ & $27.78_{\pm 0.82}$ & $30.22$ \\
    IPD, T-P (0.01\%) & 2026 & $12.92_{\pm 4.15}$ & $18.33_{\pm 5.35}$ & $13.75_{\pm 4.52}$ & $12.88_{\pm 5.55}$ & $76.90_{\pm 0.93}$ & $44.14_{\pm 0.37}$ & $27.90_{\pm 1.33}$ & $29.55$ \\
    \bottomrule
  \end{tabular}
\end{table}

\begin{table}[H]
  \caption{Accuracy by seed for Qwen3-1.7B/8B on s1K. For fine-tuned models, seeds identify independent training runs; subscripts give the within-run standard deviation over evaluation attempts.}
  \label{tab_seed_qwen17_s1k}
  \vspace{3pt}
  \centering
  \scriptsize
  \setlength{\tabcolsep}{1.6pt}
  \begin{tabular}{lrcccccccc}
    \toprule
    \textbf{Method} & \textbf{Seed} & \textbf{AIME24} & \textbf{AIME25} & \textbf{AIME26} & \textbf{HMMT26} & \textbf{MATH500} & \textbf{Olympiad} & \textbf{GPQA-D} & \textbf{Average} \\
    \midrule
    Vanilla & 42 & $34.17_{\pm 6.11}$ & $31.67_{\pm 3.56}$ & $32.50_{\pm 5.27}$ & $24.24_{\pm 4.29}$ & $88.70_{\pm 0.74}$ & $56.34_{\pm 0.52}$ & $41.41_{\pm 2.70}$ & $44.15$ \\
    Vanilla & 1 & $31.67_{\pm 4.71}$ & $34.58_{\pm 5.33}$ & $36.25_{\pm 7.00}$ & $25.00_{\pm 2.14}$ & $89.00_{\pm 0.59}$ & $58.27_{\pm 1.03}$ & $41.79_{\pm 3.51}$ & $45.22$ \\
    Vanilla & 2026 & $29.58_{\pm 4.52}$ & $31.67_{\pm 3.56}$ & $31.67_{\pm 4.71}$ & $26.89_{\pm 3.00}$ & $88.45_{\pm 0.68}$ & $56.82_{\pm 1.55}$ & $43.43_{\pm 4.06}$ & $44.07$ \\
    \midrule
    Direct Distillation & 42 & $25.42_{\pm 4.34}$ & $26.25_{\pm 5.18}$ & $25.00_{\pm 4.36}$ & $25.00_{\pm 4.51}$ & $87.00_{\pm 0.67}$ & $54.04_{\pm 0.75}$ & $24.49_{\pm 2.56}$ & $38.17$ \\
    Direct Distillation & 1 & $23.75_{\pm 6.77}$ & $26.25_{\pm 4.86}$ & $27.92_{\pm 4.34}$ & $24.24_{\pm 3.97}$ & $87.10_{\pm 0.96}$ & $54.30_{\pm 0.56}$ & $27.40_{\pm 2.31}$ & $38.71$ \\
    Direct Distillation & 2026 & $25.42_{\pm 5.02}$ & $27.50_{\pm 4.63}$ & $30.00_{\pm 4.71}$ & $21.97_{\pm 3.14}$ & $87.20_{\pm 0.49}$ & $54.15_{\pm 0.77}$ & $26.14_{\pm 1.33}$ & $38.91$ \\
    \midrule
    Student RFT & 42 & $31.25_{\pm 5.62}$ & $37.92_{\pm 2.48}$ & $37.50_{\pm 3.45}$ & $26.14_{\pm 2.25}$ & $87.90_{\pm 1.09}$ & $57.31_{\pm 0.53}$ & $39.52_{\pm 2.24}$ & $45.36$ \\
    Student RFT & 1 & $35.00_{\pm 6.67}$ & $33.75_{\pm 3.30}$ & $35.00_{\pm 4.71}$ & $28.03_{\pm 3.53}$ & $88.30_{\pm 0.53}$ & $57.46_{\pm 1.03}$ & $38.26_{\pm 1.39}$ & $45.11$ \\
    Student RFT & 2026 & $33.33_{\pm 6.42}$ & $34.58_{\pm 7.33}$ & $34.58_{\pm 5.02}$ & $27.65_{\pm 2.53}$ & $88.80_{\pm 1.43}$ & $58.09_{\pm 0.15}$ & $40.03_{\pm 2.59}$ & $45.29$ \\
    \midrule
    Teacher RFT & 42 & $32.92_{\pm 4.15}$ & $35.42_{\pm 3.05}$ & $37.92_{\pm 9.42}$ & $26.52_{\pm 3.53}$ & $88.80_{\pm 0.69}$ & $58.46_{\pm 0.72}$ & $42.42_{\pm 1.30}$ & $46.07$ \\
    Teacher RFT & 1 & $29.58_{\pm 4.86}$ & $35.00_{\pm 5.04}$ & $38.33_{\pm 6.90}$ & $25.38_{\pm 5.11}$ & $88.90_{\pm 0.89}$ & $58.16_{\pm 1.73}$ & $42.30_{\pm 2.35}$ & $45.38$ \\
    Teacher RFT & 2026 & $30.42_{\pm 4.52}$ & $34.58_{\pm 8.15}$ & $41.67_{\pm 6.90}$ & $28.79_{\pm 2.81}$ & $89.15_{\pm 1.08}$ & $57.68_{\pm 0.43}$ & $41.41_{\pm 1.30}$ & $46.24$ \\
    \midrule
    IPD, S-P (1\%) & 42 & $33.75_{\pm 6.53}$ & $35.42_{\pm 5.89}$ & $40.00_{\pm 3.98}$ & $30.30_{\pm 4.58}$ & $89.35_{\pm 1.11}$ & $59.68_{\pm 0.41}$ & $41.29_{\pm 1.12}$ & $47.11$ \\
    IPD, S-P (1\%) & 1 & $34.17_{\pm 9.04}$ & $35.42_{\pm 3.05}$ & $39.58_{\pm 5.18}$ & $25.76_{\pm 5.37}$ & $89.60_{\pm 1.31}$ & $58.53_{\pm 1.12}$ & $41.79_{\pm 2.49}$ & $46.41$ \\
    IPD, S-P (1\%) & 2026 & $35.42_{\pm 6.65}$ & $34.58_{\pm 4.34}$ & $38.75_{\pm 7.75}$ & $26.89_{\pm 4.42}$ & $89.50_{\pm 0.76}$ & $58.16_{\pm 0.90}$ & $39.27_{\pm 1.33}$ & $46.08$ \\
    \midrule
    IPD, S-P (0.1\%) & 42 & $30.83_{\pm 6.11}$ & $35.00_{\pm 1.78}$ & $39.17_{\pm 5.27}$ & $27.65_{\pm 4.11}$ & $89.60_{\pm 0.97}$ & $58.57_{\pm 0.82}$ & $42.30_{\pm 1.82}$ & $46.16$ \\
    IPD, S-P (0.1\%) & 1 & $34.58_{\pm 3.96}$ & $33.33_{\pm 4.71}$ & $39.17_{\pm 7.29}$ & $28.41_{\pm 3.60}$ & $88.80_{\pm 1.28}$ & $58.98_{\pm 0.75}$ & $43.81_{\pm 1.62}$ & $46.73$ \\
    IPD, S-P (0.1\%) & 2026 & $36.67_{\pm 2.52}$ & $31.67_{\pm 5.35}$ & $43.33_{\pm 4.71}$ & $28.41_{\pm 2.78}$ & $89.75_{\pm 0.72}$ & $59.35_{\pm 0.80}$ & $42.55_{\pm 1.91}$ & $47.39$ \\
    \midrule
    IPD, S-P (0.01\%) & 42 & $33.33_{\pm 5.63}$ & $34.58_{\pm 3.05}$ & $39.17_{\pm 6.36}$ & $29.17_{\pm 3.60}$ & $88.70_{\pm 0.95}$ & $58.57_{\pm 0.25}$ & $41.16_{\pm 3.61}$ & $46.38$ \\
    IPD, S-P (0.01\%) & 1 & $35.83_{\pm 7.51}$ & $33.33_{\pm 4.71}$ & $33.33_{\pm 6.90}$ & $30.30_{\pm 4.58}$ & $89.35_{\pm 0.81}$ & $59.12_{\pm 1.42}$ & $44.32_{\pm 2.59}$ & $46.51$ \\
    IPD, S-P (0.01\%) & 2026 & $34.58_{\pm 8.15}$ & $35.83_{\pm 4.96}$ & $35.83_{\pm 4.96}$ & $30.30_{\pm 3.97}$ & $89.35_{\pm 1.04}$ & $58.57_{\pm 0.82}$ & $40.40_{\pm 2.40}$ & $46.41$ \\
    \midrule
    IPD, T-P (1\%) & 42 & $35.83_{\pm 7.07}$ & $35.83_{\pm 6.84}$ & $40.42_{\pm 5.47}$ & $28.79_{\pm 4.58}$ & $89.50_{\pm 0.58}$ & $59.79_{\pm 0.85}$ & $47.85_{\pm 1.91}$ & $48.29$ \\
    IPD, T-P (1\%) & 1 & $32.92_{\pm 4.86}$ & $37.92_{\pm 3.54}$ & $39.17_{\pm 6.36}$ & $26.14_{\pm 3.60}$ & $90.40_{\pm 0.43}$ & $59.64_{\pm 0.79}$ & $46.84_{\pm 1.62}$ & $47.57$ \\
    IPD, T-P (1\%) & 2026 & $32.08_{\pm 5.33}$ & $37.50_{\pm 3.45}$ & $41.25_{\pm 4.34}$ & $29.17_{\pm 5.36}$ & $90.10_{\pm 0.48}$ & $59.27_{\pm 0.79}$ & $46.46_{\pm 2.37}$ & $47.98$ \\
    \midrule
    IPD, T-P (0.1\%) & 42 & $37.92_{\pm 3.96}$ & $39.58_{\pm 5.47}$ & $42.08_{\pm 6.89}$ & $32.20_{\pm 3.21}$ & $89.55_{\pm 1.20}$ & $59.53_{\pm 0.49}$ & $44.70_{\pm 3.97}$ & $49.37$ \\
    IPD, T-P (0.1\%) & 1 & $34.58_{\pm 5.89}$ & $35.00_{\pm 3.09}$ & $39.17_{\pm 8.12}$ & $28.41_{\pm 2.25}$ & $90.45_{\pm 0.93}$ & $60.27_{\pm 0.53}$ & $43.56_{\pm 4.82}$ & $47.35$ \\
    IPD, T-P (0.1\%) & 2026 & $37.50_{\pm 5.56}$ & $37.92_{\pm 9.75}$ & $36.25_{\pm 3.30}$ & $27.27_{\pm 3.62}$ & $89.85_{\pm 1.11}$ & $59.94_{\pm 0.53}$ & $44.70_{\pm 0.65}$ & $47.63$ \\
    \midrule
    IPD, T-P (0.01\%) & 42 & $37.92_{\pm 5.89}$ & $36.67_{\pm 6.17}$ & $43.75_{\pm 7.22}$ & $30.30_{\pm 3.97}$ & $90.05_{\pm 0.87}$ & $60.01_{\pm 0.66}$ & $45.08_{\pm 2.16}$ & $49.11$ \\
    IPD, T-P (0.01\%) & 1 & $32.08_{\pm 8.72}$ & $39.58_{\pm 3.30}$ & $40.00_{\pm 7.13}$ & $32.20_{\pm 3.60}$ & $89.60_{\pm 0.95}$ & $59.61_{\pm 1.14}$ & $46.34_{\pm 2.87}$ & $48.49$ \\
    IPD, T-P (0.01\%) & 2026 & $33.33_{\pm 3.56}$ & $35.00_{\pm 3.98}$ & $41.67_{\pm 4.71}$ & $32.20_{\pm 3.21}$ & $89.70_{\pm 1.15}$ & $59.79_{\pm 0.95}$ & $44.44_{\pm 2.37}$ & $48.02$ \\
    \bottomrule
  \end{tabular}
\end{table}

\clearpage
\subsection{Problem-Solving Coverage}\label{app_coverage_results}

Tables~\ref{tab_passk_qwen_s1k}--\ref{tab_passk_qwen17_s1k} report Pass@$k$
for all four settings at the largest available budget: $k=8$ for AIME and HMMT
and $k=4$ for MATH500, OlympiadBench, and GPQA-D.
Results include per-seed coverage and its arithmetic mean; samples from different seeds are not pooled
into a larger sampling budget.
The tables support the 27-of-28 comparison in Section~\ref{sec_coverage},
including the two settings absent from Figure~\ref{fig_passk}.

\begin{figure}[H]
  \centering
  \includegraphics[width=\linewidth]{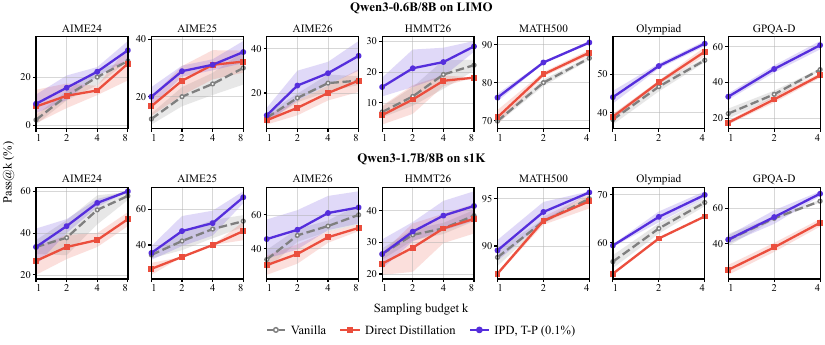}
  \caption{Pass@$k$ for Qwen3-0.6B/8B on LIMO and Qwen3-1.7B/8B on s1K. Fine-tuned curves average three training seeds; bands show one sample standard deviation.}
  \label{fig_passk_additional}
\end{figure}

Figure~\ref{fig_passk_additional} shows the full Pass@$k$ curves for Qwen3-0.6B/8B on LIMO and Qwen3-1.7B/8B on s1K.
IPD leads Vanilla and Direct Distillation on all fourteen benchmark--setting pairs at the largest matched sampling budget.

\begin{table}[H]
  \caption{Pass@8 on AIME and HMMT and Pass@4 on the other benchmarks for Qwen3-0.6B/8B on s1K. Mean rows average coverage across three seeds.}
  \label{tab_passk_qwen_s1k}
  \vspace{3pt}
  \centering
  \scriptsize
  \setlength{\tabcolsep}{3pt}
  \begin{tabular}{lrccccccc}
    \toprule
    & & \multicolumn{4}{c}{\textbf{Pass@8}} & \multicolumn{3}{c}{\textbf{Pass@4}} \\
    \cmidrule(lr){3-6}\cmidrule(lr){7-9}
    \textbf{Method} & \textbf{Seed} & \textbf{AIME24} & \textbf{AIME25} & \textbf{AIME26} & \textbf{HMMT26} & \textbf{MATH500} & \textbf{Olympiad} & \textbf{GPQA-D} \\
    \midrule
    Vanilla & 42 & $26.67$ & $33.33$ & $26.67$ & $21.21$ & $87.00$ & $53.41$ & $46.46$ \\
    Vanilla & 1 & $26.67$ & $33.33$ & $23.33$ & $21.21$ & $86.20$ & $53.12$ & $47.47$ \\
    Vanilla & 2026 & $26.67$ & $23.33$ & $26.67$ & $24.24$ & $86.20$ & $54.15$ & $47.47$ \\
    Vanilla & \textit{Mean} & $26.67$ & $30.00$ & $25.56$ & $22.22$ & $86.47$ & $53.56$ & $47.14$ \\
    \midrule
    Direct Distillation & 42 & $16.67$ & $30.00$ & $23.33$ & $27.27$ & $86.40$ & $50.30$ & $47.98$ \\
    Direct Distillation & 1 & $20.00$ & $33.33$ & $26.67$ & $24.24$ & $87.00$ & $52.08$ & $45.45$ \\
    Direct Distillation & 2026 & $16.67$ & $36.67$ & $26.67$ & $21.21$ & $88.40$ & $50.59$ & $48.48$ \\
    Direct Distillation & \textit{Mean} & $17.78$ & $33.33$ & $25.56$ & $24.24$ & $87.27$ & $50.99$ & $47.31$ \\
    \midrule
    Student RFT & 42 & $23.33$ & $40.00$ & $20.00$ & $18.18$ & $86.80$ & $53.56$ & $52.53$ \\
    Student RFT & 1 & $26.67$ & $30.00$ & $23.33$ & $18.18$ & $85.40$ & $53.71$ & $55.56$ \\
    Student RFT & 2026 & $30.00$ & $33.33$ & $23.33$ & $21.21$ & $87.00$ & $52.37$ & $57.07$ \\
    Student RFT & \textit{Mean} & $26.67$ & $34.44$ & $22.22$ & $19.19$ & $86.40$ & $53.21$ & $55.05$ \\
    \midrule
    Teacher RFT & 42 & $33.33$ & $40.00$ & $26.67$ & $30.30$ & $88.00$ & $56.08$ & $52.53$ \\
    Teacher RFT & 1 & $23.33$ & $30.00$ & $16.67$ & $27.27$ & $89.80$ & $56.23$ & $57.58$ \\
    Teacher RFT & 2026 & $26.67$ & $40.00$ & $36.67$ & $24.24$ & $88.40$ & $55.19$ & $54.04$ \\
    Teacher RFT & \textit{Mean} & $27.78$ & $36.67$ & $26.67$ & $27.27$ & $88.73$ & $55.84$ & $54.71$ \\
    \midrule
    IPD, S-P (1\%) & 42 & $30.00$ & $36.67$ & $40.00$ & $27.27$ & $88.00$ & $57.57$ & $52.02$ \\
    IPD, S-P (1\%) & 1 & $33.33$ & $36.67$ & $30.00$ & $15.15$ & $90.20$ & $57.72$ & $48.99$ \\
    IPD, S-P (1\%) & 2026 & $30.00$ & $40.00$ & $43.33$ & $30.30$ & $90.40$ & $56.97$ & $50.51$ \\
    IPD, S-P (1\%) & \textit{Mean} & $31.11$ & $37.78$ & $37.78$ & $24.24$ & $89.53$ & $57.42$ & $50.51$ \\
    \midrule
    IPD, S-P (0.1\%) & 42 & $26.67$ & $46.67$ & $30.00$ & $27.27$ & $90.60$ & $55.49$ & $48.99$ \\
    IPD, S-P (0.1\%) & 1 & $23.33$ & $36.67$ & $23.33$ & $24.24$ & $90.00$ & $55.93$ & $47.98$ \\
    IPD, S-P (0.1\%) & 2026 & $30.00$ & $36.67$ & $26.67$ & $27.27$ & $90.20$ & $54.60$ & $52.53$ \\
    IPD, S-P (0.1\%) & \textit{Mean} & $26.67$ & $40.00$ & $26.67$ & $26.26$ & $90.27$ & $55.34$ & $49.83$ \\
    \midrule
    IPD, S-P (0.01\%) & 42 & $33.33$ & $33.33$ & $30.00$ & $24.24$ & $89.60$ & $57.42$ & $55.05$ \\
    IPD, S-P (0.01\%) & 1 & $33.33$ & $40.00$ & $23.33$ & $24.24$ & $87.20$ & $55.34$ & $60.61$ \\
    IPD, S-P (0.01\%) & 2026 & $30.00$ & $40.00$ & $36.67$ & $27.27$ & $89.20$ & $56.53$ & $55.56$ \\
    IPD, S-P (0.01\%) & \textit{Mean} & $32.22$ & $37.78$ & $30.00$ & $25.25$ & $88.67$ & $56.43$ & $57.07$ \\
    \midrule
    IPD, T-P (1\%) & 42 & $33.33$ & $43.33$ & $30.00$ & $24.24$ & $90.20$ & $58.61$ & $67.17$ \\
    IPD, T-P (1\%) & 1 & $30.00$ & $46.67$ & $40.00$ & $27.27$ & $90.60$ & $58.16$ & $57.07$ \\
    IPD, T-P (1\%) & 2026 & $40.00$ & $33.33$ & $23.33$ & $36.36$ & $89.80$ & $56.08$ & $60.61$ \\
    IPD, T-P (1\%) & \textit{Mean} & $34.44$ & $41.11$ & $31.11$ & $29.29$ & $90.20$ & $57.62$ & $61.62$ \\
    \midrule
    IPD, T-P (0.1\%) & 42 & $33.33$ & $43.33$ & $36.67$ & $24.24$ & $90.00$ & $57.57$ & $57.58$ \\
    IPD, T-P (0.1\%) & 1 & $26.67$ & $40.00$ & $26.67$ & $33.33$ & $88.80$ & $56.38$ & $61.62$ \\
    IPD, T-P (0.1\%) & 2026 & $30.00$ & $40.00$ & $33.33$ & $33.33$ & $89.80$ & $56.53$ & $59.60$ \\
    IPD, T-P (0.1\%) & \textit{Mean} & $30.00$ & $41.11$ & $32.22$ & $30.30$ & $89.53$ & $56.82$ & $59.60$ \\
    \midrule
    IPD, T-P (0.01\%) & 42 & $33.33$ & $30.00$ & $33.33$ & $30.30$ & $88.80$ & $56.53$ & $62.63$ \\
    IPD, T-P (0.01\%) & 1 & $33.33$ & $36.67$ & $30.00$ & $39.39$ & $88.60$ & $54.90$ & $69.19$ \\
    IPD, T-P (0.01\%) & 2026 & $26.67$ & $40.00$ & $30.00$ & $21.21$ & $89.40$ & $55.34$ & $68.18$ \\
    IPD, T-P (0.01\%) & \textit{Mean} & $31.11$ & $35.56$ & $31.11$ & $30.30$ & $88.93$ & $55.59$ & $66.67$ \\
    \bottomrule
  \end{tabular}
\end{table}

\begin{table}[H]
  \caption{Pass@8 on AIME and HMMT and Pass@4 on the other benchmarks for EXAONE 4.0 1.2B/32B on s1K. Mean rows average coverage across three seeds.}
  \label{tab_passk_exaone_s1k}
  \vspace{3pt}
  \centering
  \scriptsize
  \setlength{\tabcolsep}{3pt}
  \begin{tabular}{lrccccccc}
    \toprule
    & & \multicolumn{4}{c}{\textbf{Pass@8}} & \multicolumn{3}{c}{\textbf{Pass@4}} \\
    \cmidrule(lr){3-6}\cmidrule(lr){7-9}
    \textbf{Method} & \textbf{Seed} & \textbf{AIME24} & \textbf{AIME25} & \textbf{AIME26} & \textbf{HMMT26} & \textbf{MATH500} & \textbf{Olympiad} & \textbf{GPQA-D} \\
    \midrule
    Vanilla & 42 & $60.00$ & $70.00$ & $76.67$ & $45.45$ & $95.40$ & $80.42$ & $74.24$ \\
    Vanilla & 1 & $60.00$ & $66.67$ & $73.33$ & $54.55$ & $96.40$ & $80.42$ & $76.26$ \\
    Vanilla & 2026 & $63.33$ & $70.00$ & $80.00$ & $51.52$ & $95.60$ & $79.08$ & $73.74$ \\
    Vanilla & \textit{Mean} & $61.11$ & $68.89$ & $76.67$ & $50.51$ & $95.80$ & $79.97$ & $74.75$ \\
    \midrule
    Direct Distillation & 42 & $56.67$ & $73.33$ & $73.33$ & $39.39$ & $96.00$ & $70.62$ & $59.09$ \\
    Direct Distillation & 1 & $60.00$ & $60.00$ & $70.00$ & $39.39$ & $96.00$ & $69.88$ & $57.58$ \\
    Direct Distillation & 2026 & $53.33$ & $73.33$ & $70.00$ & $42.42$ & $96.60$ & $70.03$ & $59.60$ \\
    Direct Distillation & \textit{Mean} & $56.67$ & $68.89$ & $71.11$ & $40.40$ & $96.20$ & $70.18$ & $58.75$ \\
    \midrule
    Student RFT & 42 & $60.00$ & $73.33$ & $80.00$ & $54.55$ & $96.00$ & $80.27$ & $73.74$ \\
    Student RFT & 1 & $56.67$ & $70.00$ & $73.33$ & $51.52$ & $96.00$ & $80.27$ & $73.74$ \\
    Student RFT & 2026 & $63.33$ & $73.33$ & $76.67$ & $54.55$ & $96.60$ & $80.86$ & $76.77$ \\
    Student RFT & \textit{Mean} & $60.00$ & $72.22$ & $76.67$ & $53.54$ & $96.20$ & $80.46$ & $74.75$ \\
    \midrule
    Teacher RFT & 42 & $73.33$ & $76.67$ & $83.33$ & $60.61$ & $97.60$ & $79.53$ & $76.26$ \\
    Teacher RFT & 1 & $66.67$ & $83.33$ & $80.00$ & $60.61$ & $97.00$ & $81.60$ & $76.77$ \\
    Teacher RFT & 2026 & $63.33$ & $83.33$ & $80.00$ & $54.55$ & $96.60$ & $82.49$ & $76.77$ \\
    Teacher RFT & \textit{Mean} & $67.78$ & $81.11$ & $81.11$ & $58.59$ & $97.07$ & $81.21$ & $76.60$ \\
    \midrule
    IPD, S-P (1\%) & 42 & $56.67$ & $80.00$ & $90.00$ & $54.55$ & $96.40$ & $82.34$ & $75.25$ \\
    IPD, S-P (1\%) & 1 & $63.33$ & $73.33$ & $86.67$ & $51.52$ & $96.00$ & $82.20$ & $75.76$ \\
    IPD, S-P (1\%) & 2026 & $70.00$ & $83.33$ & $86.67$ & $54.55$ & $95.80$ & $81.31$ & $74.24$ \\
    IPD, S-P (1\%) & \textit{Mean} & $63.33$ & $78.89$ & $87.78$ & $53.54$ & $96.07$ & $81.95$ & $75.08$ \\
    \midrule
    IPD, S-P (0.1\%) & 42 & $73.33$ & $73.33$ & $83.33$ & $57.58$ & $97.40$ & $81.16$ & $72.22$ \\
    IPD, S-P (0.1\%) & 1 & $63.33$ & $80.00$ & $80.00$ & $48.48$ & $96.40$ & $83.09$ & $76.26$ \\
    IPD, S-P (0.1\%) & 2026 & $70.00$ & $70.00$ & $76.67$ & $63.64$ & $96.40$ & $83.38$ & $79.29$ \\
    IPD, S-P (0.1\%) & \textit{Mean} & $68.89$ & $74.44$ & $80.00$ & $56.57$ & $96.73$ & $82.54$ & $75.93$ \\
    \midrule
    IPD, S-P (0.01\%) & 42 & $66.67$ & $80.00$ & $86.67$ & $57.58$ & $96.20$ & $80.86$ & $75.76$ \\
    IPD, S-P (0.01\%) & 1 & $66.67$ & $76.67$ & $83.33$ & $63.64$ & $95.20$ & $83.09$ & $83.84$ \\
    IPD, S-P (0.01\%) & 2026 & $63.33$ & $76.67$ & $80.00$ & $60.61$ & $95.80$ & $81.31$ & $78.28$ \\
    IPD, S-P (0.01\%) & \textit{Mean} & $65.56$ & $77.78$ & $83.33$ & $60.61$ & $95.73$ & $81.75$ & $79.29$ \\
    \midrule
    IPD, T-P (1\%) & 42 & $63.33$ & $80.00$ & $83.33$ & $54.55$ & $97.00$ & $82.79$ & $75.25$ \\
    IPD, T-P (1\%) & 1 & $66.67$ & $80.00$ & $86.67$ & $54.55$ & $96.20$ & $82.05$ & $81.82$ \\
    IPD, T-P (1\%) & 2026 & $66.67$ & $83.33$ & $83.33$ & $57.58$ & $97.40$ & $82.64$ & $76.77$ \\
    IPD, T-P (1\%) & \textit{Mean} & $65.56$ & $81.11$ & $84.44$ & $55.56$ & $96.87$ & $82.49$ & $77.95$ \\
    \midrule
    IPD, T-P (0.1\%) & 42 & $66.67$ & $76.67$ & $83.33$ & $60.61$ & $96.60$ & $79.97$ & $76.77$ \\
    IPD, T-P (0.1\%) & 1 & $60.00$ & $80.00$ & $80.00$ & $51.52$ & $97.00$ & $79.38$ & $74.75$ \\
    IPD, T-P (0.1\%) & 2026 & $66.67$ & $73.33$ & $83.33$ & $60.61$ & $96.60$ & $79.82$ & $77.27$ \\
    IPD, T-P (0.1\%) & \textit{Mean} & $64.44$ & $76.67$ & $82.22$ & $57.58$ & $96.73$ & $79.72$ & $76.26$ \\
    \midrule
    IPD, T-P (0.01\%) & 42 & $66.67$ & $76.67$ & $83.33$ & $54.55$ & $96.20$ & $82.05$ & $73.23$ \\
    IPD, T-P (0.01\%) & 1 & $60.00$ & $80.00$ & $83.33$ & $48.48$ & $97.40$ & $83.23$ & $73.23$ \\
    IPD, T-P (0.01\%) & 2026 & $63.33$ & $80.00$ & $86.67$ & $57.58$ & $96.80$ & $82.49$ & $76.26$ \\
    IPD, T-P (0.01\%) & \textit{Mean} & $63.33$ & $78.89$ & $84.44$ & $53.54$ & $96.80$ & $82.59$ & $74.24$ \\
    \bottomrule
  \end{tabular}
\end{table}

\begin{table}[H]
  \caption{Pass@8 on AIME and HMMT and Pass@4 on the other benchmarks for Qwen3-0.6B/8B on LIMO. Mean rows average coverage across three seeds.}
  \label{tab_passk_qwen_limo}
  \vspace{3pt}
  \centering
  \scriptsize
  \setlength{\tabcolsep}{3pt}
  \begin{tabular}{lrccccccc}
    \toprule
    & & \multicolumn{4}{c}{\textbf{Pass@8}} & \multicolumn{3}{c}{\textbf{Pass@4}} \\
    \cmidrule(lr){3-6}\cmidrule(lr){7-9}
    \textbf{Method} & \textbf{Seed} & \textbf{AIME24} & \textbf{AIME25} & \textbf{AIME26} & \textbf{HMMT26} & \textbf{MATH500} & \textbf{Olympiad} & \textbf{GPQA-D} \\
    \midrule
    Vanilla & 42 & $26.67$ & $33.33$ & $26.67$ & $21.21$ & $87.00$ & $53.41$ & $46.46$ \\
    Vanilla & 1 & $26.67$ & $33.33$ & $23.33$ & $21.21$ & $86.20$ & $53.12$ & $47.47$ \\
    Vanilla & 2026 & $26.67$ & $23.33$ & $26.67$ & $24.24$ & $86.20$ & $54.15$ & $47.47$ \\
    Vanilla & \textit{Mean} & $26.67$ & $30.00$ & $25.56$ & $22.22$ & $86.47$ & $53.56$ & $47.14$ \\
    \midrule
    Direct Distillation & 42 & $33.33$ & $30.00$ & $30.00$ & $18.18$ & $87.40$ & $55.79$ & $44.44$ \\
    Direct Distillation & 1 & $20.00$ & $36.67$ & $20.00$ & $18.18$ & $87.60$ & $54.60$ & $44.95$ \\
    Direct Distillation & 2026 & $23.33$ & $30.00$ & $26.67$ & $18.18$ & $88.80$ & $56.97$ & $42.42$ \\
    Direct Distillation & \textit{Mean} & $25.56$ & $32.22$ & $25.56$ & $18.18$ & $87.93$ & $55.79$ & $43.94$ \\
    \midrule
    Student RFT & 42 & $13.33$ & $40.00$ & $23.33$ & $24.24$ & $88.20$ & $51.63$ & $50.00$ \\
    Student RFT & 1 & $20.00$ & $40.00$ & $16.67$ & $21.21$ & $85.60$ & $53.12$ & $48.99$ \\
    Student RFT & 2026 & $30.00$ & $30.00$ & $26.67$ & $18.18$ & $86.00$ & $54.60$ & $51.01$ \\
    Student RFT & \textit{Mean} & $21.11$ & $36.67$ & $22.22$ & $21.21$ & $86.60$ & $53.12$ & $50.00$ \\
    \midrule
    Teacher RFT & 42 & $23.33$ & $43.33$ & $36.67$ & $27.27$ & $88.60$ & $55.64$ & $58.59$ \\
    Teacher RFT & 1 & $30.00$ & $40.00$ & $40.00$ & $24.24$ & $88.00$ & $56.82$ & $51.01$ \\
    Teacher RFT & 2026 & $23.33$ & $40.00$ & $33.33$ & $24.24$ & $88.20$ & $54.90$ & $53.54$ \\
    Teacher RFT & \textit{Mean} & $25.56$ & $41.11$ & $36.67$ & $25.25$ & $88.27$ & $55.79$ & $54.38$ \\
    \midrule
    IPD, S-P (1\%) & 42 & $23.33$ & $40.00$ & $33.33$ & $24.24$ & $90.40$ & $56.38$ & $53.03$ \\
    IPD, S-P (1\%) & 1 & $33.33$ & $46.67$ & $26.67$ & $27.27$ & $90.20$ & $59.20$ & $54.04$ \\
    IPD, S-P (1\%) & 2026 & $23.33$ & $43.33$ & $33.33$ & $24.24$ & $89.40$ & $57.86$ & $52.02$ \\
    IPD, S-P (1\%) & \textit{Mean} & $26.67$ & $43.33$ & $31.11$ & $25.25$ & $90.00$ & $57.81$ & $53.03$ \\
    \midrule
    IPD, S-P (0.1\%) & 42 & $16.67$ & $43.33$ & $23.33$ & $30.30$ & $89.80$ & $57.27$ & $52.53$ \\
    IPD, S-P (0.1\%) & 1 & $23.33$ & $43.33$ & $30.00$ & $27.27$ & $89.20$ & $55.64$ & $56.06$ \\
    IPD, S-P (0.1\%) & 2026 & $26.67$ & $36.67$ & $30.00$ & $27.27$ & $89.60$ & $55.64$ & $52.02$ \\
    IPD, S-P (0.1\%) & \textit{Mean} & $22.22$ & $41.11$ & $27.78$ & $28.28$ & $89.53$ & $56.18$ & $53.54$ \\
    \midrule
    IPD, S-P (0.01\%) & 42 & $33.33$ & $43.33$ & $33.33$ & $24.24$ & $89.00$ & $57.27$ & $52.53$ \\
    IPD, S-P (0.01\%) & 1 & $30.00$ & $36.67$ & $26.67$ & $24.24$ & $89.00$ & $55.64$ & $51.01$ \\
    IPD, S-P (0.01\%) & 2026 & $33.33$ & $26.67$ & $30.00$ & $24.24$ & $87.80$ & $55.49$ & $50.00$ \\
    IPD, S-P (0.01\%) & \textit{Mean} & $32.22$ & $35.56$ & $30.00$ & $24.24$ & $88.60$ & $56.13$ & $51.18$ \\
    \midrule
    IPD, T-P (1\%) & 42 & $30.00$ & $36.67$ & $36.67$ & $27.27$ & $91.60$ & $58.46$ & $56.57$ \\
    IPD, T-P (1\%) & 1 & $30.00$ & $36.67$ & $33.33$ & $21.21$ & $91.80$ & $57.27$ & $56.06$ \\
    IPD, T-P (1\%) & 2026 & $30.00$ & $36.67$ & $30.00$ & $30.30$ & $89.20$ & $58.61$ & $58.59$ \\
    IPD, T-P (1\%) & \textit{Mean} & $30.00$ & $36.67$ & $33.33$ & $26.26$ & $90.87$ & $58.11$ & $57.07$ \\
    \midrule
    IPD, T-P (0.1\%) & 42 & $26.67$ & $33.33$ & $43.33$ & $27.27$ & $91.00$ & $58.61$ & $58.08$ \\
    IPD, T-P (0.1\%) & 1 & $33.33$ & $33.33$ & $30.00$ & $27.27$ & $90.20$ & $57.27$ & $62.12$ \\
    IPD, T-P (0.1\%) & 2026 & $33.33$ & $40.00$ & $36.67$ & $30.30$ & $90.60$ & $57.72$ & $62.12$ \\
    IPD, T-P (0.1\%) & \textit{Mean} & $31.11$ & $35.56$ & $36.67$ & $28.28$ & $90.60$ & $57.86$ & $60.77$ \\
    \midrule
    IPD, T-P (0.01\%) & 42 & $26.67$ & $50.00$ & $33.33$ & $30.30$ & $89.00$ & $57.27$ & $56.57$ \\
    IPD, T-P (0.01\%) & 1 & $30.00$ & $36.67$ & $36.67$ & $27.27$ & $89.80$ & $58.90$ & $55.56$ \\
    IPD, T-P (0.01\%) & 2026 & $40.00$ & $36.67$ & $30.00$ & $27.27$ & $88.00$ & $59.35$ & $56.06$ \\
    IPD, T-P (0.01\%) & \textit{Mean} & $32.22$ & $41.11$ & $33.33$ & $28.28$ & $88.93$ & $58.51$ & $56.06$ \\
    \bottomrule
  \end{tabular}
\end{table}

\begin{table}[H]
  \caption{Pass@8 on AIME and HMMT and Pass@4 on the other benchmarks for Qwen3-1.7B/8B on s1K. Mean rows average coverage across three seeds.}
  \label{tab_passk_qwen17_s1k}
  \vspace{3pt}
  \centering
  \scriptsize
  \setlength{\tabcolsep}{3pt}
  \begin{tabular}{lrccccccc}
    \toprule
    & & \multicolumn{4}{c}{\textbf{Pass@8}} & \multicolumn{3}{c}{\textbf{Pass@4}} \\
    \cmidrule(lr){3-6}\cmidrule(lr){7-9}
    \textbf{Method} & \textbf{Seed} & \textbf{AIME24} & \textbf{AIME25} & \textbf{AIME26} & \textbf{HMMT26} & \textbf{MATH500} & \textbf{Olympiad} & \textbf{GPQA-D} \\
    \midrule
    Vanilla & 42 & $60.00$ & $50.00$ & $56.67$ & $39.39$ & $95.20$ & $67.51$ & $64.14$ \\
    Vanilla & 1 & $56.67$ & $56.67$ & $66.67$ & $39.39$ & $94.60$ & $68.99$ & $64.65$ \\
    Vanilla & 2026 & $56.67$ & $53.33$ & $56.67$ & $36.36$ & $95.20$ & $68.55$ & $63.13$ \\
    Vanilla & \textit{Mean} & $57.78$ & $53.33$ & $60.00$ & $38.38$ & $95.00$ & $68.35$ & $63.97$ \\
    \midrule
    Direct Distillation & 42 & $43.33$ & $53.33$ & $53.33$ & $36.36$ & $94.40$ & $65.28$ & $49.49$ \\
    Direct Distillation & 1 & $46.67$ & $46.67$ & $50.00$ & $42.42$ & $94.20$ & $65.58$ & $53.54$ \\
    Direct Distillation & 2026 & $50.00$ & $43.33$ & $53.33$ & $33.33$ & $95.60$ & $65.43$ & $52.53$ \\
    Direct Distillation & \textit{Mean} & $46.67$ & $47.78$ & $52.22$ & $37.37$ & $94.73$ & $65.43$ & $51.85$ \\
    \midrule
    Student RFT & 42 & $56.67$ & $53.33$ & $63.33$ & $42.42$ & $94.40$ & $67.36$ & $62.63$ \\
    Student RFT & 1 & $56.67$ & $53.33$ & $60.00$ & $45.45$ & $95.00$ & $68.69$ & $58.08$ \\
    Student RFT & 2026 & $63.33$ & $53.33$ & $53.33$ & $45.45$ & $96.00$ & $68.84$ & $62.12$ \\
    Student RFT & \textit{Mean} & $58.89$ & $53.33$ & $58.89$ & $44.44$ & $95.13$ & $68.30$ & $60.94$ \\
    \midrule
    Teacher RFT & 42 & $53.33$ & $60.00$ & $66.67$ & $39.39$ & $95.40$ & $67.51$ & $69.70$ \\
    Teacher RFT & 1 & $53.33$ & $63.33$ & $70.00$ & $36.36$ & $95.20$ & $68.99$ & $64.65$ \\
    Teacher RFT & 2026 & $63.33$ & $60.00$ & $73.33$ & $45.45$ & $96.60$ & $67.80$ & $68.69$ \\
    Teacher RFT & \textit{Mean} & $56.67$ & $61.11$ & $70.00$ & $40.40$ & $95.73$ & $68.10$ & $67.68$ \\
    \midrule
    IPD, S-P (1\%) & 42 & $56.67$ & $50.00$ & $63.33$ & $42.42$ & $95.20$ & $68.69$ & $63.13$ \\
    IPD, S-P (1\%) & 1 & $56.67$ & $56.67$ & $66.67$ & $39.39$ & $94.80$ & $67.51$ & $62.63$ \\
    IPD, S-P (1\%) & 2026 & $60.00$ & $53.33$ & $66.67$ & $39.39$ & $94.40$ & $67.21$ & $62.12$ \\
    IPD, S-P (1\%) & \textit{Mean} & $57.78$ & $53.33$ & $65.56$ & $40.40$ & $94.80$ & $67.80$ & $62.63$ \\
    \midrule
    IPD, S-P (0.1\%) & 42 & $56.67$ & $53.33$ & $63.33$ & $36.36$ & $95.00$ & $69.14$ & $62.63$ \\
    IPD, S-P (0.1\%) & 1 & $56.67$ & $60.00$ & $66.67$ & $39.39$ & $95.60$ & $69.29$ & $65.15$ \\
    IPD, S-P (0.1\%) & 2026 & $53.33$ & $53.33$ & $70.00$ & $42.42$ & $95.40$ & $69.44$ & $63.64$ \\
    IPD, S-P (0.1\%) & \textit{Mean} & $55.56$ & $55.56$ & $66.67$ & $39.39$ & $95.33$ & $69.29$ & $63.80$ \\
    \midrule
    IPD, S-P (0.01\%) & 42 & $60.00$ & $56.67$ & $63.33$ & $42.42$ & $94.60$ & $69.44$ & $59.09$ \\
    IPD, S-P (0.01\%) & 1 & $60.00$ & $53.33$ & $63.33$ & $39.39$ & $95.20$ & $70.47$ & $70.71$ \\
    IPD, S-P (0.01\%) & 2026 & $56.67$ & $63.33$ & $66.67$ & $45.45$ & $95.40$ & $67.80$ & $61.11$ \\
    IPD, S-P (0.01\%) & \textit{Mean} & $58.89$ & $57.78$ & $64.44$ & $42.42$ & $95.07$ & $69.24$ & $63.64$ \\
    \midrule
    IPD, T-P (1\%) & 42 & $63.33$ & $56.67$ & $66.67$ & $42.42$ & $95.40$ & $70.92$ & $69.70$ \\
    IPD, T-P (1\%) & 1 & $56.67$ & $56.67$ & $70.00$ & $39.39$ & $96.40$ & $70.03$ & $70.71$ \\
    IPD, T-P (1\%) & 2026 & $53.33$ & $63.33$ & $60.00$ & $42.42$ & $95.20$ & $68.55$ & $68.18$ \\
    IPD, T-P (1\%) & \textit{Mean} & $57.78$ & $58.89$ & $65.56$ & $41.41$ & $95.67$ & $69.83$ & $69.53$ \\
    \midrule
    IPD, T-P (0.1\%) & 42 & $60.00$ & $63.33$ & $70.00$ & $45.45$ & $95.60$ & $69.14$ & $68.69$ \\
    IPD, T-P (0.1\%) & 1 & $60.00$ & $66.67$ & $70.00$ & $36.36$ & $95.80$ & $70.62$ & $69.19$ \\
    IPD, T-P (0.1\%) & 2026 & $60.00$ & $70.00$ & $53.33$ & $42.42$ & $95.60$ & $70.03$ & $66.67$ \\
    IPD, T-P (0.1\%) & \textit{Mean} & $60.00$ & $66.67$ & $64.44$ & $41.41$ & $95.67$ & $69.93$ & $68.18$ \\
    \midrule
    IPD, T-P (0.01\%) & 42 & $60.00$ & $60.00$ & $76.67$ & $42.42$ & $95.20$ & $68.84$ & $70.71$ \\
    IPD, T-P (0.01\%) & 1 & $53.33$ & $63.33$ & $60.00$ & $42.42$ & $95.00$ & $69.73$ & $67.68$ \\
    IPD, T-P (0.01\%) & 2026 & $53.33$ & $56.67$ & $66.67$ & $45.45$ & $96.20$ & $69.29$ & $67.17$ \\
    IPD, T-P (0.01\%) & \textit{Mean} & $55.56$ & $60.00$ & $67.78$ & $43.43$ & $95.47$ & $69.29$ & $68.52$ \\
    \bottomrule
  \end{tabular}
\end{table}

% \clearpage

% \section{Limitations and Broader Impacts}\label{app_limitations}

% \paragraph{Limitations.}
% This study targets reasoning models with at most 1.7B student parameters, so it
% does not establish that the same distillation behavior persists for larger
% students. The evaluation covers mathematical reasoning and GPQA-D rather than open-ended
% generation or interactive agents.

% \paragraph{Broader impacts.}
% Transferring stronger reasoning into smaller models reduces inference cost and
% broadens access to local reasoning. The same efficiency can lower the cost of
% misuse, and students can inherit teacher errors or biases, so high-stakes uses
% require domain-specific evaluation.

\end{document}